%% file: main.tex
\documentclass[review]{elsarticle}

\usepackage{lineno,hyperref}

\usepackage{float}
\usepackage{array}
\usepackage{longtable}
\usepackage{xspace}
\usepackage{amssymb}
\usepackage{amsmath}
\usepackage{paralist}
\usepackage{xfrac}
\usepackage{rotating}
\usepackage{multirow}
\usepackage{epstopdf}
\usepackage{graphicx}

\usepackage{microtype}
\DeclareMathAlphabet{\mathcal}{OMS}{cmsy}{m}{n}  
\usepackage[usenames,dvipsnames]{xcolor}
\usepackage[normalem]{ulem}  

\graphicspath{{images/}}

\definecolor{green}{rgb}{0.0,0.8,0.0}
\newcommand{\etal}{{~et~al.~}}

\newcommand{\kintof}{{Kinect$^{\text{ToF}}$}}
\newcommand{\kinsl}{{Kinect$^{\text{SL}}$}}

\newcommand{\arctann}{\operatorname{arctan2}\nolimits}

\journal{Journal of Computer Vision and Image Understanding }

\begin{document}

\begin{frontmatter}

\title{Kinect Range Sensing: Structured-Light versus Time-of-Flight Kinect}

\author{Hamed Sarbolandi}
\cortext[]{Corresponding author}
\ead{hamed.sarbolandi@uni-siegen.de}
\author{Damien Lefloch}
\author{Andreas Kolb}

\address{Institute for Vision and Graphics, University of Siegen, Germany}




\begin{abstract}
  Recently, the new Kinect One has been issued by Microsoft, providing
  the next generation of real-time range sensing devices based on the
  Time-of-Flight (ToF) principle. As the first Kinect version was using a
  structured light approach, one would expect various differences in
  the characteristics of the range data delivered by both devices.

  This paper presents a detailed and in-depth comparison between both
  devices. In order to conduct the comparison, we propose a framework
  of seven different experimental setups, which is a generic basis
  for evaluating range cameras such as Kinect. The experiments have
  been designed with the goal to capture individual effects of the
  Kinect devices as isolatedly as possible and in a way, that they can
  also be adopted, in order to apply them to any other range sensing
  device. The overall goal of this paper is to provide a solid insight
  into the pros and cons of either device. Thus, scientists that are
  interested in using Kinect range sensing cameras in their specific
  application scenario can directly assess the expected, specific
  benefits and potential problem of either device.
\end{abstract}

\begin{keyword}
Depth Sensor\sep 3D \sep Kinect \sep Evaluation
\end{keyword}

\end{frontmatter}


\section{Introduction and Related Works}
\label{sec:intro}
In the last decade, several new range sensing devices have been
developed and have been made available for application development at
affordable costs. In 2010, Microsoft, in cooperation with PrimeSense
released a structured-light (SL) based range sensing camera, the
so-called Kinect\texttrademark, that delivers reliable depth images at
VGA resolution at 30 Hz, coupled with an RGB-color camera at the same
image resolution. Even though the camera was mainly designed for
gaming, it achieved great popularity in the scientific community where
researchers have developed a huge amount of innovative applications
that are related to different fields such as online 3D
reconstruction~\cite{keller:2013:PBF, newcombe:2011:KINFU,
  niessner:2013:KINFUH}, medical applications and health
care~\cite{gallo2011controller, bauer13range-in-healtcare}, augmented
reality~\cite{vera2011augmented}, etc.  Recently Microsoft released an
update of their Kinect\texttrademark\ camera in the context of their
next generation of console (XBox One) that is now based on
Time-of-Flight (ToF) principle.

Both range sensing principles, SL and ToF, are quite different and are
subject to a variety of error sources (see Sec.~\ref{sec:principle}).
This paper is meant to deeply evaluate both Kinect\texttrademark
cameras, denoted as \kinsl\ and \kintof\ in the following, in order to
extract their pros and cons which are relevant for any application
incorporating this kind of device. Thus, we explicitly do not try to
evaluate the devices with respect to a set of specific application
scenarios, but we designed a set of seven different experimental setups
as a generic basis for evaluating range cameras such as Kinect.

Several studies can be found in the literature that compare and
evaluate the depth precision of both principles. However, this work is
the first study comparing both versions of the Kinect cameras and
offering detailed descriptions under which conditions one is superior
to the other. Since Kinect\texttrademark cameras are targeting the
consumer market and have known sales of several millions devices, we
believe that our work will be valuable for a large number of follow-up
research projects.

\paragraph{Prior Work}

A complete discussion on prior work in SL- and ToF-based range sensing
would clearly go beyond the scope of this paper. Thus, we give a brief
and exemplary overview on related work in the context SL- and
ToF-based range sensing and focus on papers that compare different
range sensing approaches and devices. In Sec.~\ref{sec:principle.err}
we further refer to some key papers that deal with specific
characteristics of SL and ToF range data. Additionally, we refer the
reader to the surveys of Berger\etal\cite{berger2013state} and
Han\etal\cite{han2013enhanced} on the \kinsl\, as well as to the
survey on Time-of-Flight cameras by Kolb\etal\cite{kolbTOF10}.

Kuhnert and Stommel~\cite{kuhnert06fusion} demonstrate a first
integration of ToF- and stereo cameras.
Beder\etal\cite{beder:2007:CompTOFSV} evaluate and compare ToF cameras
to a stereo-vision setup. Both papers emphasize that ToF and stereo
data are at least partially complementary and thus an integration
significantly improves the quality of range data. Furthermore, the \kintof\ does not use triangulation for depth calculation, and thus it does not suffer much from occlusion. As it will be shown in Sec.~\ref{sec:results:dynamic}, the occluded area in a static scene is around 5\% compared to \kinsl\ which is around 20\%. Besides Evangelidis\etal\cite{evangelidis2015fusion} has also used a ToF range camera, in comparison with \kinsl, \kintof\ would be a better choice specifically to be utilized in depth-stereo approach. For further details
on ToF-stereo fusion we refer the reader to
Nair\etal\cite{nair13tof-stereo-fusion}.
In the domain of robotics,
Wiedemann\etal\cite{wiedemann:2008:TOFCharact} compare different ToF
cameras from different manufacturers. They analyze the sensor
characteristics of such systems and the application potential for
mobile robots. In their work, they address several problems such as
sensor calibration, automatic integration time and data filtering
schemes for outliers measurements removal.
Stoyanov\etal\cite{stoyanov:2011:DEPTHACCURACY,Stoyanov:2013:CER}
compare the accuracy of two ToF cameras and the \kinsl\ camera to a
precise laser range sensor (aLRF). However their evaluation
methodology does not take into account the different error sources
given by real-time range sensing cameras.  The follow-up work by
Langmann\etal\cite{langmann:2012:DepthComparison} compares a ToF
camera (pmdtec CamCube 41k) with the \kinsl. Lateral resolution of
depth measurements are given using a three dimensional Siemens
star-like shape. The depth linearity is also compared using precise
linear rail. The authors conclude that both cameras have different
drawbacks and advantages and thus are meant to be used for different
applications.  Meister\etal\cite{Meister2012can} discuss the
properties of the 3D data acquired with a \kinsl\ camera and fused
into a consistent 3D Model using the so-called
KinectFusion-pipeline~\cite{newcombe:2011:KINFU} in order to provide
ground truth data for low-level image processing. The ``targetbox''
scene used by Meister\etal\cite{Meister2012can}, also called ``HCI
Box'', consists of several object arranged in a $1\times1\times0.5$m
box. Nair\etal\cite{nair13ground-truth} discuss quality measures for
good ground truth data as well as measurement and simulation
approaches to generate this kind of data. We generally opted against
this kind of ground truth scenery, as this approach does often not allow a
proper separation of the individual error sources and, thus, it would
be nearly impossible to transfer results to another application
scenario.

In their book about ToF cameras Hansard\etal\cite{hansard13time}
compare between ToF cameras and the \kinsl. Their comparison focuses
on different material classes. They use $13$ diffuse (``class A''),
$11$ specular (``class B'') and $12$ translucent (``class C'') objects
or object variants for which they acquire geometric ground truth using
an additional 3D scanner and applying white matte spray on each object
surface. As result, they provide root mean square error (RMSE) and
standard deviation (SD).

Compared to all prior work, in this paper we focus on a set of
experimental setups handling an as complete as possible list of
characteristic sensor effects and evaluate these effects for the
\kinsl\ and the \kintof\ cameras presented in
Sec.~\ref{sec:principle}. Before presenting the experiments and
results, we discuss the fundamental problem raised by any attempt to
compare these devices in Sec.~\ref{sec:comparison}. In
Sec.~\ref{sec:results} we present our experiments, that are all
designed in such a way that individual sensor effects can be captured as
isolatedly as possible and that the experiments are reproducible for
other range sensing cameras.

\section{Devices Principle}
\label{sec:principle}

\subsection{Structured Light Cameras - \kinsl}
\label{sec:principle.sl}
Even though the principle of structured light (SL)
range sensing is comparatively old, the launch of the Microsoft
Kinect\texttrademark\ (\kinsl) in 2010 as interaction device for the
XBox~360 clearly demonstrates the maturity of the underlying
principle.

\paragraph*{Technical Foundations}

\begin{figure*}[t!]
  \centering
  \includegraphics[width=.8\textwidth]{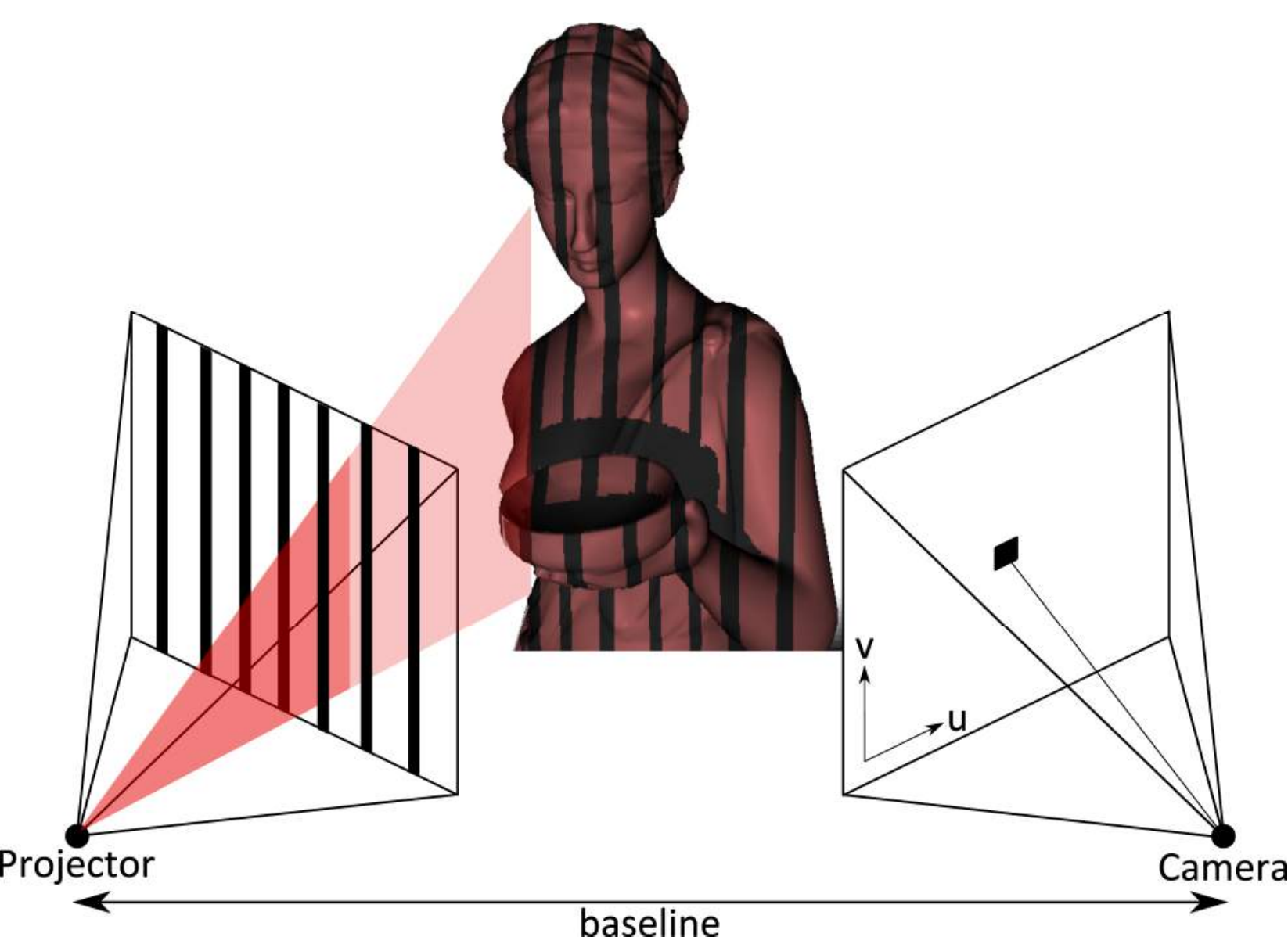} 
  \caption{Principle of structured light based systems.}
  \label{fig:kinsl-scheme}
\end{figure*}

In the structured light approach is an active stereo-vision technique.
A sequence of known patterns is sequentially projected onto an
object, which gets deformed by geometric shape of the object. The
object is then observed from a camera from a different direction. By
analyzing the distortion of the observed pattern, i.e. the disparity
from the original projected pattern, depth information can be
extracted; see Fig.~\ref{fig:kinsl-scheme}.

Knowing the intrinsic parameters of the camera, i.e. the \emph{focal
  length} $f$ and additionally the \emph{baseline} $b$ between the
observing camera and the projector, the depth of pixel $(x,y)$ can be
computed using the disparity value $m(x,y)$ for this pixel as $d =
\frac{b\cdot f}{m(x,y)}$. As the disparity $m(x,y)$ is usually given
in pixel-units, the focal length is also converted to pixel units,
i.e. $f = \frac{f_x^{\text{metric}}}{s_{\text{px}}}$, where
$s_{\text{px}}$ denotes the pixel size. In most cases, the camera and
the projector are only horizontally displaced, thus the disparity
values are all given as horizontal distances. In this case
$s_{\text{px}}$ resembles the horizontal pixel size. The depth range
and the depth accuracy relate to the baseline, i.e. longer baselines
allow for robust depth measurements at long distances.

There are different options to design the projection patterns for a SL range sensor. Several approaches were proposed based on the SL principle in order to
estimate the disparity resulting from the deformation of the projected
light patterns. In the simplest case the
stripe-pattern sequence realizes a binary code which is used to decode
the direction from an object point is illuminated by the beamer. Based
on this principle, Hall-Holt and Rusinkiewicz~\cite{hallholtSBC01}
introduced a real-time camera based 3D system. The authors show that
they could achieve full 3D reconstruction of objects using an
automatic registration of different rotated range maps.

Zhang\etal\cite{zhangRSA02} investigates the benefit of
projection patterns composed of alternative color stripes creating
color transitions that are matched with observed edges. Their matching
algorithm is faster and eliminates the global smoothness assumptions
from the standard SL matching algorithm. Similarly,
Fechteler\etal\cite{fechtelerFAH07} uses this color pattern to
reconstruct at high-resolution human face using only two sequential
patterns, which leads to a reduced computational complexity.

Additionally, Zhang and Huang~\cite{zhangHRR04} proposes an high resolution SL camera based on the use of color fringes pattern and phase-shifting techniques.  Their system was designed to capture and reconstruct at high frame rate (up to 40Hz) dynamic
deformable objects such as human face. 

SL cameras, such as the \kinsl, use a low number of patterns, maybe
only one, to obtain a depth estimation of the scenery at a ``high''
frame rate ($30$~FPS). Typically, it is composed of an near infra-red
(NIR) laser projector combined with a monochrome CMOS camera which
captures depth variations of object surfaces in the scene.

\begin{figure*}[t!]
  \centering
  \includegraphics[width=\textwidth]{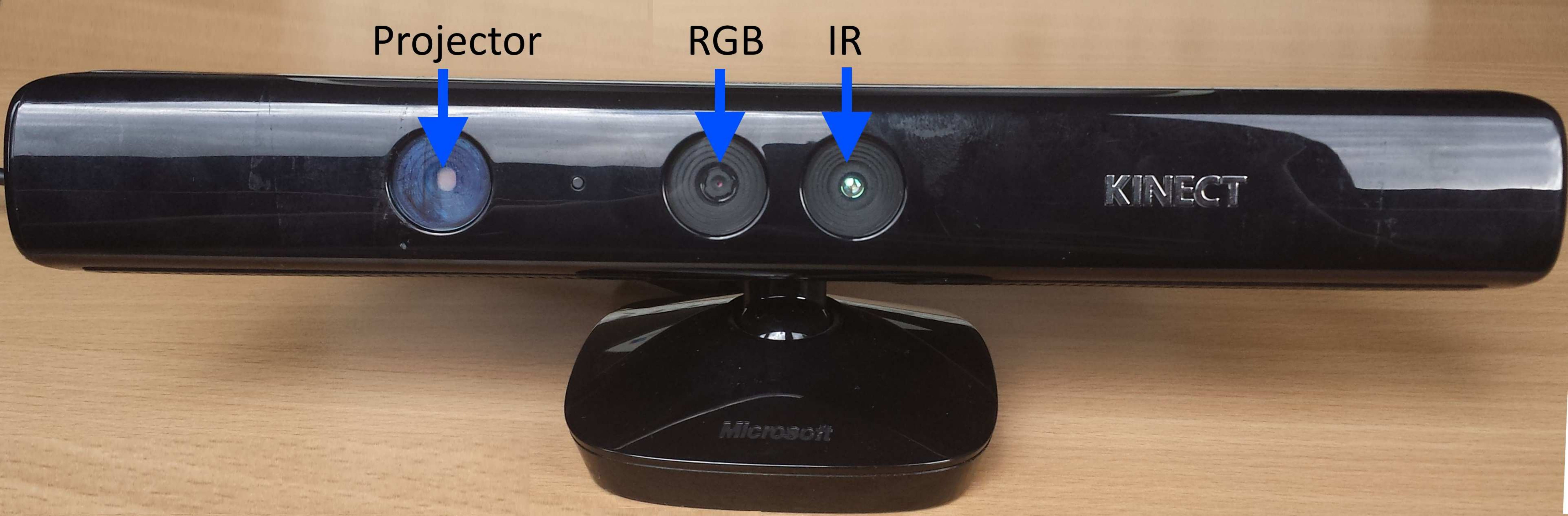} 
  \caption{Sensor placement within a \kinsl\ camera.  The baseline is
    of approximately $7.5$cm.}
  \label{fig:kinsl-principle}
\end{figure*}

The \kinsl\ camera is based on the standard structured light principle
where the device is composed of two cameras, i.e. a color RGB and a
monochrome NIR camera, and an NIR projector including a laser diode at
$850$nm wavelength. The baseline between the
NIR projector and the NIR camera is $7.5$cm see
Fig.~\ref{fig:kinsl-principle}. The NIR projector uses a known and
fixed dot pattern to illuminate the scenery. 

Simple triangulation techniques are later on used to compute the depth
information between the projected pattern seen by the NIR camera and
the input pattern stored on the unit. For each pixel $p_i$, depth is
estimated by finding the best correlation pattern patch, typically in
a $9 \times 9$ pixel window, on the NIR image with the corresponding
projection pattern. The disparity value is given by this best
match. Note that the \kinsl\ device performs internally an
interpolation of the best match operation in order to achieve
sub-pixel accuracy of $\frac{1}{8}$ pixel.  A detailed description of
the Kinect disparity map computation can be found at the ROS.org
community website~\cite{Konolige12openKinectSL}, where the \kinsl's
disparity map computation has been reverse engineered and a complete
calibration procedure is deduced.

\subsection{Time-of-Flight (ToF) Cameras}
\label{sec:principle.tof}

The ToF technology is based on measuring the time that light emitted
by an illumination unit requires to travel to an object and back to
the sensor array~\cite{lefloch13foundation}. In the last decade, this
principle has found realization in microelectronic devices,
i.e. chips, resulting in new range-sensing devices, the so-called
\emph{ToF cameras}. Here, we will explain the basic principle of
operation of ToF-cameras. It should be noted that for the specific
device of the new \kintof\ camera, issued by Microsoft Corp. in
conjunction with the XBox~360 game console, only little technical
detail is known.

\begin{figure*}[t!]
  \centering
  \scalebox{0.75}{\input{images/pmd-principle.tex}}
  \caption{The ToF phase-measurement principle.}
  \label{fig:principle.tof}
\end{figure*}
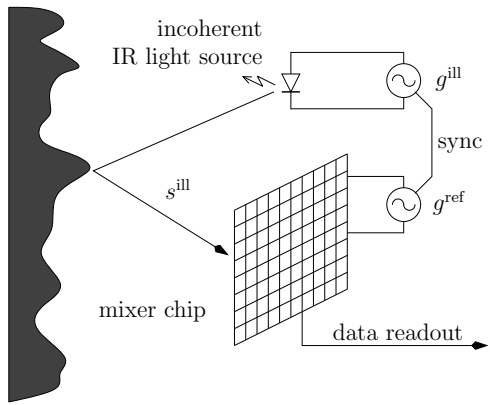

The \kintof\ utilizes the \emph{Continuous Wave (CW) Intensity
  Modulation} approach, which is most commonly used in ToF
cameras. The general idea is to actively illuminate the scene under
observation using near infrared (NIR) intensity-modulated, periodic
light (see Figure~\ref{fig:principle.tof}). Due to the distance
between the camera and the object (sensor and illumination are assumed
to be at the same location), and the finite speed of light $c$, a
\emph{time shift} $\phi [s]$ is caused in the optical signal which is
equivalent to a \emph{phase shift} in the periodic signal. This shift
is detected in each sensor pixel by a so-called \emph{mixing}
process. The time shift can be easily transformed into the
sensor-object distance as the light has to travel the distance twice,
i.e.  $ d = \frac{c\phi}{4\pi}$. 

From the technical perspective, the generator signal $g^{\text{ill}}$
driving the illumination unit results in the intensity modulated
signal which, after being reflected by the scene, results in an incident
optical signal $s^{\text{ill}}$ on each sensor pixel. Note, that the
optical signal may be deformed by nonlinear effects e.g. in the LEDs
of the illumination unit. The incident signal $s^{\text{ill}}$ is
correlated with the reference generator signal $g^{\text{ref}}$. This
mixing approach yields the correlation function which is sampled in
each pixel
$$
C[g^{\text{ill}},g^{\text{ref}}] = s\otimes g = \lim_{T\rightarrow\infty}\int_{-T/2}^{T/2} s^{\text{ill}}(t)\cdot
g^{\text{ref}}(t)\;\mathrm{d}t.
$$  
The phase shift is computed using several correlation measurements
with varying illumination and reference signals $g^{\text{ill}}_i$ and
$g^{\text{ref}}_i$, respectively, using some kind of demodulation
function, i.e.
$$
\phi =
{\cal
  G}(A_0,A_1,\ldots,A_n),\quad\text{with }A_i=C[g^{\text{ill}}_i,g^{\text{ref}}_i],\.i=1,\ldots,n.
$$ 
Frequently, $A_i$ is called \emph{phase image} or \emph{correlation
  image}. We will use the latter notation in order to prevent
confusion with the phase shift ($\propto$~distance). Practically, the
correlation images are acquired sequentially, however there is the
theoretic option to acquire all correlation images in parallel,
e.g. by having different phase shifts for neighboring pixels. Note, that due to the periodicity of the reference signal, any ToF-camera has a unique unambiguous measurement range.

The first ToF cameras like the prototypes from
pmdtechnolgies~\cite{xuSPP99} used sinusoidal signals
$g^{\text{ill}}(t)=\cos(2\pi f_m t)$ with a constant modulation
frequency $f_m$ and a reference signal equal to $g^{\text{ill}}$ with
an additional phase offset $\tau$,
i.e. $g^{\text{ref}}(t)=g^{\text{ill}}(t+\tau)$. For this approach,
usually four correlation images $A_i=C[\cos(2\pi f_m\,\cdot),\cos(2\pi
  f_m\,\cdot+\tau_i)]$ for $\tau_i=i\cdot\sfrac\pi2,\;i=0,1,2,3$ are
acquired leading to a distance value of
$$
\phi = {\cal G}(A_0,A_1,A_2,A_3) = \arctann(A_3-A_1, A_0-A_2)/f_m,
$$ 
where $\arctann(y, x)$ is the angle between the positive x-axis and the point
given by the coordinates $(x,y)$.

The \kintof\ camera applies this CW intensity modulation
approach~\cite{xu2013method}. Blake\etal\cite{blake15openKinectToF}
reverse engineered the \kintof-driver. This revealed, that the
\kintof\ acquires $10$ correlation images, from which nine correlation
images are used for a three-phase reconstruction approach based on
phase shifts of $0^\circ, 120^\circ$ and $240^\circ$ at three
different frequencies. Using multiple modulation frequencies the measurement range can be exceeded~\cite{droeschel2010multi} Although the \kintof\ camera can obtain depth values for distances longer than 9 meters, the official driver masks the distances further than around 4.5 meters.

The purpose of the tenth correlation image is
still not clear.  Even though the technical specifics of the
\kintof\ have not been explicitly revealed by Microsoft, it definitely
applies the basic principle of correlation as described above. The
illumination unit consists of a laser diode at $850$nm wavelength.

In
Sec.~\ref{sec:comparison} we discuss further technical details
regarding the \kintof\ driver.

\subsection{Error Sources for \kinsl\ and \kintof}
\label{sec:principle.err}

SL and ToF cameras are active imaging systems that use standard optics
to focus the reflected light onto the chip area. Therefore, the
typical optical effects like shifted optical centers and lateral
distortion need to be corrected, which can be done using classical
intrinsic camera calibration techniques. Beyond this camera specific
calibration issues, SL and ToF cameras possess several  specific
error sources, which are discussed in the following and which also
apply to \kintof\ and/or \kinsl. As a detailed discussion of prior work
in relation to these error sources would go beyond the scope of this
paper, we only give some relevant links to prior work that relates to
the individual effects for either system.

{\bfseries Ambient Background Light [SL, ToF]:} As any other camera,
ToF and SL cameras can suffer from ambient background light, as it can
either lead to over-saturation in case of too long exposure times in
relation to the objects' distance and/or reflectivity, e.g. causing
problems to SL-systems in detecting the light pattern. Both, the
\kintof\ and the \kinsl\ are utilized with a band-pass filter,
suppressing background light out of the range of the
illumination. \kintof\ provides a suppression of background intensity
on the chip.

For ToF cameras specific circuitry has been developed, e.g. the
\emph{Suppression of Background Intensity} approach for PMD
cameras~\cite{ringbeck2007multidimensional} that electronically filter
out the DC-part of the light. For SL systems outdoor application is
usually hard to achieve, which has also been stated for the
\kinsl~\cite{el2012study}.

{\bfseries Multi-Device Interference [SL, ToF]:} Similar to any other
active sensing approach, the parallel use of several Kinect cameras
may lead to interference problems, i.e.\ the active illumination of
one camera influences the result of another camera.

For \kinsl\ the potential interference problem given by multiple NIR
patterns projected into the scene is very difficult to
solve. Butler\etal\cite{butlerSNS12} propose a ``Shake`n'Sense'' setup
where one (or each) \kinsl-device is continuously shaken using an
imbalanced rotating motor. Thus, the projected pattern performs a high
frequency motion that appears significantly blurred for another
device.  An alternative approach is introduced by
Berger\etal\cite{bergerMMC11}. They add steerable hardware shutters to
the \kinsl-devices' illumination units resulting in a time-multiplex
approach. For ToF cameras the signal shape can be altered in order to
prevent multi-device interference, e.g.  for sinusoidal signal shapes different modulation frequencies can simply be used to decouple the
devices~\cite{kim2009multi}.

{\bfseries Temperature Drift [SL, ToF]:} A common effect to many
technical devices is the drift of the system output, i.e. the distance
values in the case of Kinect cameras, during the device warm-up. The
major difference between the SL and the ToF approach is that an SL
camera usually does not produce as much heat as a ToF camera. This is
due to the fact, that the required illumination power to cover the
full scene width and depth in order to get a sufficient
\emph{signal-to-noise} (SNR) for the optical signal for a ToF camera
is beyond the power needed to generate the relatively sparse
point-based pattern applied by the \kinsl. As a consequence, the
\kinsl\ can be cooled passively whereas the \kintof\ requires active
cooling.

For the \kinsl\ significant temperature drift has been reported by
Fiedler and M\"uller~\cite{fiedler2013impact}.  Early ToF-camera
studies e.g. from Kahlmann\etal\cite{kahlmannCFI06} of the
Swissranger\texttrademark\ camera exhibit the clear impact of this
warm-up on the range measurement. More recently smaller ToF cameras
for close range applications such as the camboard-nano series provided
my pmdtechnolgies do not require active cooling, however, no
temperature drift investigations have been reported so far.

\begin{figure*}[t!]
  \centering
  \includegraphics[height=.2\textheight]{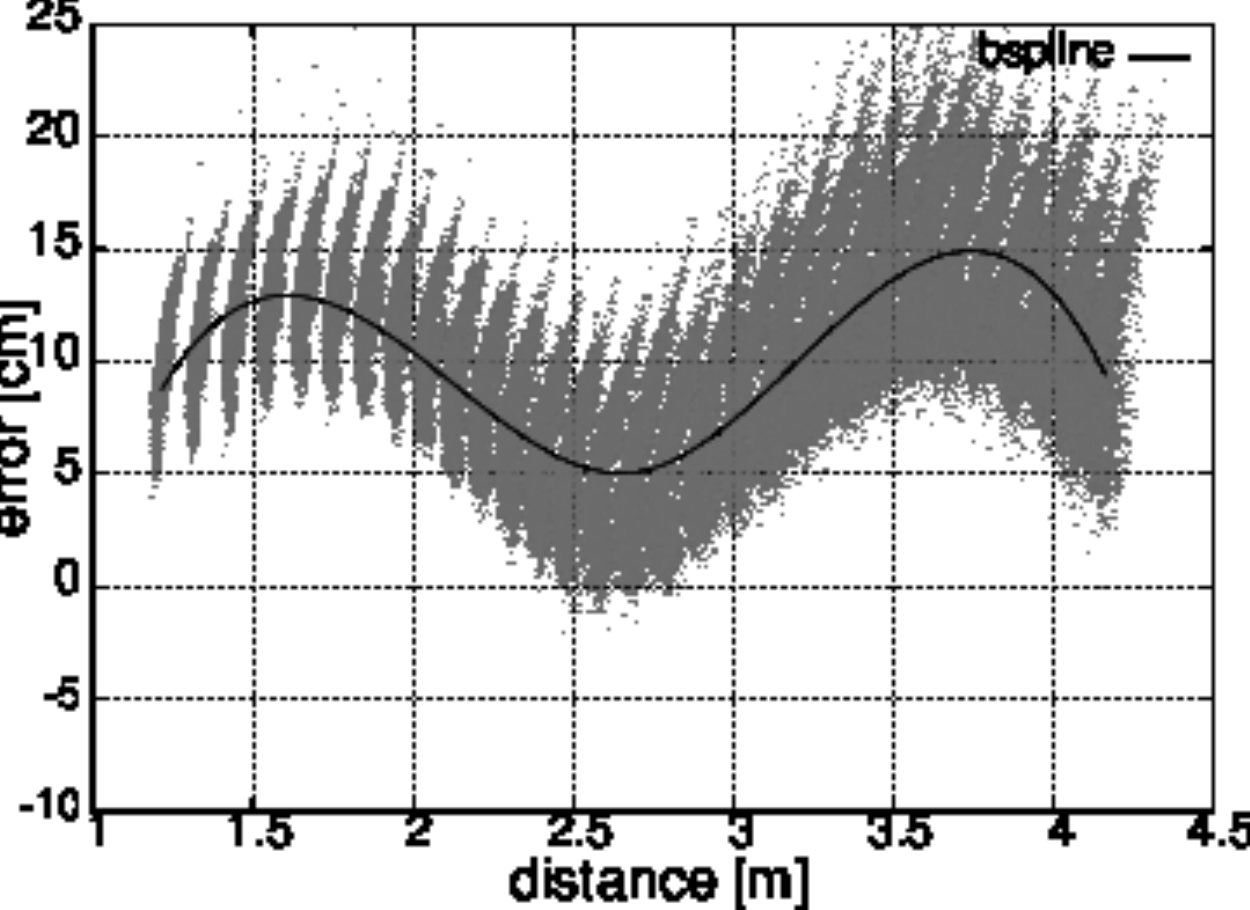} \;
  \includegraphics[height=.2\textheight]{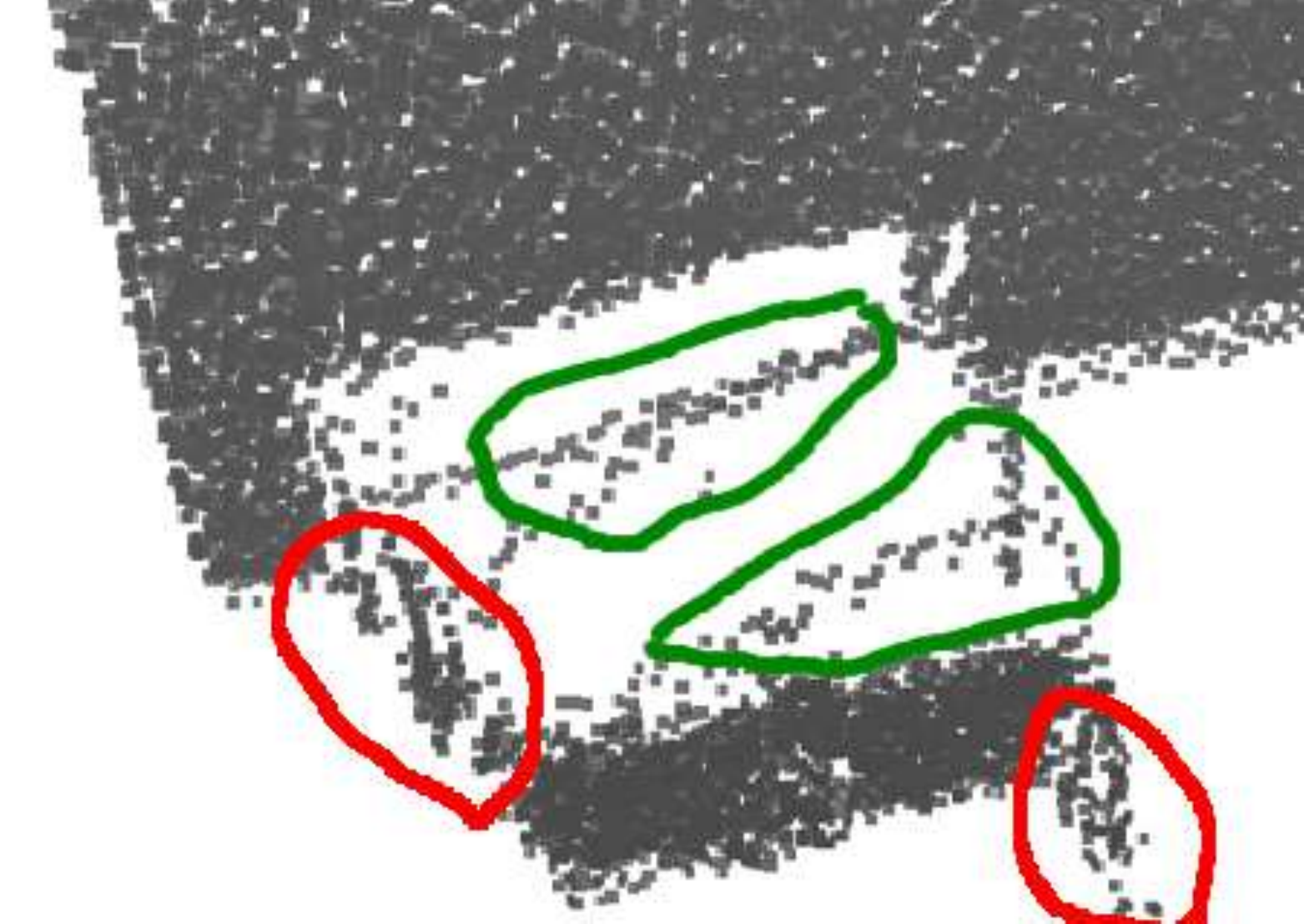} \\
  \includegraphics[height=.2\textheight,width=.5\textwidth]{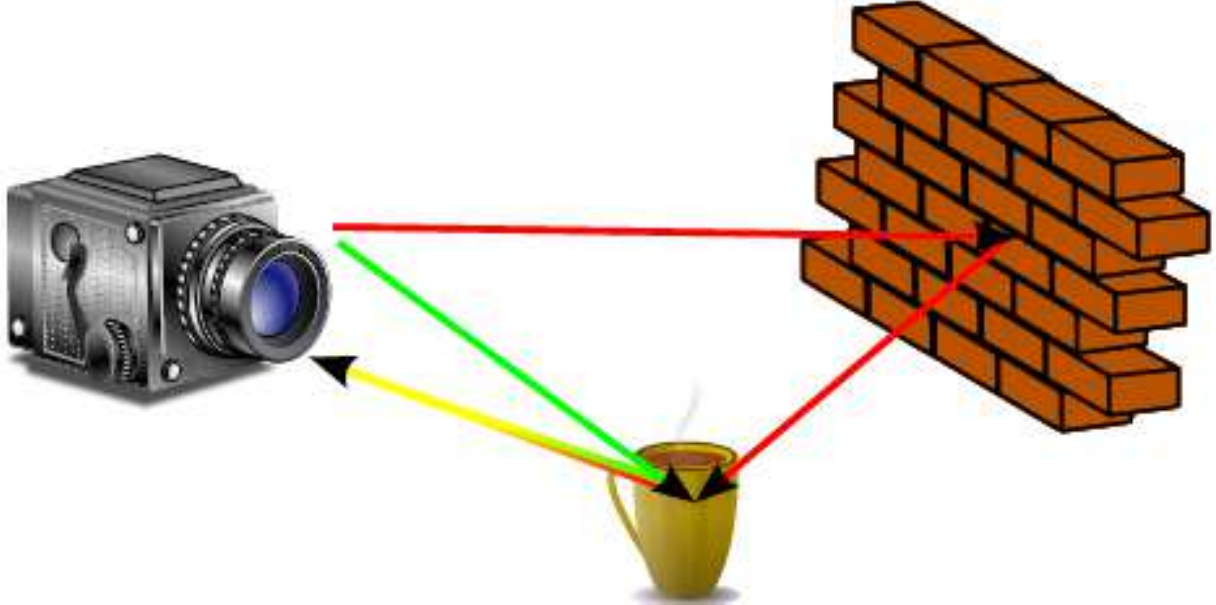} \;
  \includegraphics[height=.2\textheight]{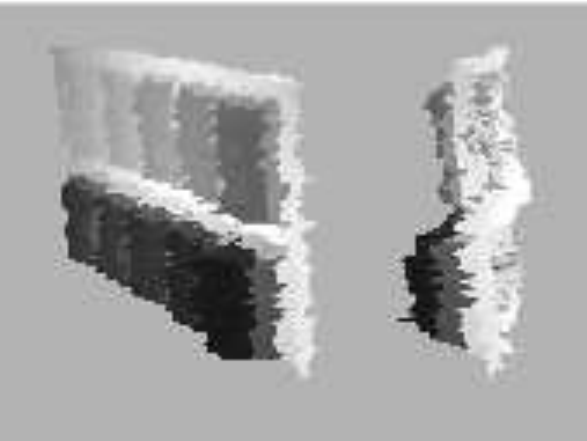} \;
  \caption{Error sources of ToF cameras. Top left: Systematic
    (wiggling) error for all pixels (gray) and fitted mean deviation
    (black). Top right: Motion artifacts (red) and flying pixels
    (green) for a horizontally moving planar object in front of a
    wall. Bottom left: Schematic illustration of multi-path effects
    due to reflections in the scene. Bottom right: Acquisition of a planar
    gray-scale checkerboard reveals the intensity related distance
    error. (Image courtesy: \cite{kolbTOF10}, Eurographics Association, 2010.)}
  \label{fig:foundation.tof-error}
\end{figure*}

{\bfseries Systematic Distance Error [SL, ToF]:} Both Kinect cameras
suffer from systematic error in their depth measurement. For the
\kinsl\ the error is mainly due to inadequate calibration and
restricted pixel resolution for estimation of the point locations in the
image plane, leading to imprecise pixel coordinates of the reflected
points of the light pattern~\cite{khoshelhamAAR12}. Further range
deviations for the \kinsl\ result from the comparably coarse
quantization of the depth values which increases for further distances
from the camera. For \kintof, on the other hand, the distance
calculation based on the mixing of different optical signals $s$ with
reference signals $g^{\text{ref}}$ requires either an approximation to
the assumed, e.g. a sinusoidal signal shape or an approximation to the
phase demodulation function $\cal{G}$. Both approximations lead to a
systematic error in the depth measurement. In case of an approximated
sinusoidal shape this effect is also called ``wiggling'' (see
Figure~\ref{fig:foundation.tof-error}, top left).  The systematic error may
depend on other factors, such as the exposure time.

For \kinsl\ Khoshelham and Elberink~\cite{khoshelhamAAR12} present a
detailed analysis of its accuracy and depth resolution. They conclude
that the systematic error is below some $3$cm, however it increases on
the periphery of the range image and for increasing object-camera
distance.  Smisek\etal\cite{smisek3WK11} present a geometric method to
calibrate the systematic error of the
\kinsl. Herreta\etal\cite{herreraJDA12} proposed a joint calibration
approach for the color and the depth camera of the \kinsl.  Correction
schemes applied to reduce the systematic error of ToF cameras with
sinusoidal reference signals simply model the depth deviation using a
look-up-table~\cite{kahlmannRIT07} or function fitting, e.g. using
b-splines~\cite{lindnerLAD06}.

{\bfseries Depth Inhomogeneity [SL, ToF]} At object boundaries, a
pixel may observe inhomogeneous depth values. Due to the structured
light principle, occlusion may happen at object boundaries where parts
of the scene are not illuminated by the infra-red beam which results
in a lack of depth information in those regions (invalid pixels).  For
ToF cameras, the mixing process results in a superimposed signal
caused by light reflected from different depths, so-called \emph{mixed
  pixels}. In the context of ToF cameras these pixels are sometimes
called \emph{flying pixels}. The mixed, or flying signal leads to
wrong distance values; see Figure~\ref{fig:foundation.tof-error}, top
right.

There are simple methods relying on geometric models that give good
results in identifying flying pixel, e.g. by estimating the depth
variance which is extremely high for flying
pixel~\cite{sabovIAC08}. Denoising techniques, such as a median filter,
can be used to correct some of the flying pixels.

Note that flying pixels are directly related to a more general
problem, i.e. the multi-path problem; see below.

{\bfseries Multi-Path Effects [SL, ToF]} Multi-path effects relate to
an error source common to active measurement systems: The active light
may not only travel the direct path from the illumination unit via the
object's surface to the detector, but may additionally travel
\emph{indirect paths}, i.e. being scattered by highly reflective
objects in the scene or within the lens systems or the housing of the
camera itself, see Fig.~\ref{fig:foundation.tof-error} bottom left. In
the context of computer graphics this effect is known as global
illumination. For ToF cameras these multiple responses of the active
light are superimposed in each pixel leading to an altered signal not
resembling the directly reflected signal and thus a wrong
distance. For \kinsl\ indirect illumination mainly causes problems for
highly reflecting surfaces, as dots of the pattern may be projected at
other objects in the scene. However, objects with a flat angle to the
camera will lead to a complete lack of depth information (see also
Sec.~\ref{sec:results:reflect}).

For ToF cameras several correction schemes for multi-path effects have
been proposed for sinusoidal signal shapes. Falie and
Buzuloiu~\cite{falieDEC08} assume that the indirect effects are of
rather low spatial frequency and analyze the pixel's neighborhood to
detect the low-frequency indirect
component. Dorrington\etal\cite{dorringtonSTR11} present an analytic
formulation for the signal superposition resulting in a non-linear
optimization scheme per pixel using different modulation frequencies.

{\bfseries Intensity-Related Distance Error [ToF]} Considering a
highly reflecting object and a second object with the same distance to
the camera but with low reflectivity in the relevant NIR range, a
reduced SNR is expected. Beyond this, it has frequently been reported
that ToF cameras have a non-zero biased distance offset for objects
with low NIR reflectivity (see Figure~\ref{fig:foundation.tof-error}, bottom
right).

Lindner\etal\cite{lindnerTOF10} tackle the specific intensity-related
error using phenomenological approaches. In general, there are at
least two possible explanations for this intensity-related effect. The
first assumption explains this effect is a specific variant of a
multi-path effect, the second one puts this effect down to the
non-linear pixel response for low amounts of incident intensity.

{\bfseries Semitransparent and Scattering Media [SL, ToF]} As for most
active measuring devices, media that does not perfectly reflect the
incident light potentially causes errors for ToF and SL cameras. In
case of ToF cameras, light scattered within semitransparent media
usually leads to an additional phase delay due to a reduced speed of
light.

The investigations done by Hansard\etal\cite{hansard13time} give a
nice overview for specular and translucent, i.e. semitransparent and
scattering media for ToF cameras with sinusoidal reference signal and
the \kinsl. 
Kadambi\etal\cite{kadambi2013coded} show that their coding method (originally designed to solve multi-path errors for ToF cameras) is able to recover depth of near-transparent objects using their resulting time-profile (transient imaging).
Finally, a detailed state-of-the-art report is given by Ihrke\etal\cite{ihrke2010transparent}
where different methods are described in order to robustly acquire and reconstruct such challenging media.

{\bfseries Dynamic Scenery [SL, ToF]} One key assumption for any
camera-based system is that each pixel observes a single object point
during the whole acquisition process. This assumption is violated in
case of moving objects or moving cameras, resulting in motion
artifacts. In real scenes, motion may alter the true depth.  Even
though \kinsl\ acquires depth using only a single NIR image of the
projected pattern, a moving object and/or camera leads to improper
detection of the pattern in the affected region.  ToF cameras as the
\kintof\ require several correlation images per depth image.
Furthermore, their correlation measurements get affected by a change
of reflectivity observed by a pixel during the acquisition. Processing
the acquired correlation images ignoring the motion present during
acquisition leads to erroneous distance values at object boundaries
(see Figure~\ref{fig:foundation.tof-error}, top right).

However, no real investigations have been done yet for the \kinsl\ to
study the effect of motion blur on the depth measurement
quality. Nevertheless the work of Butler\etal\cite{butlerSNS12} uses
the motion blur property to solve the problem of multiple \kinsl\
devices interference.

For ToF cameras several motion compensation schemes have been
proposed. Schmidt and J\"ahne~\cite{schmidtEAR11} detect motion
artifacts using temporal gradients of the correlation images $A_i$,
i.e. a large gradient in one of the correlation images indicates
motion. This approach also performs a correction using extrapolated
information from prior frames; see also discussion in
Hansard\etal\cite {hansard13time}, Sec.~1.3.3.  Since motion artifacts
result from in-plane motion between subsequent correction images,
several approaches use optical flow methods in order to re-align the
individual correlation images. Lindner and Kolb~\cite{lindnerCOM09}
apply a fast optical flow algorithm~\cite{ZachADB07} three times in
order to align the four correlation images $A_0,A_1,A_2,A_3$ to the
first correlation image $A_0$. As optical flow algorithms are
computationally very expensive, these approaches significantly reduce
the frame rates for real-time processing. A faster approach is motion
detection and correction using block-matching techniques applied
pixels where motion has been detected~\cite{hoeggRTM13}.

\begin{figure}
  \centering  \begin{tabular}{cccc}
    & \small Mean & \small SD &  \small RMSE (plane fit)\\
    \begin{rotate}{90}\small\kinsl\ Offic. \end{rotate}&
    \includegraphics[trim = 32mm 86mm 11mm 98mm,clip,width=.3\textwidth]{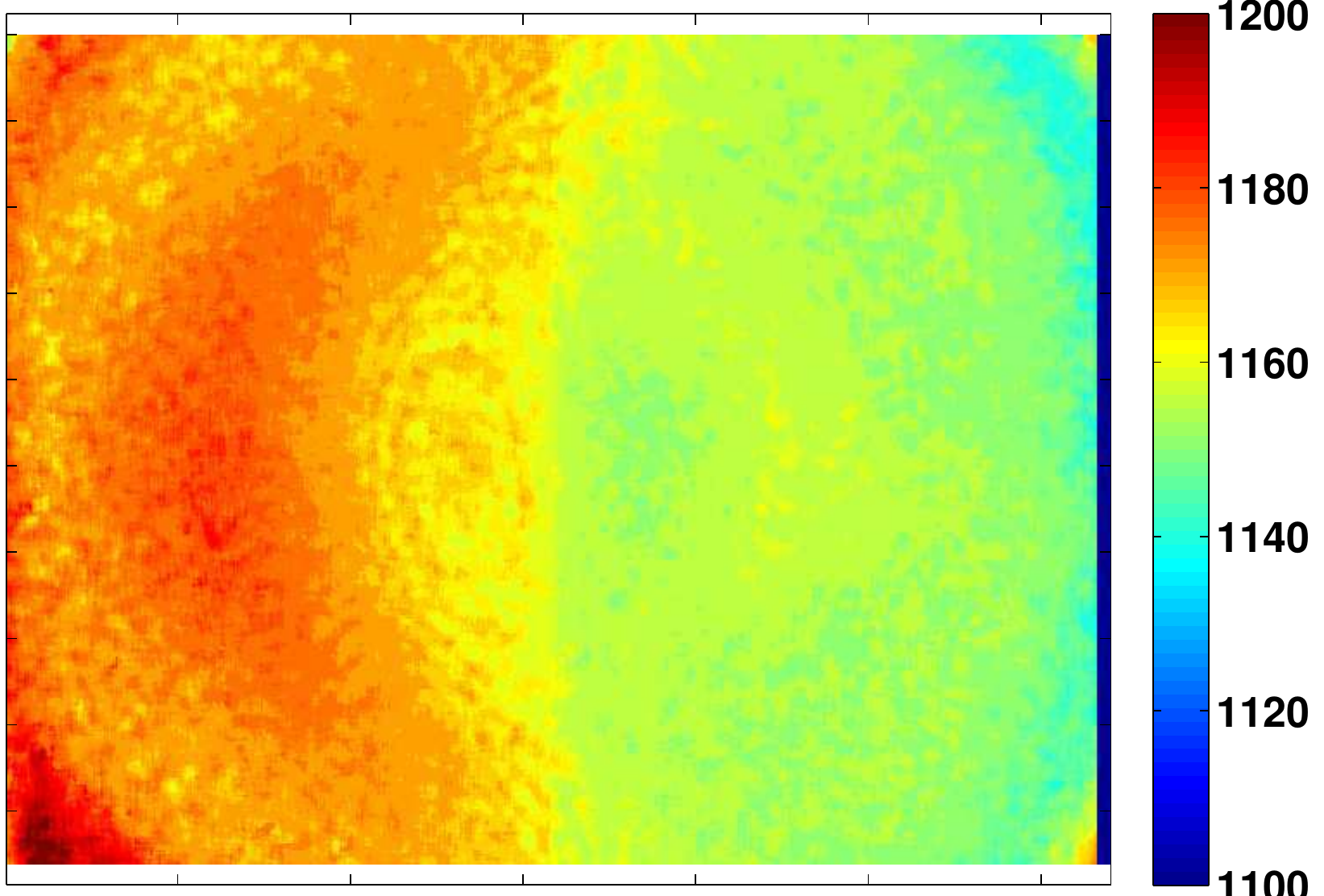} &
    \includegraphics[trim = 32mm 86mm 11mm 98mm,clip,width=.3\textwidth]{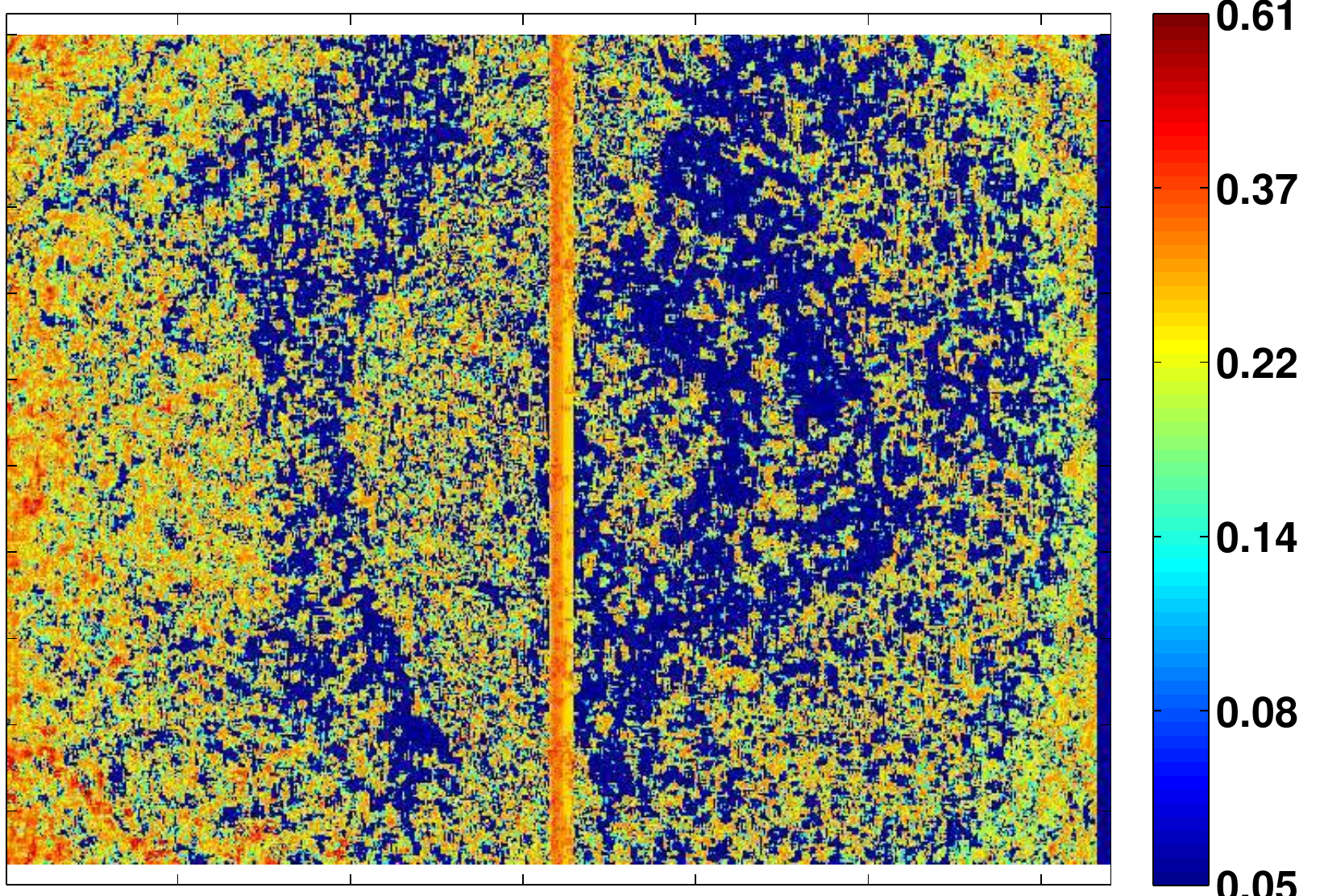} &
    \includegraphics[trim = 32mm 86mm 11mm 98mm,clip,width=.3\textwidth]{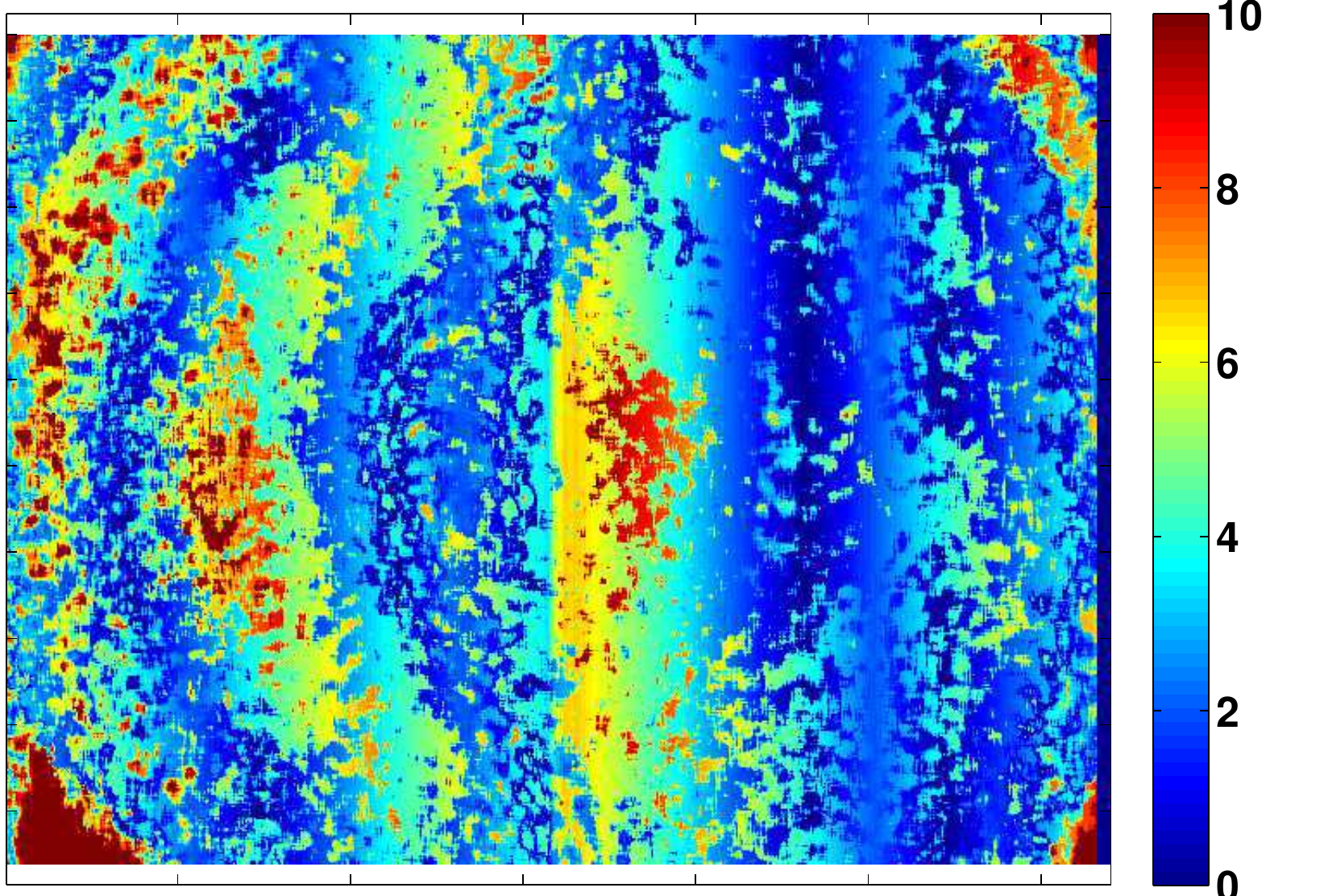}\\
    \begin{rotate}{90}\small\kinsl~Post\end{rotate}&
    \includegraphics[trim = 32mm 86mm 11mm 98mm,clip,width=.3\textwidth]{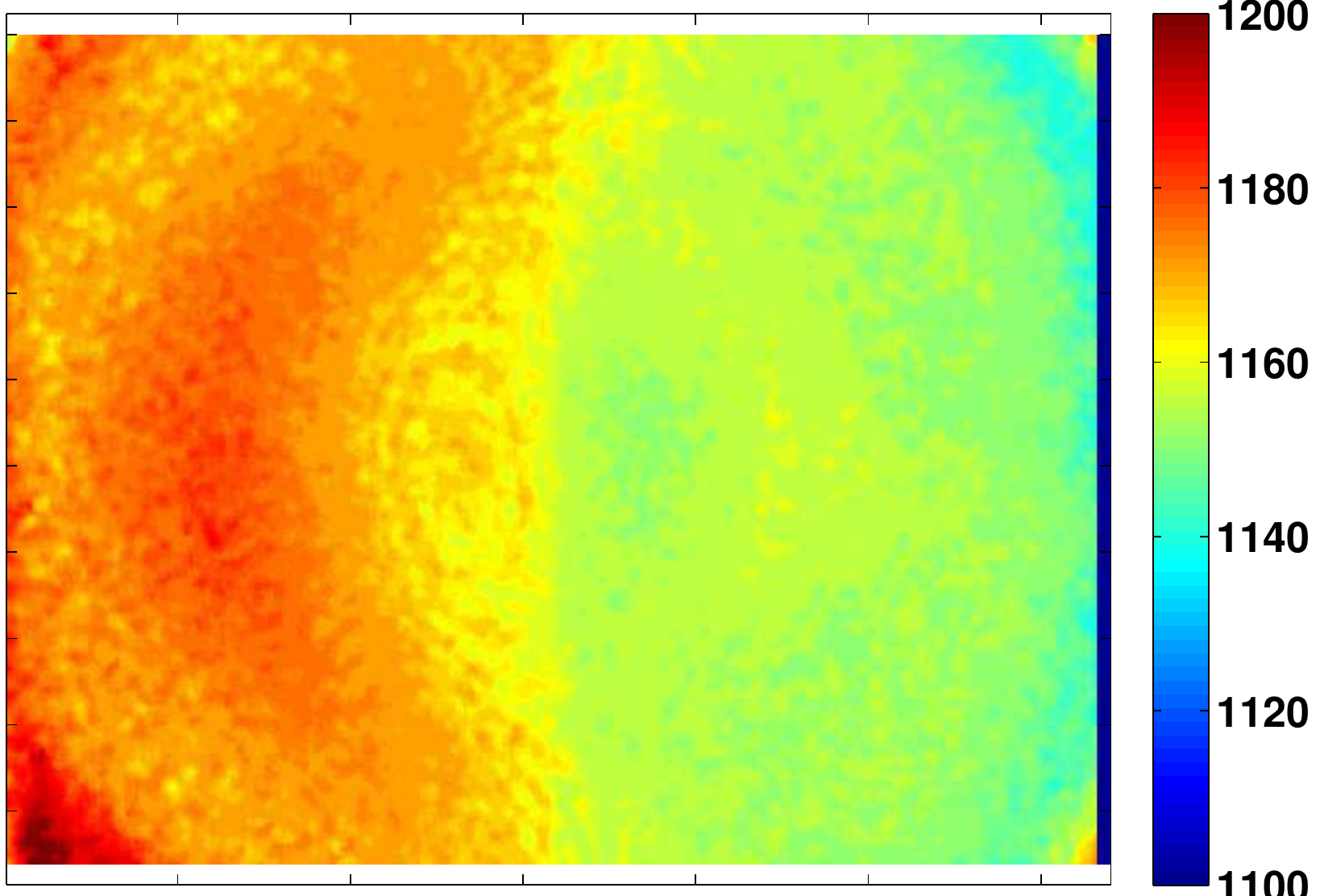} &
    \includegraphics[trim = 32mm 86mm 11mm 98mm,clip,width=.3\textwidth]{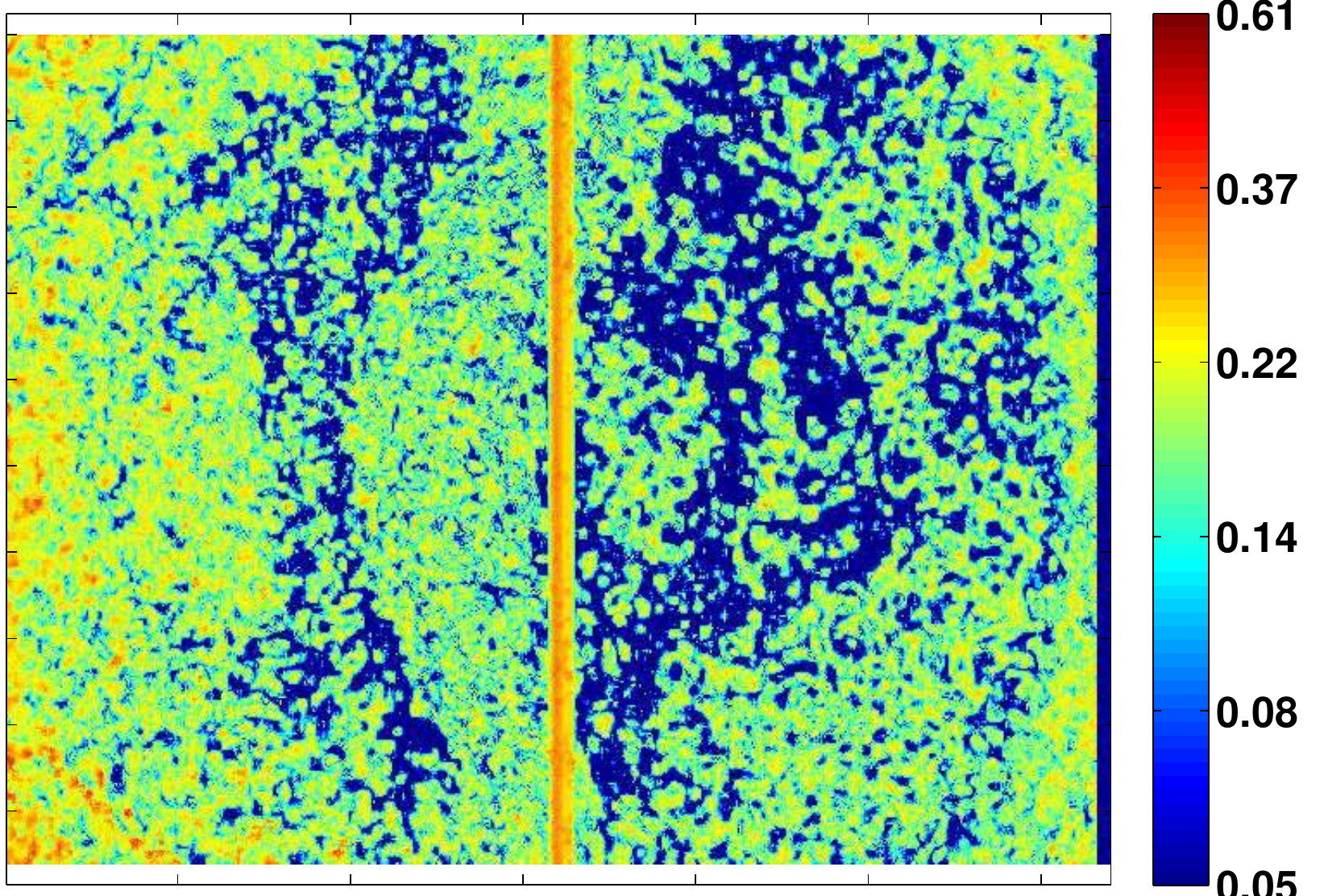} &
    \includegraphics[trim = 32mm 86mm 11mm 98mm,clip,width=.3\textwidth]{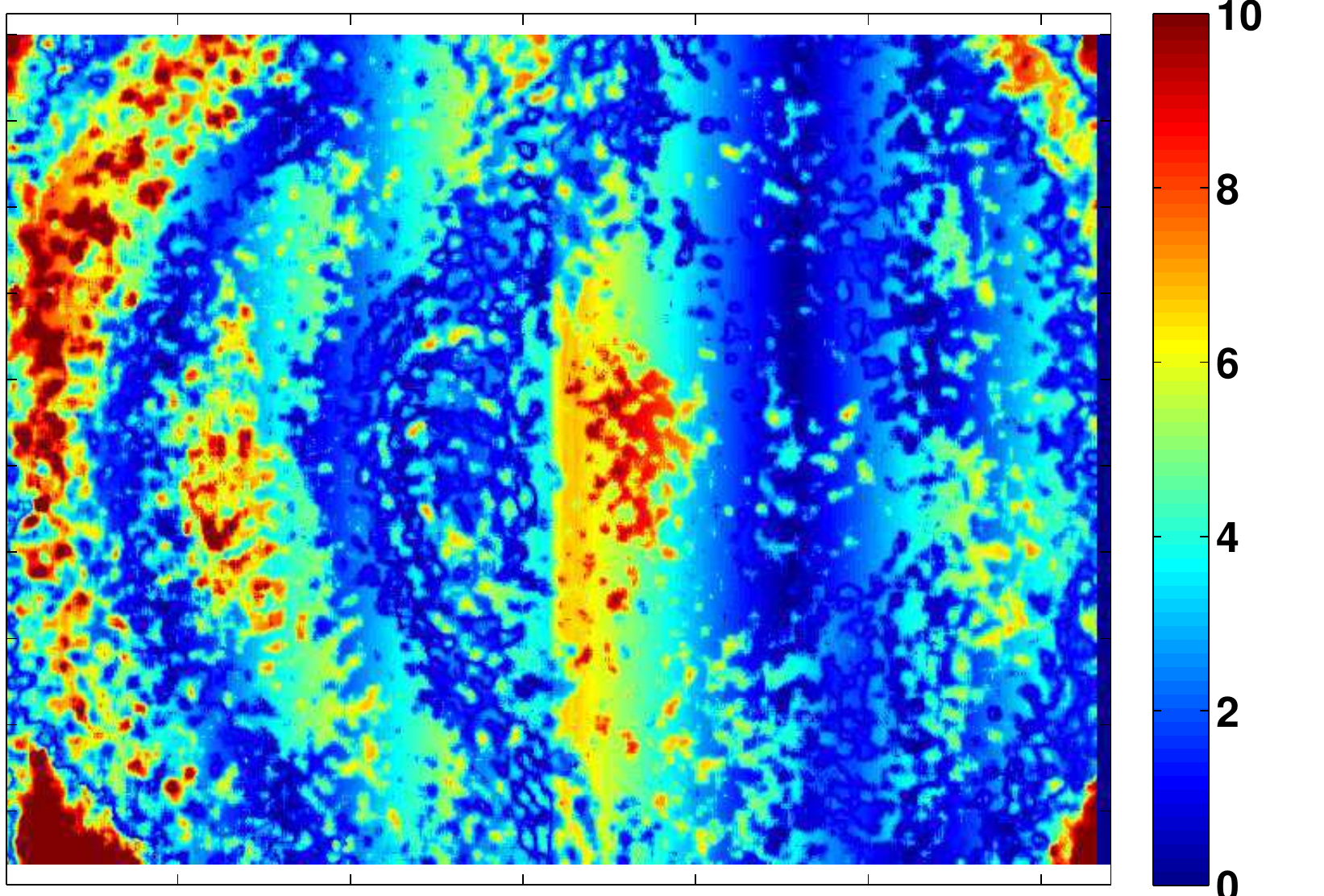}\\
    \begin{rotate}{90}\small\kintof\ Offic. \end{rotate}&
    \includegraphics[trim = 32mm 86mm 11mm 98mm,clip,width=.3\textwidth]{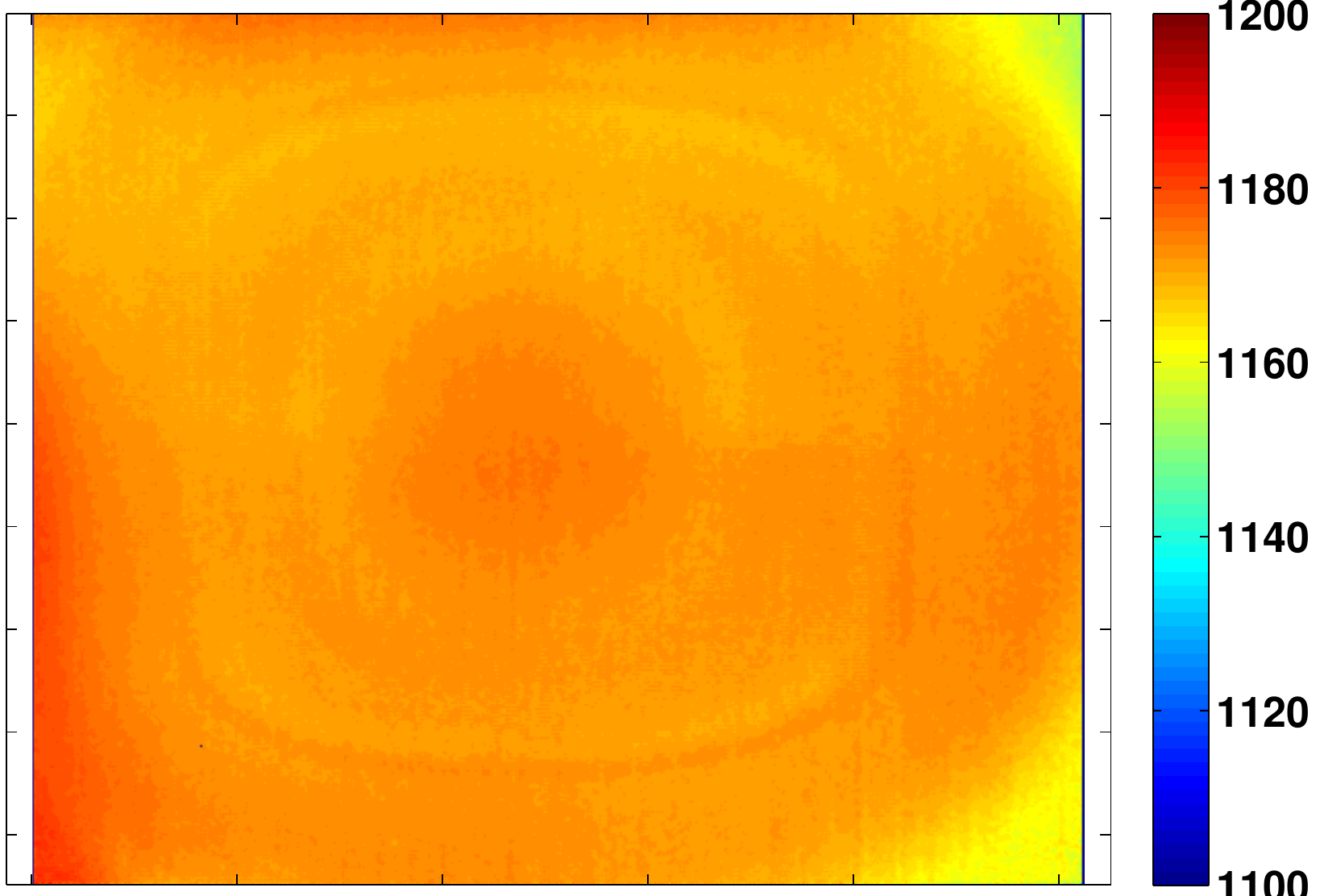} &
    \includegraphics[trim = 32mm 86mm 11mm 98mm,clip,width=.3\textwidth]{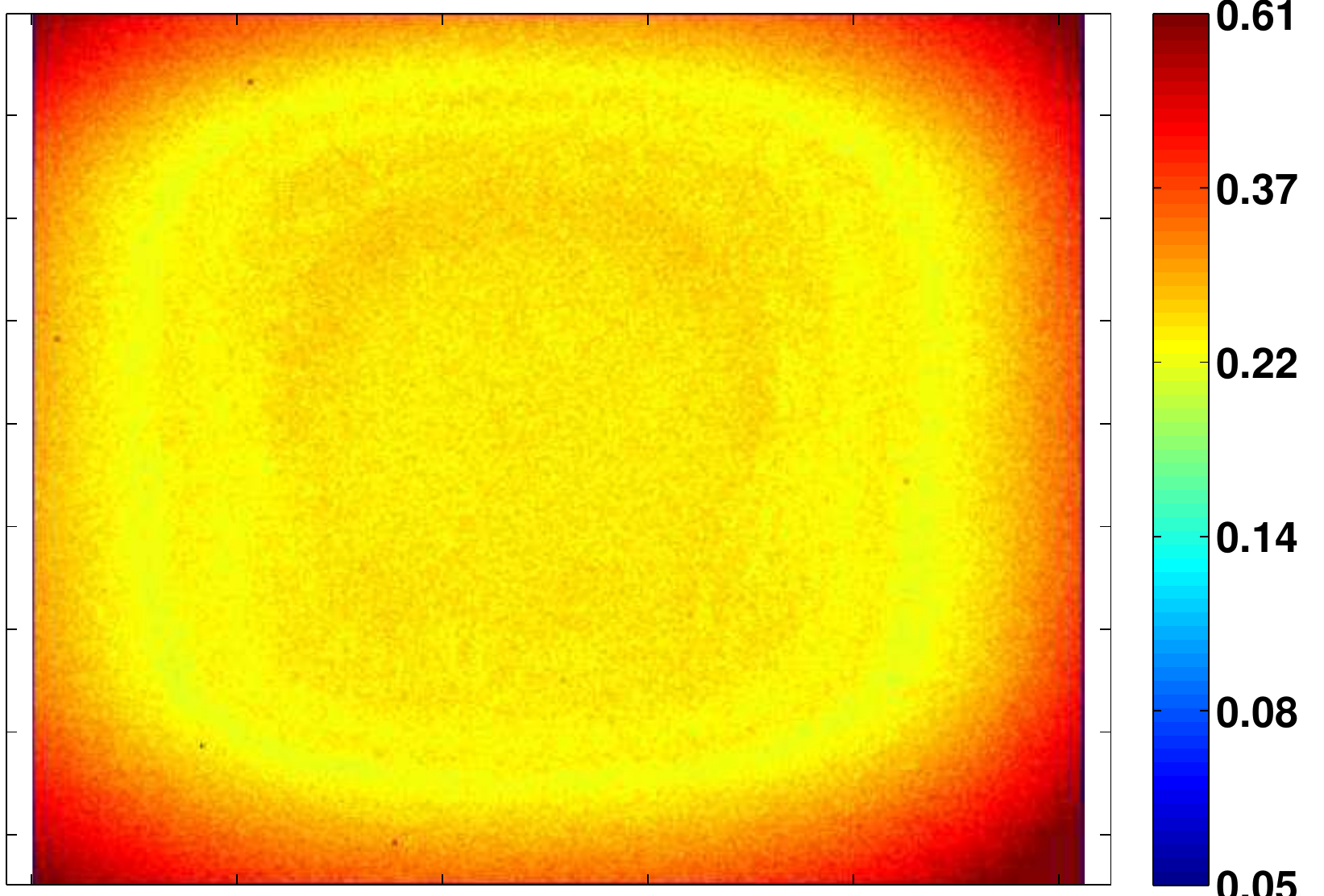} &
    \includegraphics[trim = 32mm 86mm 11mm 98mm,clip,width=.3\textwidth]{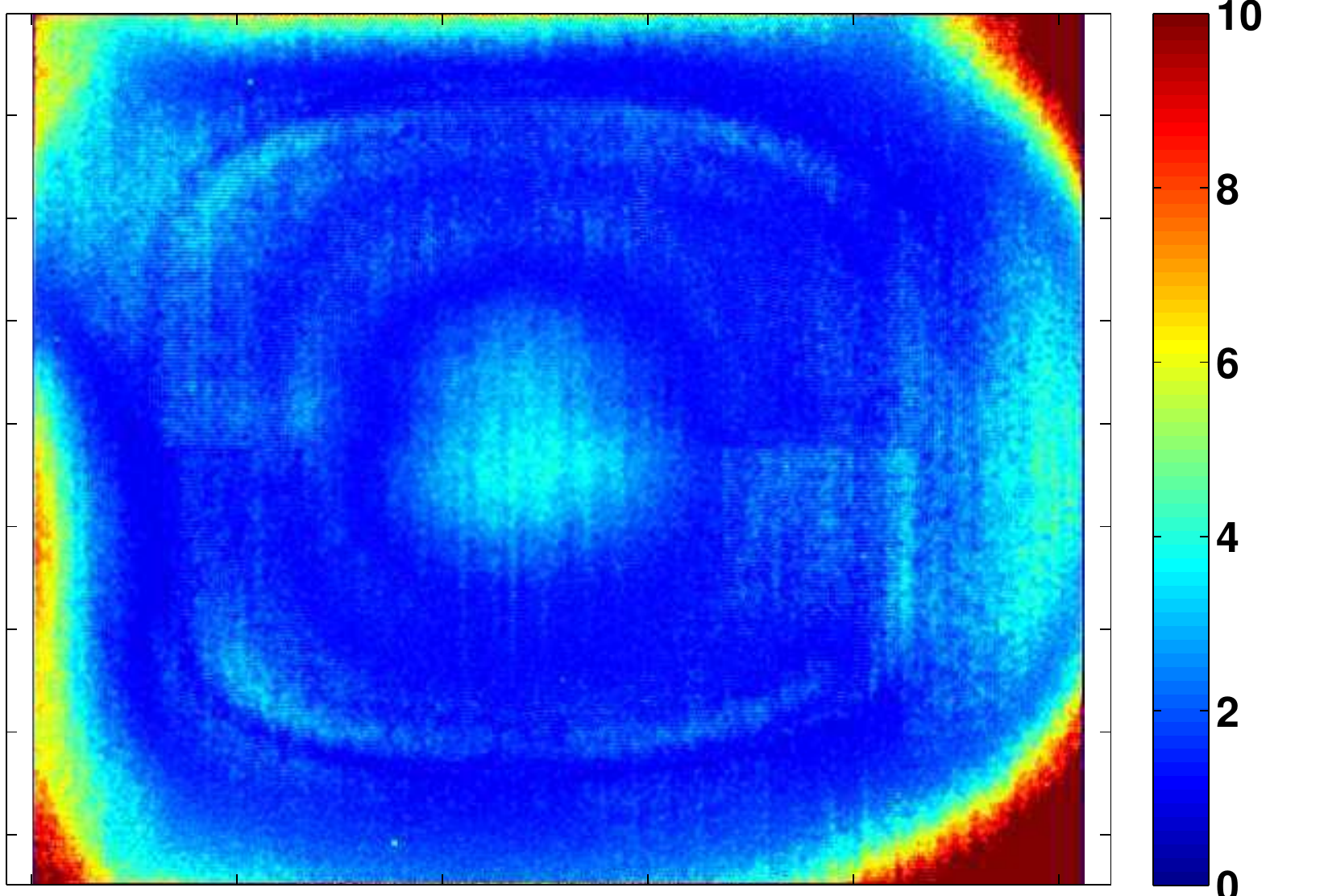}\\
    \begin{rotate}{90}\small\kintof~Open\end{rotate}&
    \includegraphics[trim = 32mm 86mm 11mm 98mm,clip,width=.3\textwidth]{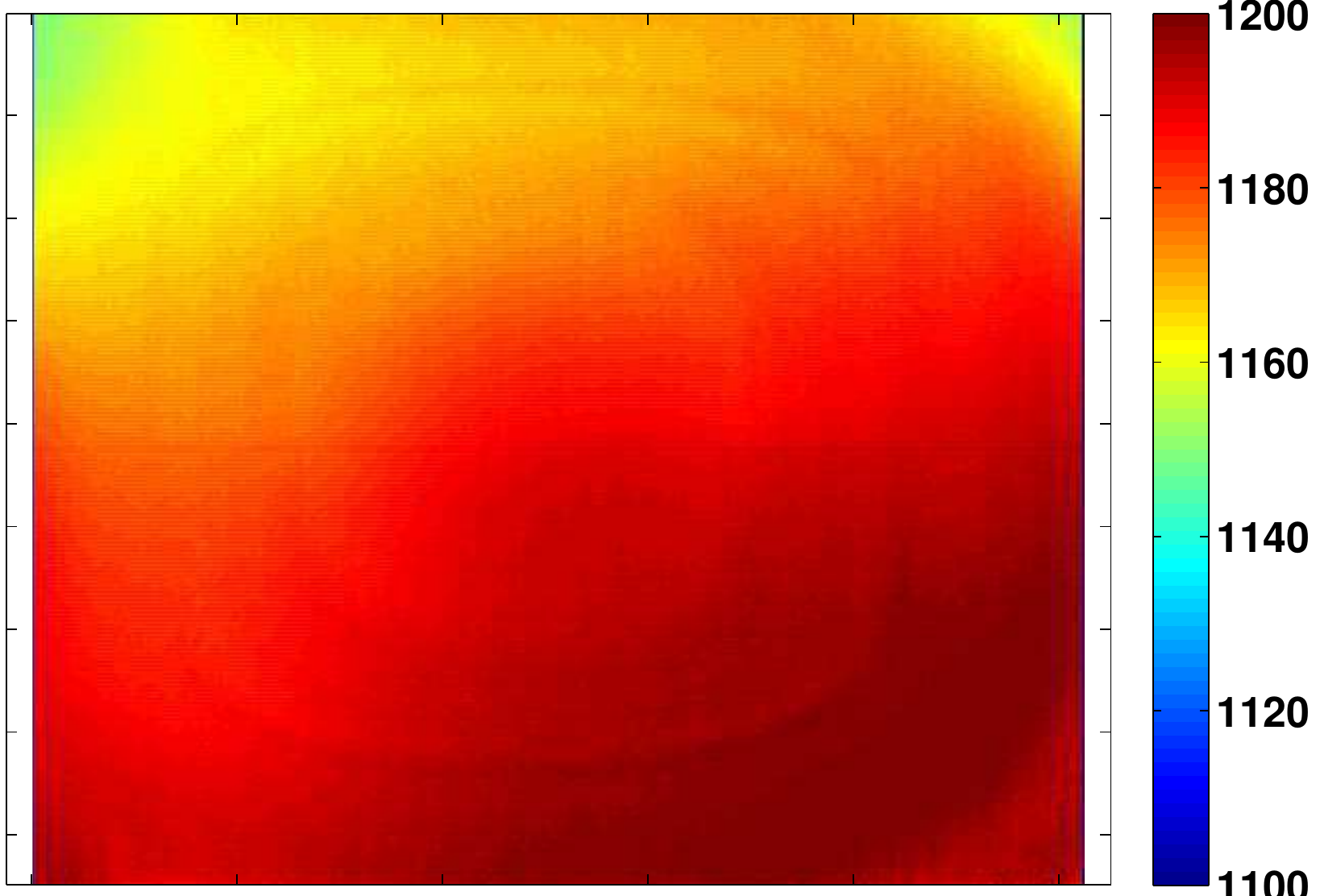} &
    \includegraphics[trim = 32mm 86mm 11mm 98mm,clip,width=.3\textwidth]{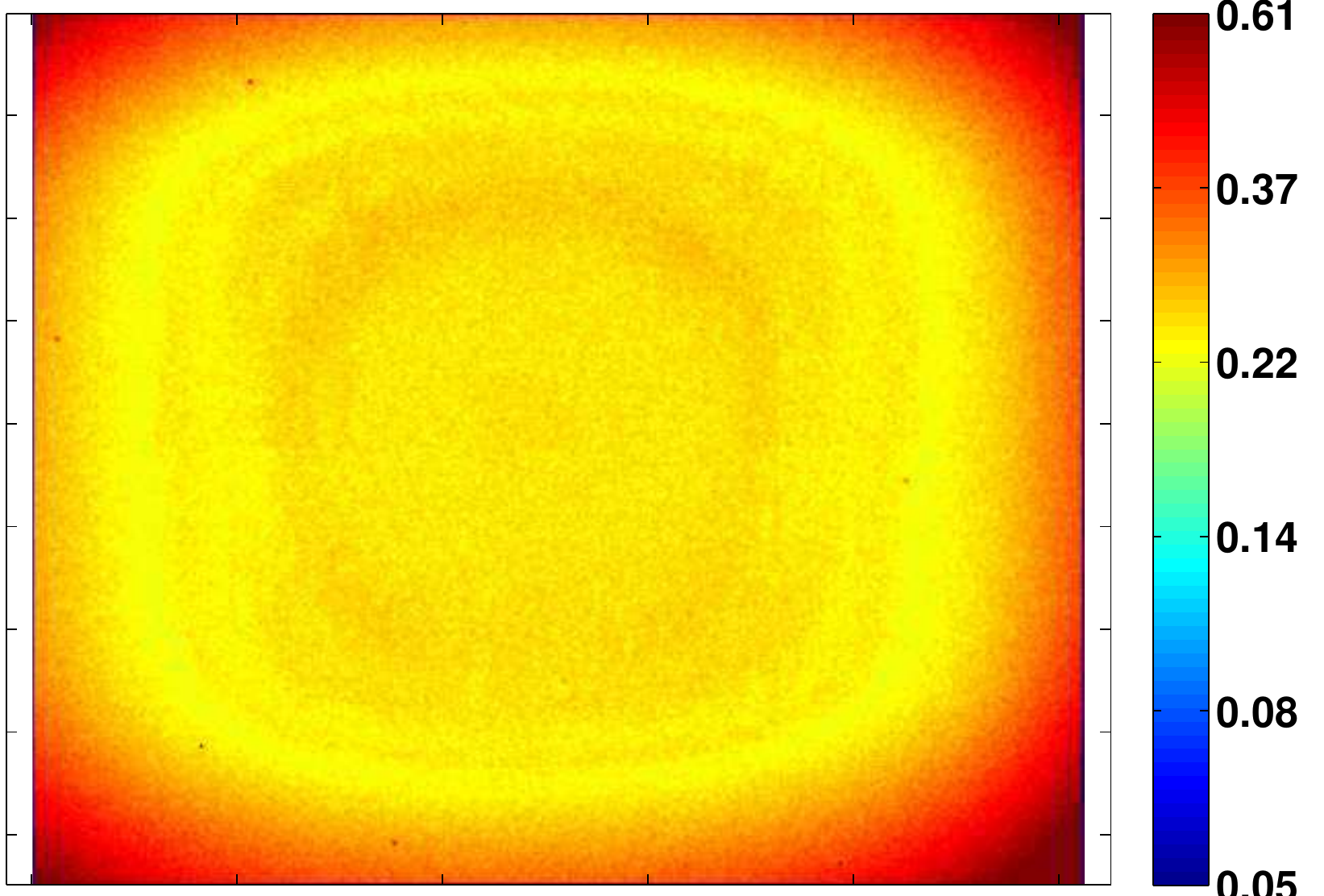} &
    \includegraphics[trim = 32mm 86mm 11mm 98mm,clip,width=.3\textwidth]{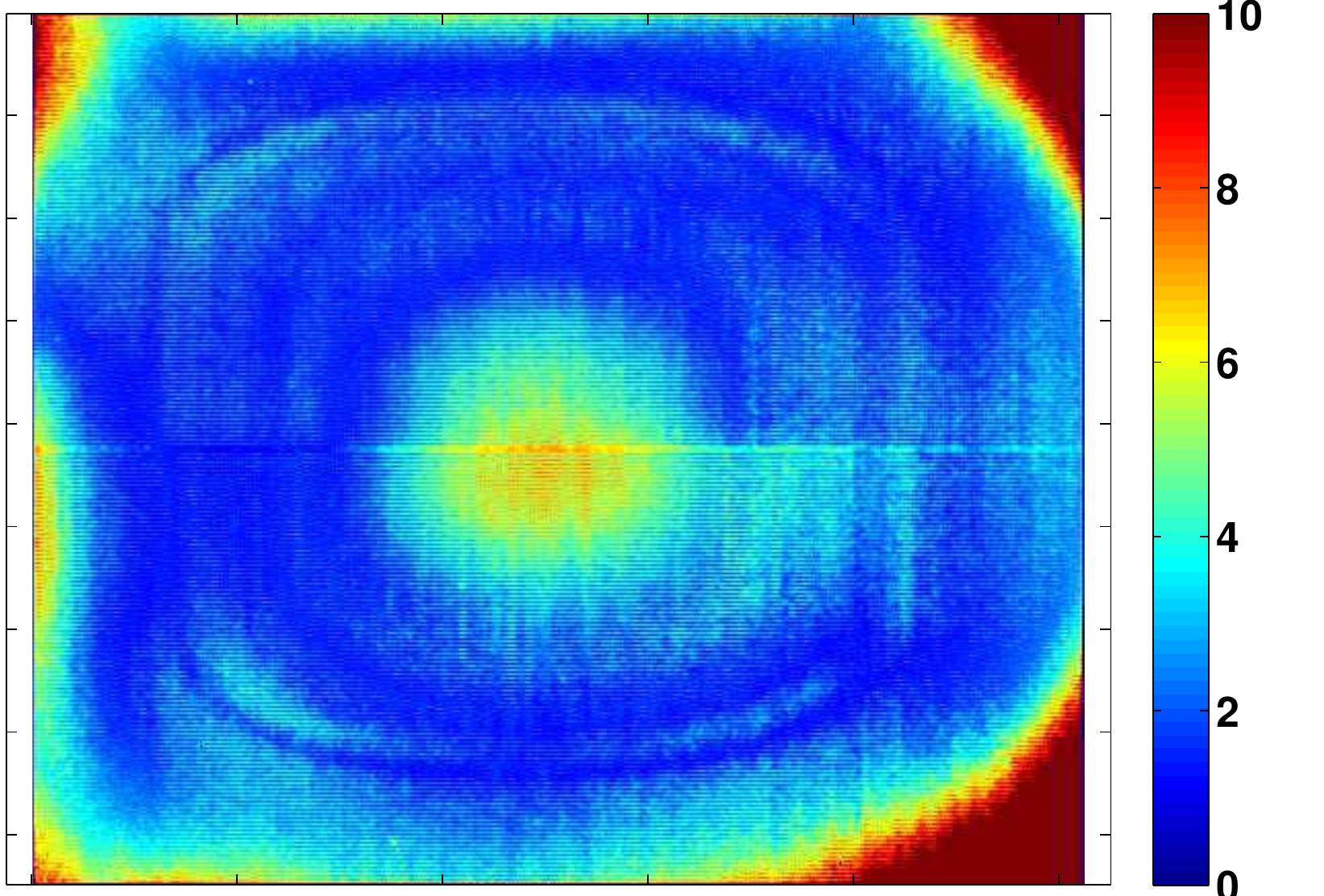}\\
    \begin{rotate}{90}\small\kintof~Raw\end{rotate}&
    \includegraphics[trim = 32mm 86mm 11mm 98mm,clip,width=.3\textwidth]{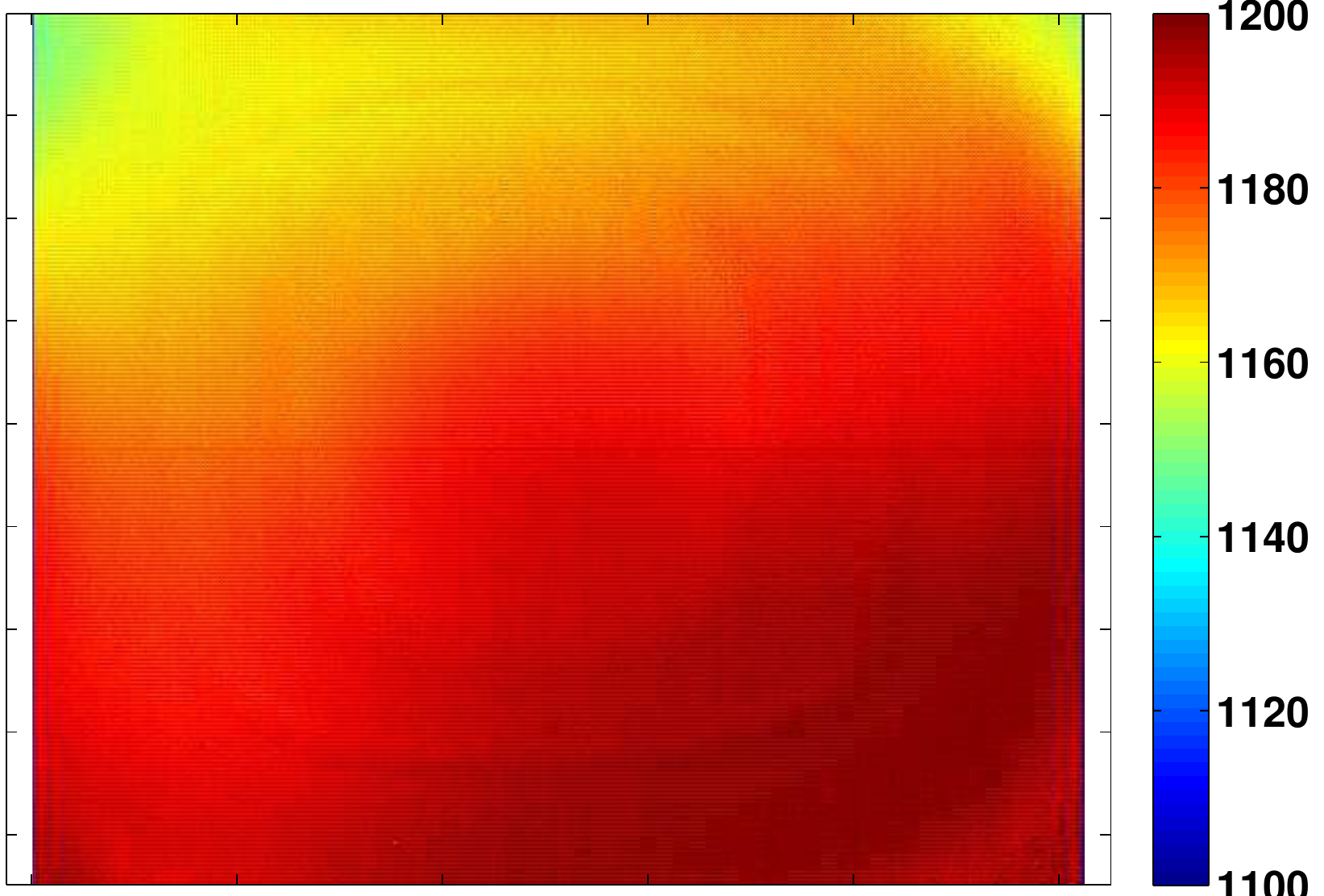} &
    \includegraphics[trim = 32mm 86mm 11mm 98mm,clip,width=.3\textwidth]{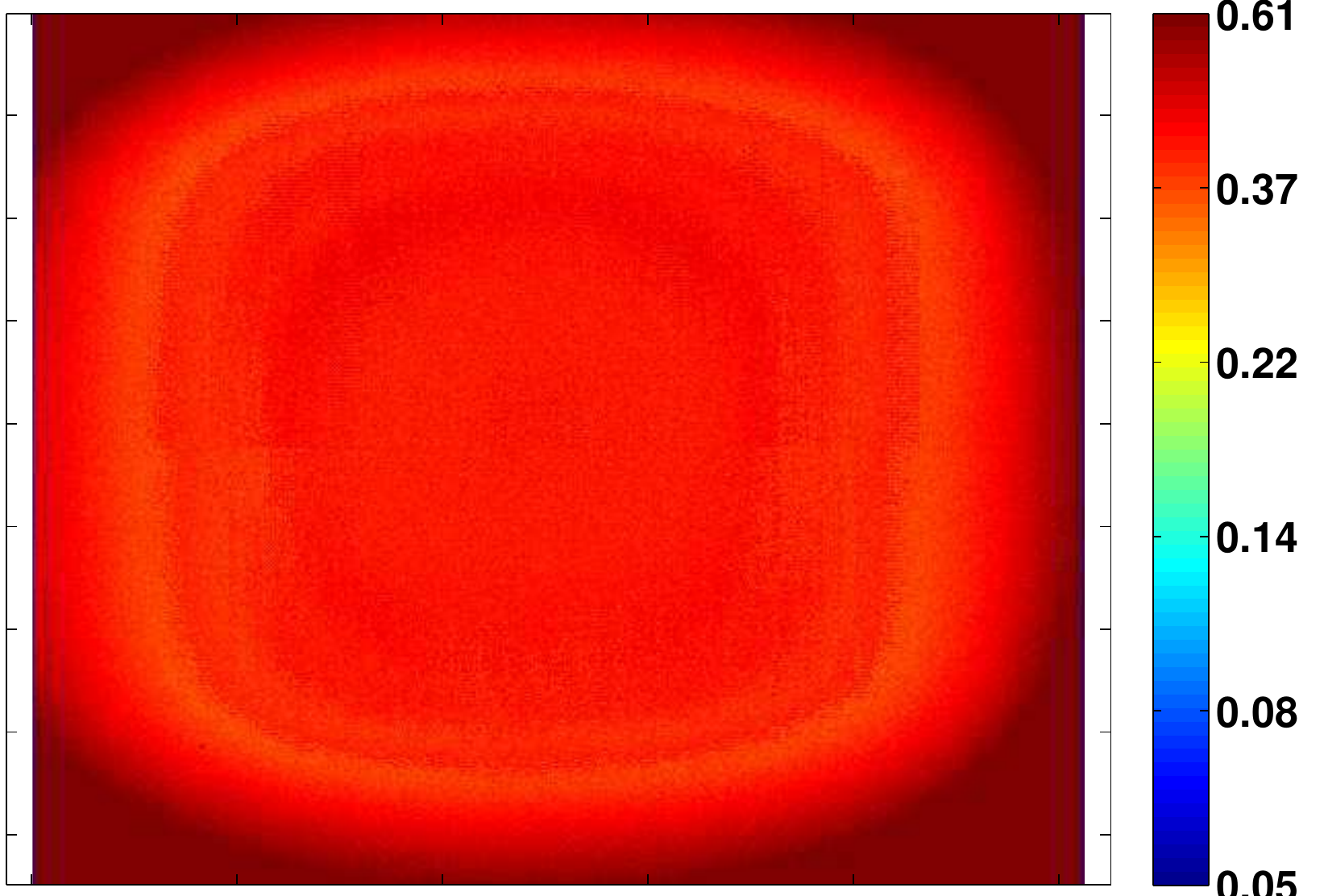} &
    \includegraphics[trim = 32mm 86mm 11mm 98mm,clip,width=.3\textwidth]{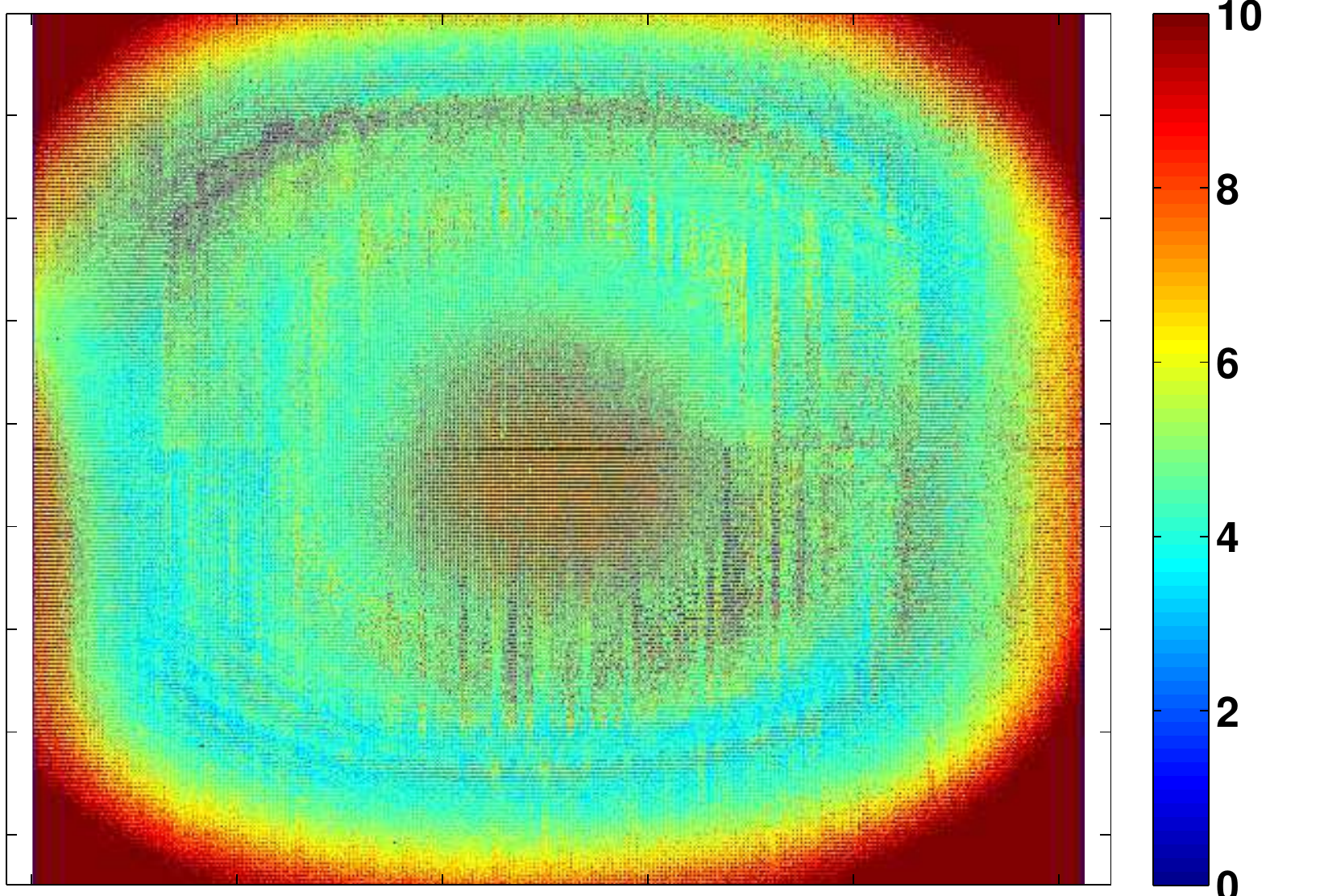}
  \end{tabular}
  \caption{Statistics of $200$~frame for \kinsl\ and
    \kintof\ acquiring a planar wall (values in $mm$): Mean (left
    col.), standard deviation (middle col.)  and RMSE with respect of
    a fitted plane (right) for the \kinsl\ (official driver,
    1$^{\text{st}}$ row, post-filtered range images, 2$^{\text{nd}}$
    row) and for the \kintof (official driver, 3$^{\text{rd}}$ row,
    the re-engineered OpenKinect driver, 4$^{\text{th}}$ row, and the
    raw range data delivered by the OpenKinect driver, 5$^{\text{th}}$
    row).}
  \label{fig:driver-filter-planar}
\end{figure}

\begin{figure}[t!]
  \centering
  \begin{tabular}{cccc}
    \includegraphics[trim = 60mm 42mm 53mm 23mm,clip,width=.22\textwidth]{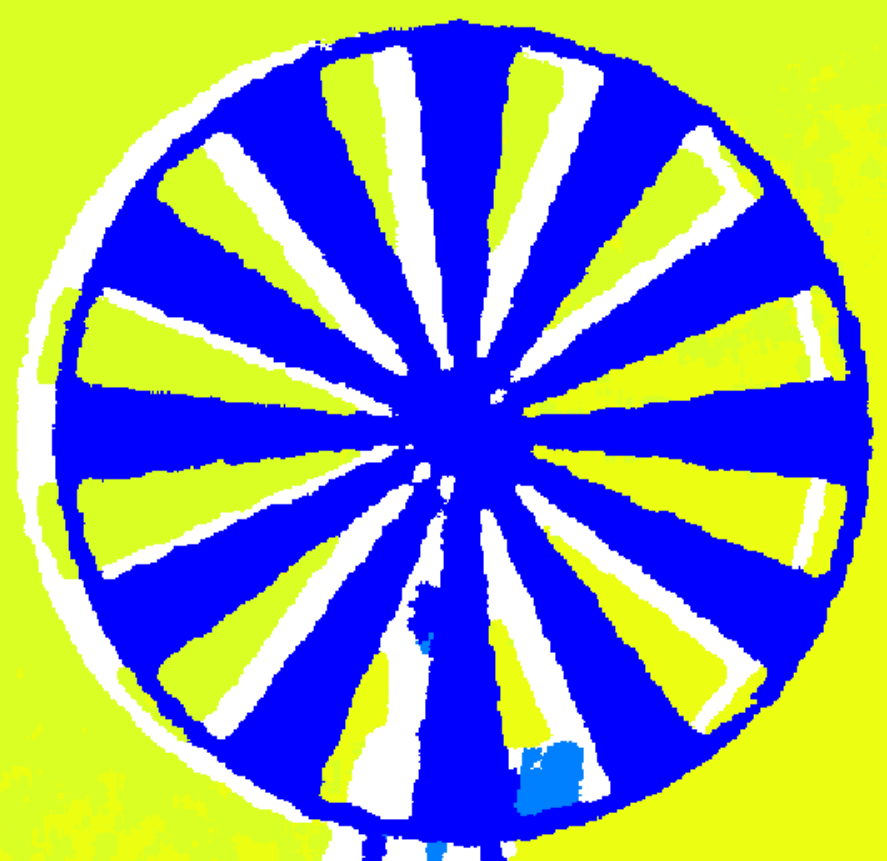} &
    \includegraphics[trim = 60mm 42mm 53mm 23mm,clip,width=.22\textwidth]{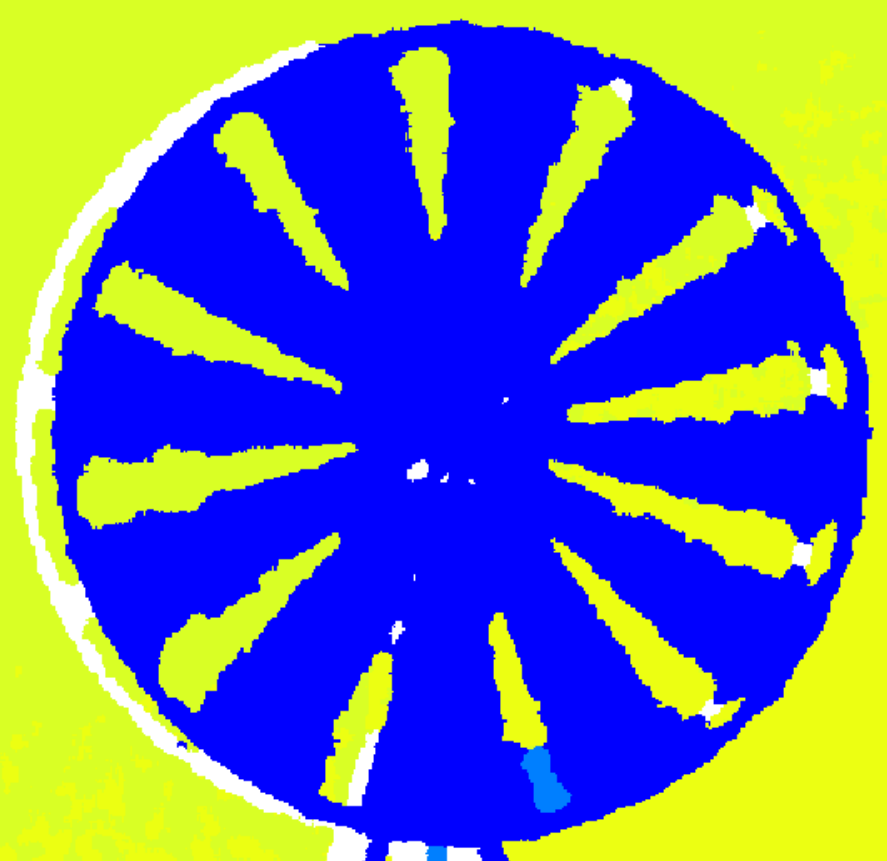} &
	    \includegraphics[trim = 60mm 40mm 53mm 23mm,clip,width=.22\textwidth]{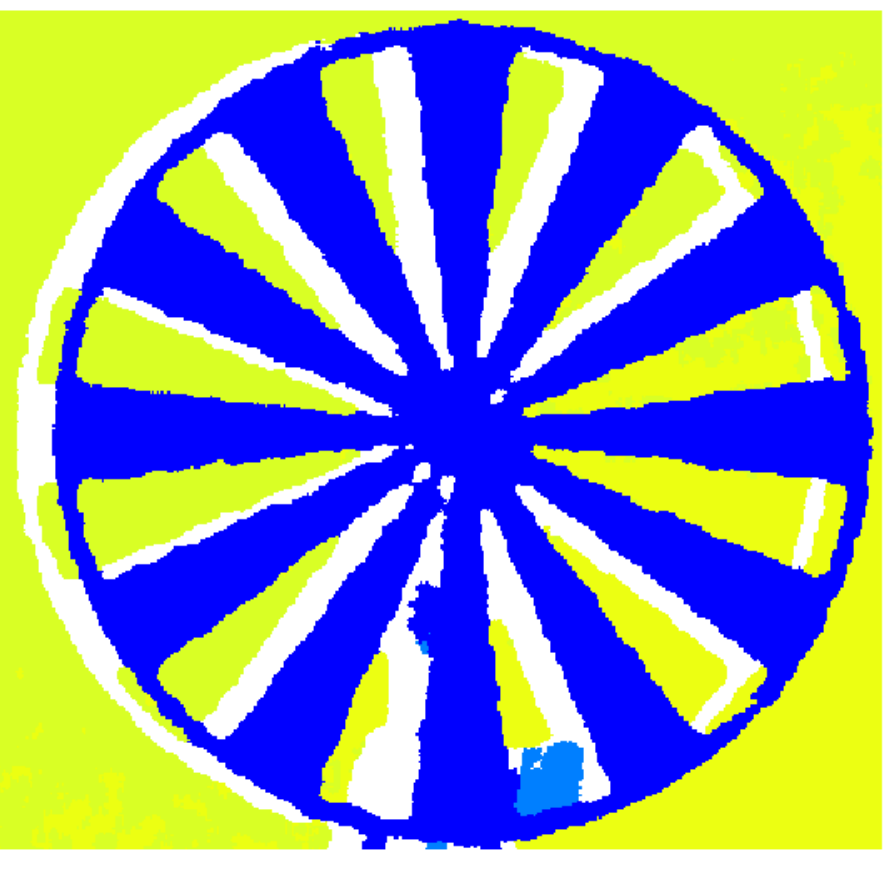} &
    \includegraphics[trim = 60mm 40mm 53mm 23mm,clip,width=.22\textwidth]{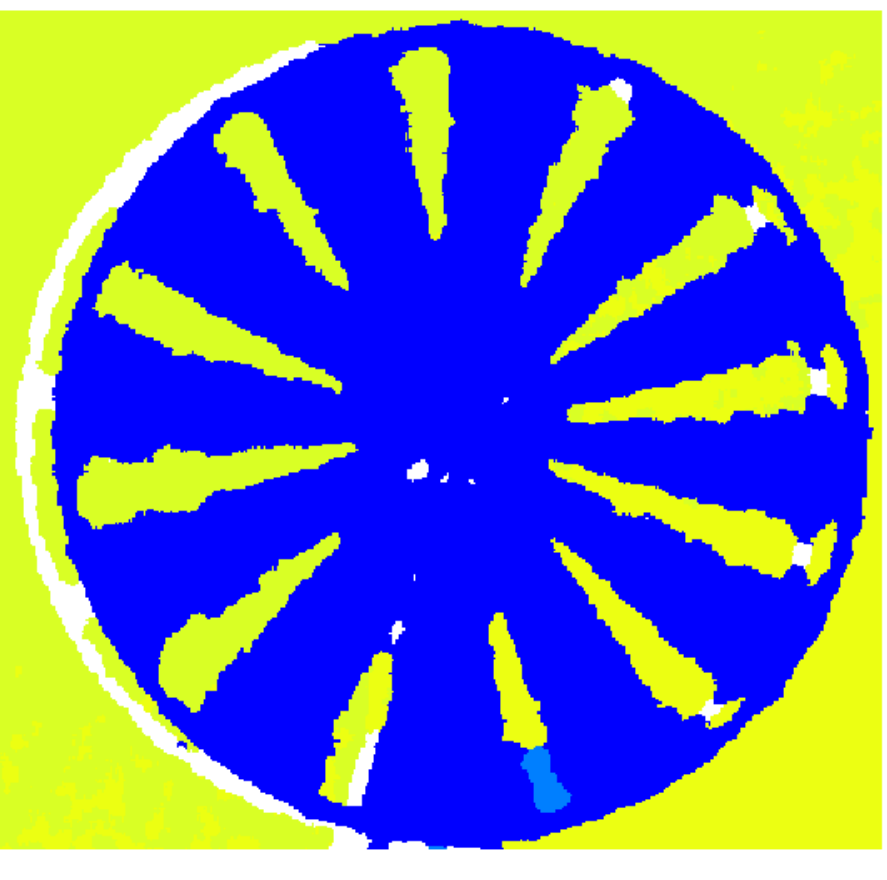}\\
    \multicolumn{2}{c}{\small \kinsl\ Offic.} &
    \multicolumn{2}{c}{\small \kinsl\ Post} \\
    \includegraphics[trim = 60mm 58mm 70mm 26mm,clip,width=.22\textwidth]{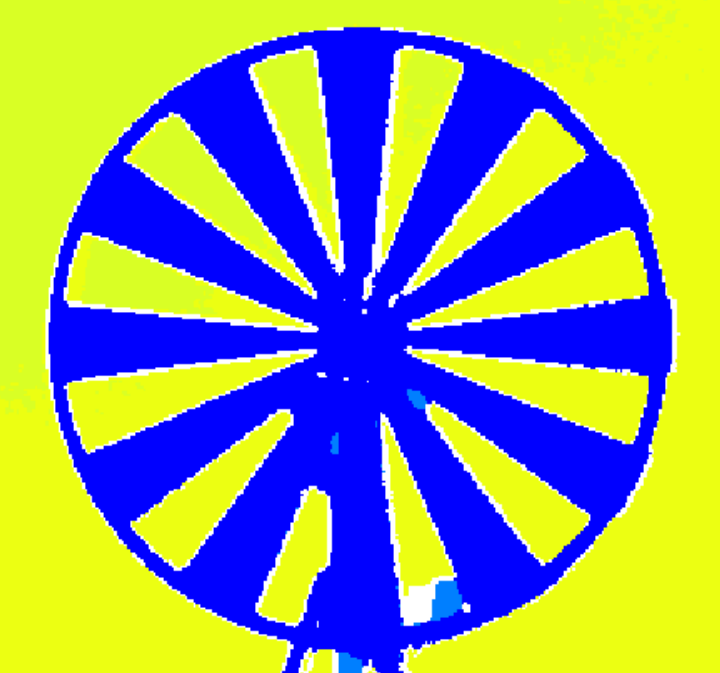} & 
    \includegraphics[trim = 60mm 58mm 70mm 26mm,clip,width=.22\textwidth]{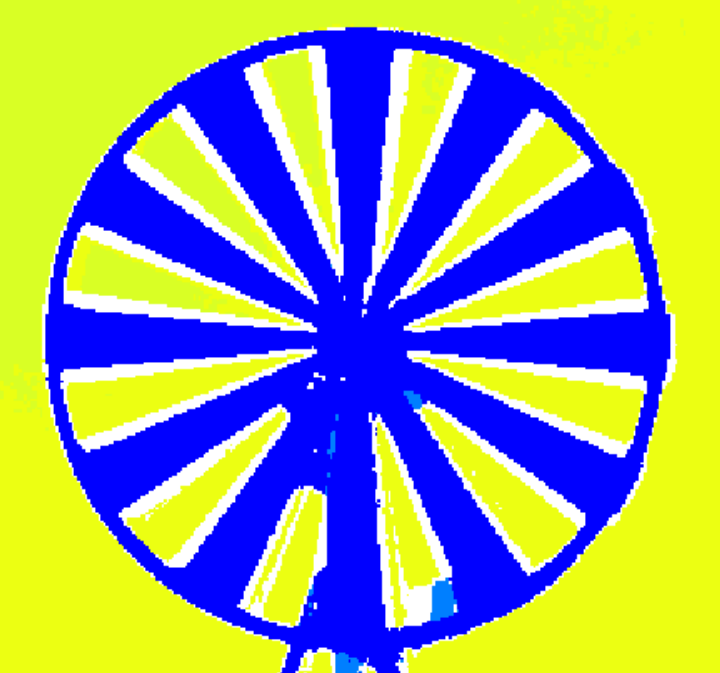} &
    \includegraphics[trim = 60mm 40mm 70mm 40mm,clip,width=.22\textwidth]{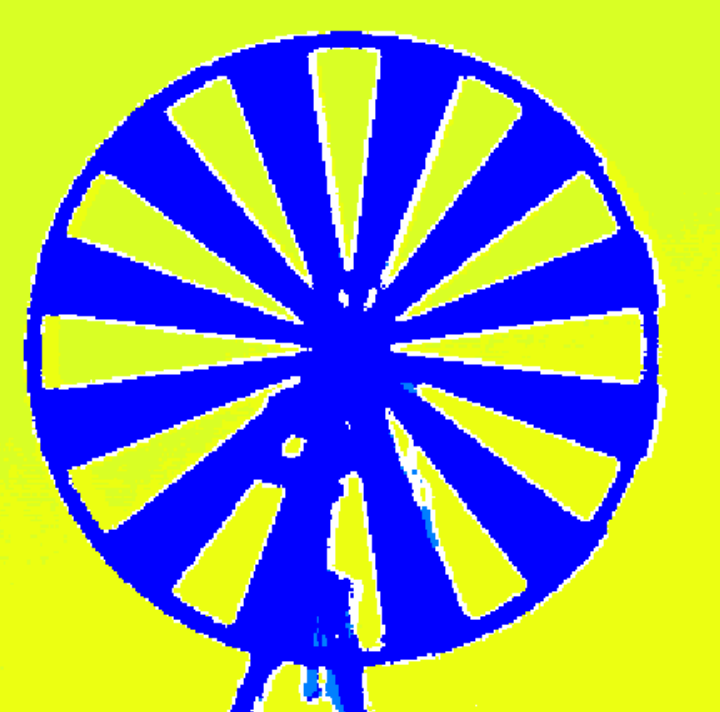} &
    \includegraphics[trim = 60mm 40mm 70mm 40mm,clip,width=.22\textwidth]{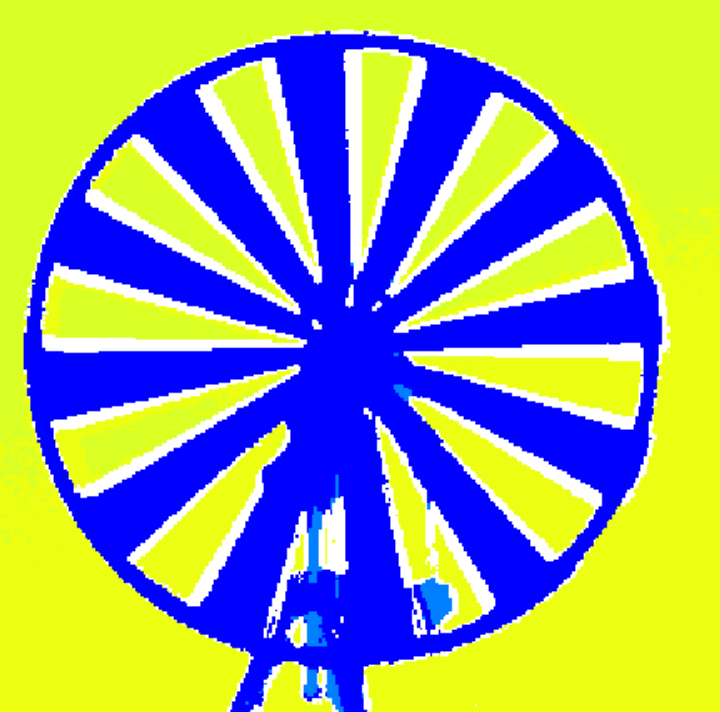}\\    
    \multicolumn{2}{c}{\small \kintof\ Offic.} &
    \multicolumn{2}{c}{\small \kintof\ Open} \\
    & \includegraphics[trim = 60mm 40mm 70mm 40mm,clip,width=.22\textwidth]{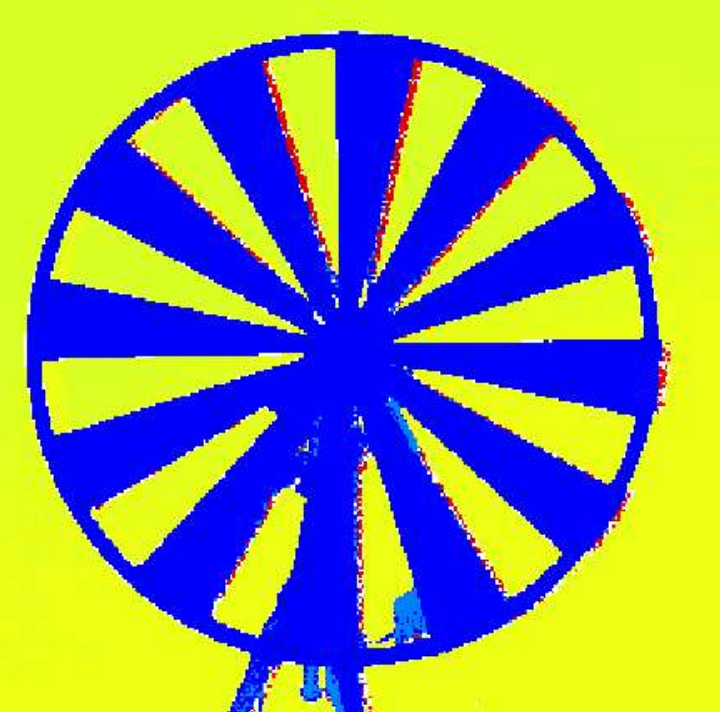} &
    \includegraphics[trim = 60mm 40mm 70mm 40mm,clip,width=.22\textwidth]{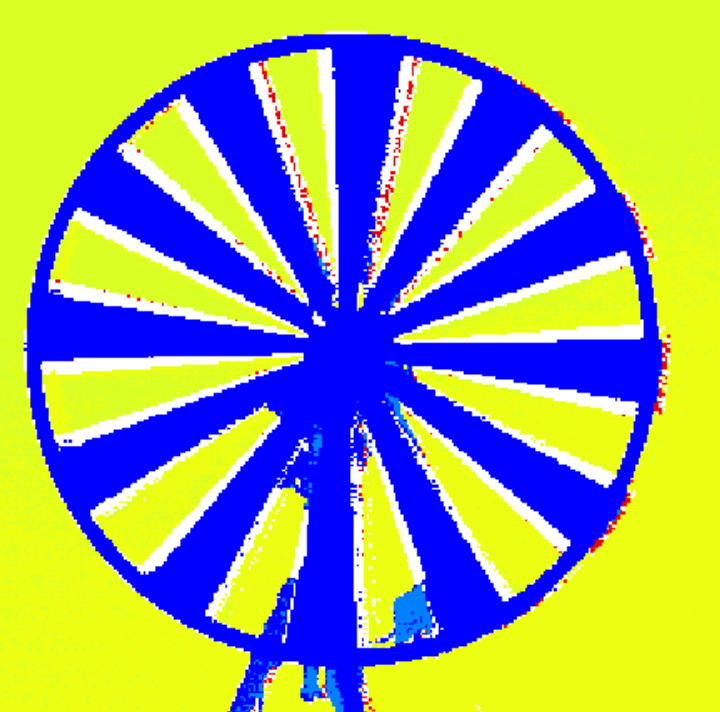} &
    \includegraphics[trim = 170mm 20mm 0mm 0mm,clip,height=.16\textheight]{images/MotionArtifacts/MotionK2Speed12800filter0-eps-converted-to.pdf} \\
   & \multicolumn{2}{c}{\small \kintof\ Raw} &
  \end{tabular}
  \caption{Single depth frame for a Siemens star for \kinsl\ and
    \kintof, range in $mm$: The range images are acquired for the
    static (left image) and the rotating star ($60$~RPM, right image)
    for \kinsl\ (official driver, top left, and post-filtered range
    images, top right) and for the \kintof (official driver, middle
    left, the re-engineered OpenKinect driver, middle right, and the
    raw range data delivered by the OpenKinect driver, bottom).  White
    color indicates invalid pixels.}
  \label{fig:driver-filter-motion}
\end{figure}

\section{General Considerations for Comparing \kinsl\ and \kintof}
\label{sec:comparison}

Before presenting the experimental setups and the comparison between
the two Kinect devices, we have to consider the limitations which
this kind of comparison encounters. For both, the \kinsl\ and the
\kintof\ camera, there are no official, publicly available reference
implementations which explain all stages from raw data acquisition to
the final range data delivery. Thus, any effect observed may either
relate to the sensor hardware, i.e. to the measurement principle as
such, or to the algorithms applied to raw data or, in a
post-processing manner, to the range data which are integrated in the
camera systems.

Anticipating the further discussion in this section, we explicitly
opted to work with both Kinect cameras in a ``black box'' manner using
the official drivers, as it is impossible to achieve ``fair
conditions'' for the comparison, i.e. a comparison which neutralizes
the effects from diverse filters applied in range data
processing. This is mainly due to the fact, that data processing is
applied on the level of raw data, i.e. disparity maps or correction images,
as well as on the level of range data; see detailed discussion
below. Attempts to reverse engineer the processing functionality
usually do not lead to the same data quality; see below. Thus,
taking the devices as they are, including the official, closed-source
drivers, is the most appropriate approach from the perspective in
utilizing them for any kind of application.

However, the disparity map from the \kinsl\ is different from common
representation, i.e.  $0$ disparity value does not refer to an
infinite distance. According to the reverse engineered disparity map
computation from ROS.org, the disparity map is normalized and
quantized between $0$ and $2047$ (using 11 bits storage), that requires
a more complex mapping function in order to convert disparity into
depth values. Note, that the quantization of the disparity map leads
to quantization of range values, which in some cases negatively
influences the statistical analysis or, in some cases, make it
completely useless. For example, it is impossible to derive a
per-pixel noise model for the \kinsl\ taking only individual pixel
distance measurements of a static scene; see
Sec.~\ref{sec:results:parameters} and
Nguyen\etal\cite{nguyen12kinectsl-noise}.

Different alternatives have been proposed for depth value estimation
for \kinsl\ disparity maps~\cite{khoshelhamAAR12}. In general, it is
possible to access the raw data of the \kinsl\ camera, i.e. infrared
image of the scene with the dots pattern, but it would go far beyond
the scope of this paper to provide further insight into the \kinsl's
method of operation by reverse engineering. On the other hand, solely
post-processing the delivered range data hardly improves the quality;
see below.

As described in Sec.~\ref{sec:principle.tof}. the \kintof\ camera
applies the CW approach. Additionally, the reverse engineered
OpenKinect driver~\cite{blake15openKinectToF} gives insight into some
details of data processing applied in the \kintof\.  In a first
processing stage, the correction images are converted to intermediate
images. At this stage a bilateral filter is applied. In a second
stage, the final range data is computed out of the intermediate
images, joining the three different range values retrieved from the
three frequencies. At this level, an outlier (or flying pixel) removal
is applied. The OpenKinect driver allows to deactivate the two
filters, thus the delivered range data can be considered as being
based raw correction images.

The described functionality allows data access on several levels, i.e.
\begin{itemize}
\item \emph{\kinsl~Offic} and \emph{\kintof~Offic}: Range data as
  delivered by the official driver provided by
  Microsoft for the
  \kinsl\footnote{\href{http://go.microsoft.com/fwlink/?LinkID=323588}{Kinect
      For Windows SDK $1.8$}} and for the \kintof\ using the Developer
  Preview
  driver\footnote{\href{http://www.microsoft.com/en-us/download/details.aspx?id=43661}{Kinect
      For Windows SDK $2.0$ (JuneSDK)}}.
\item \emph{\kinsl~Post}: Additional post-processing using
  filtering; here we use a bilateral filter. Note, that the filter has
  to operate on data with already masked out, i.e. invalid pixels.
\item \emph{\kintof~Open}: The reengineered data processing of the
  OpenKinect driver by Blake\etal\cite{blake15openKinectToF}.
\item \emph{\kintof~Raw}: The reengineered data processing of the
  OpenKinect driver by Blake\etal\cite{blake15openKinectToF} with
  deactivated filtering, i.e. range data directly computed from the
  raw data.
\end{itemize}

We apply these five different options to a simple static scene, where
the cameras observe a planar wall, analyzing the statistics for
$200$~frames; see Fig.~\ref{fig:driver-filter-planar} and
Sec.~\ref{sec:results:dynamic}). For this scenario, the data is
comparable among different drivers of a device, as the cameras have not been
moved while switching to a different driver. However, the data is not fully
comparable between \kinsl\ and \kintof. Additionally, we used a
dynamic scenery with a rotating Siemens star; see
Fig.~\ref{fig:driver-filter-motion} and
Sec.~\ref{sec:results:dynamic}).

The results for the static wall and the Siemens star are presented in
Figs.~\ref{fig:driver-filter-planar} and
\ref{fig:driver-filter-motion}, respectively. The results can be
summarized as follows:
\begin{itemize}
\item Post-processing the \kinsl-data does not improve the quality, as
  the problematic regions of the range image are already masked out;
  see Fig.~\ref{fig:driver-filter-motion}, top row. The quality of the
  \kinsl\ device is mainly driven by strong depth quantization
  artifacts, which get apparent in the standard deviation; see
  Fig.~\ref{fig:driver-filter-planar}, middle column, first two rows.
\item The quality of the OpenKinect driver~\cite{blake15openKinectToF}
  stays somewhat behind the official \kintof-driver; see
  Fig~\ref{fig:driver-filter-planar}, 3$^{\text{rd}}$ and
  4$^{\text{th}}$ row, i.e. the reverse engineering appears to be
  functionally not fully complete.
\item Disabling the internal filters for the \kintof\ mainly shows
  negative effects for the rotating Siemens star; see
  Fig.~\ref{fig:driver-filter-motion}. The filtering of the correction
  images and the flying pixel removal clearly removes the artifacts at
  the jumping edges of the Siemens star.
\end{itemize}

\section{Experimental Results And Comparison}
\label{sec:results}

\newcommand{\f}{$\bullet$}
\newcommand{\s}{$\circ$}
\begin{table}
  \begin{tabular}[H]{|p{.45\textwidth}|c|c|c|c|c|c|c|c|c|} \hline
    \centering
    \rule{0pt}{38mm}
    \bf Test-Scenarios$\backslash$Effect & 
    ~~\begin{rotate}{90}\small\bfseries Amb. Backgr. Light\end{rotate}&
    ~~\begin{rotate}{90}\small\bfseries Multi-Device Interf.\end{rotate}&
    ~~\begin{rotate}{90}\small\bfseries Temperature Drift\end{rotate}&
    ~~\begin{rotate}{90}\small\bfseries Systematic Error\end{rotate}&
    ~~\begin{rotate}{90}\small\bfseries Depth Inhomogeneity\end{rotate}&
    ~~\begin{rotate}{90}\small\bfseries Multipath Effect\end{rotate}&
    ~~\begin{rotate}{90}\small\bfseries Intens.-Rel. Error\end{rotate}&
    ~~\begin{rotate}{90}\small\bfseries Semitrans. \& Scatter.\end{rotate}&
    ~~\begin{rotate}{90}\small\bfseries Dynamic Scenery\end{rotate}\\ \hline\hline
    \bfseries Ambient Background Light  &  &\f&  &  &  &  &  &  & \\ \hline
    \bfseries Multi-Device Interference &\f&\s&  &  &  &  &  &  & \\ \hline
    \bfseries Device Warm-Up            &  &  &\f&  &  &  &  &  & \\ \hline
    \bfseries Rail Depth Tracking       &  &  &  &\f&  &  &\f&  & \\ \hline
    \bfseries Semitransparent Liquid    &  &  &  &  &  &\s&  &\f& \\ \hline
    \bfseries Reflective Board          &  &  &  &  &  &\f&  &  & \\ \hline
    \bfseries Turning Siemens star      &  &  &  &  &\f&  &  &  & \f \\ \hline
  \end{tabular} 
  \caption{The different effects relevant SL- and ToF-based range
    sensing systems and their relation to the designed test
    scenarios. Each test addresses primarily one or two separable
    effects denoted by \f\ and may address also secondary effects,
    denoted by \s.}        
  \label{tab:test-effect}      
\end{table}

In section \ref{sec:results.ambient}--\ref{sec:results:dynamic} we
present the different test scenarios we designed in order to capture
specific error sources of the \kinsl\ and the \kintof-cameras. Before
going into the scenarios, in Sec.~\ref{sec:results:parameters} we will
briefly present the camera parameters and the pixel statistics.

Our major design goal for the test scenarios was to capture individual
effects as isolatedly as possible. Furthermore, we designed the
scenarios in a way, that they can be reproduced in order to adopt them
to any other range sensing system that works in a similar depth
range. Tab.~\ref{tab:test-effect} gives an overview of the different
test scenarios and the effects they address; see also
Sec.~\ref{sec:principle.err}.  We focus on the range of $500$mm to
$3000$mm\footnote{\href{http://msdn.microsoft.com/en-us/library/hh438998.aspx}{Microsoft
    Developer network, Kinect sensor}} as we operate the \kinsl\ in
the so-called near-range-mode, which is optimized for this depth
range. Also it covers the depth range supported by \kintof\ which is
$500$mm to
$4500$mm\footnote{\href{http://www.microsoft.com/en-us/kinectforwindows/meetkinect/features.aspx}{Kinect
    for windows, features}}.

For all tests we utilize a \kinsl\ (Kinect for Windows v1 sensor with
activated near mode) and a \kintof\ (Microsoft camera prototype
available from the Developer Preview Program). Data access for the
\kinsl\ is done via the official driver provided by
Microsoft\footnote{\href{http://go.microsoft.com/fwlink/?LinkID=323588}{Kinect
    For Windows SDK $1.8$}} and for the \kintof\ using the Developer
Preview
driver\footnote{\href{http://www.microsoft.com/en-us/download/details.aspx?id=43661}{Kinect
    For Windows SDK $2.0$ (JuneSDK)}}. All
data evaluations have been done using Matlab.

The major quantitative results for the comprehensive comparison are
summarized in Tab.~\ref{tab:conclusion} indicating the major
differences, strengths and limitations of both systems.

At this point, we want to refer to the discussion in
Sec.~\ref{sec:comparison} state explicitly, that both Kinect cameras
are used in a ``black box'' manner. Thus, even though we refer to
characteristics of the specific range measurement techniques, the
resulting effects may not only relate to the sensor hardware, i.e. to
the measurement principle as such, but also to the post-processing
integrated into the cameras

\subsection{Camera Parameters and Noise Statistics}
\label{sec:results:parameters}

\begin{table}[t]
  \centering
  \begin{tabular}{ | l | l | l |}
    \hline
    Parameter & \kinsl & \kintof \\  \hline
    Resolution & 640$\times$480 &  512$\times$424\\
    Focal Length [px]& (583.46, 583.46) & (370.79, 370.20) \\
    Center [px] & (318.58, 251.55) & (263.35, 202.61) \\
    Dist. ($k_1$; $k_2$; $k_3$; $p_1$; $p_2$) &
    -0.07377, 0.1641, 0, 0, 0 & 0.09497,
    -0.2426, 0, \\ && 0.00076, -0.00017 \\ \hline
  \end{tabular}
  \caption{Camera parameters of the depth sensors for \kinsl\ and
    \kintof. The distortion coefficients are radial ($k_1,k_2,k_3$) and
    tangential distortion ($p_1,p_2$).}
  \label{tab:camera-parameters}
\end{table}

\begin{table}[b]
  \centering
  \begin{tabular}{ | l | l | l | l | l |}
    \hline
    Pixel & Gaussian ($\mu, \sigma$) & $RMSD_g$ & Poisson ($\lambda, \delta_x$) & $RMSD_p$\\ \hline
   
    Center & [1023.91, 4.42] & 0.0025 & [17.47, 1007.47] & 0.0024\\
    Intermed. & [1074.77, 3.77] & 0.0017 & [14.10, 1061.40] & 0.0017 \\
    Corner & [1127.21, 24.03] & 0.0019 & [101.73, 1030.02] & 0.0019 \\
 
\hline
  \end{tabular}
  \caption{Temporal statistics of three different pixels of
    \kintof\ sensor with the corresponding Gaussian and Poisson
    fits. The value $\delta_x$ [mm] denotes the shift applied range values
    to match the Poisson distribution.}
  \label{tab:noise-fits}
\end{table}

  \begin{figure}[t]
  \centering  \begin{tabular}{ccc}
    \small Center Pixel &
    \small Intermed. Pixel &
    \small Corner Pixel \\
    \includegraphics[width=.3\textwidth]{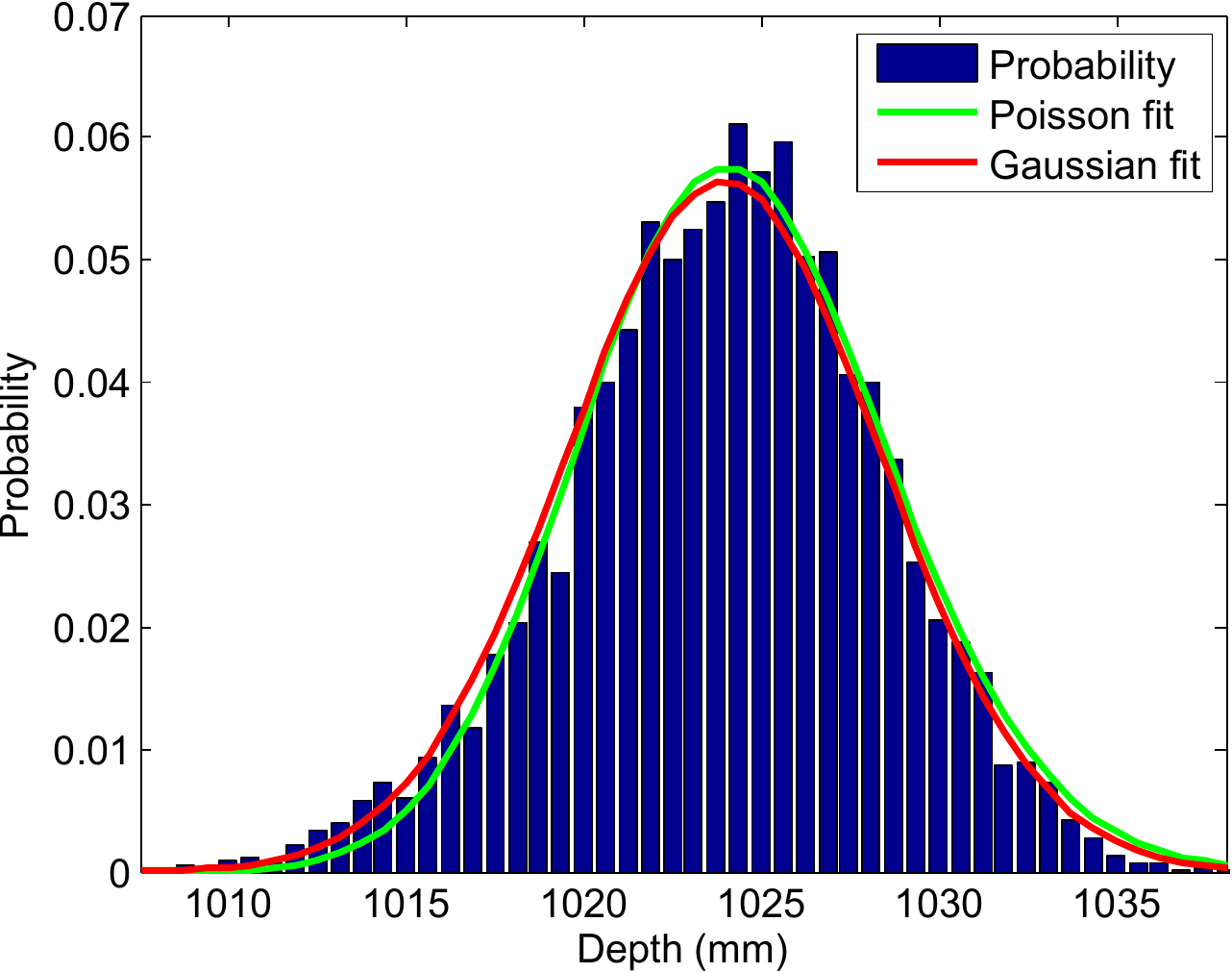}&
    \includegraphics[width=.3\textwidth]{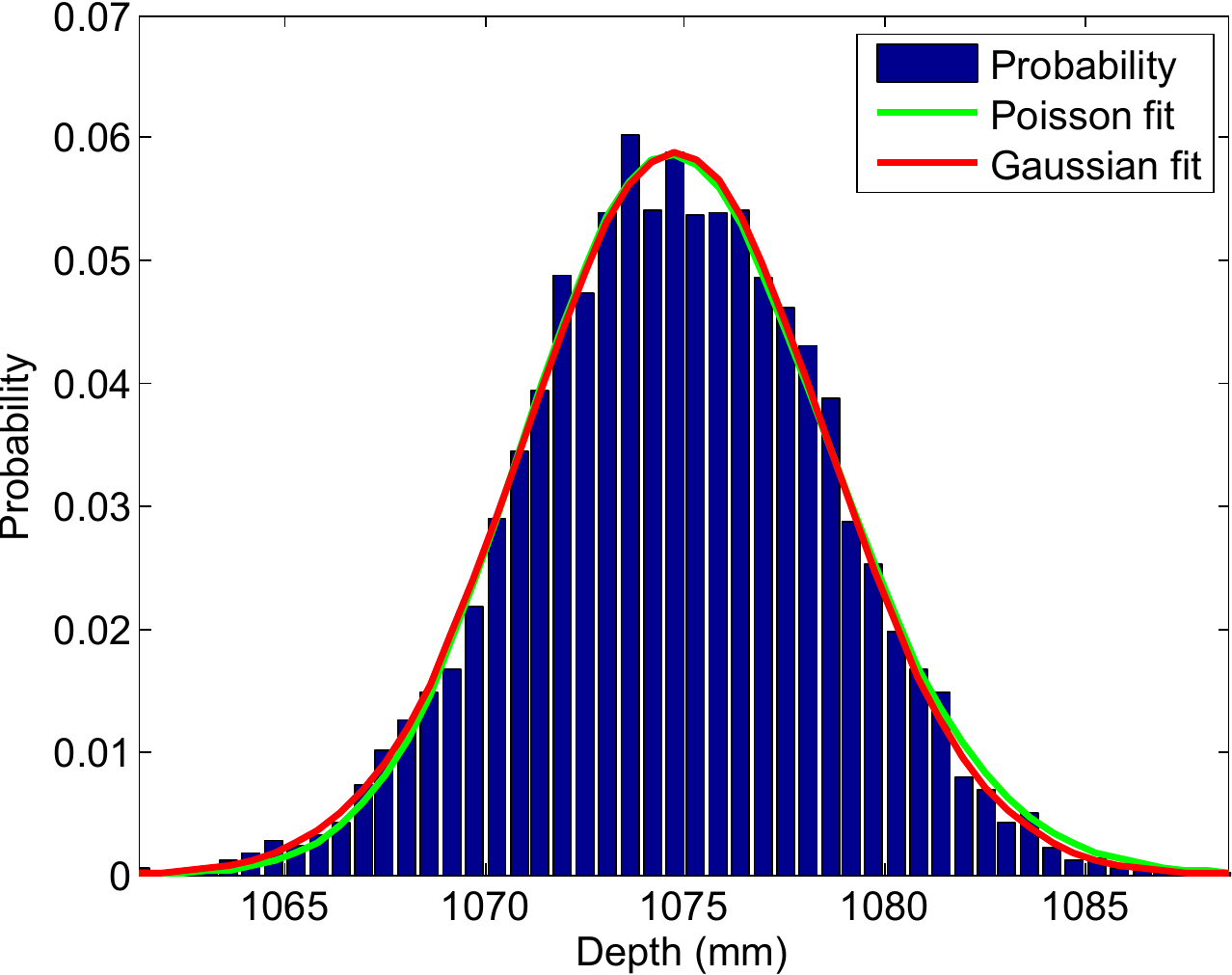}&    
    \includegraphics[width=.3\textwidth]{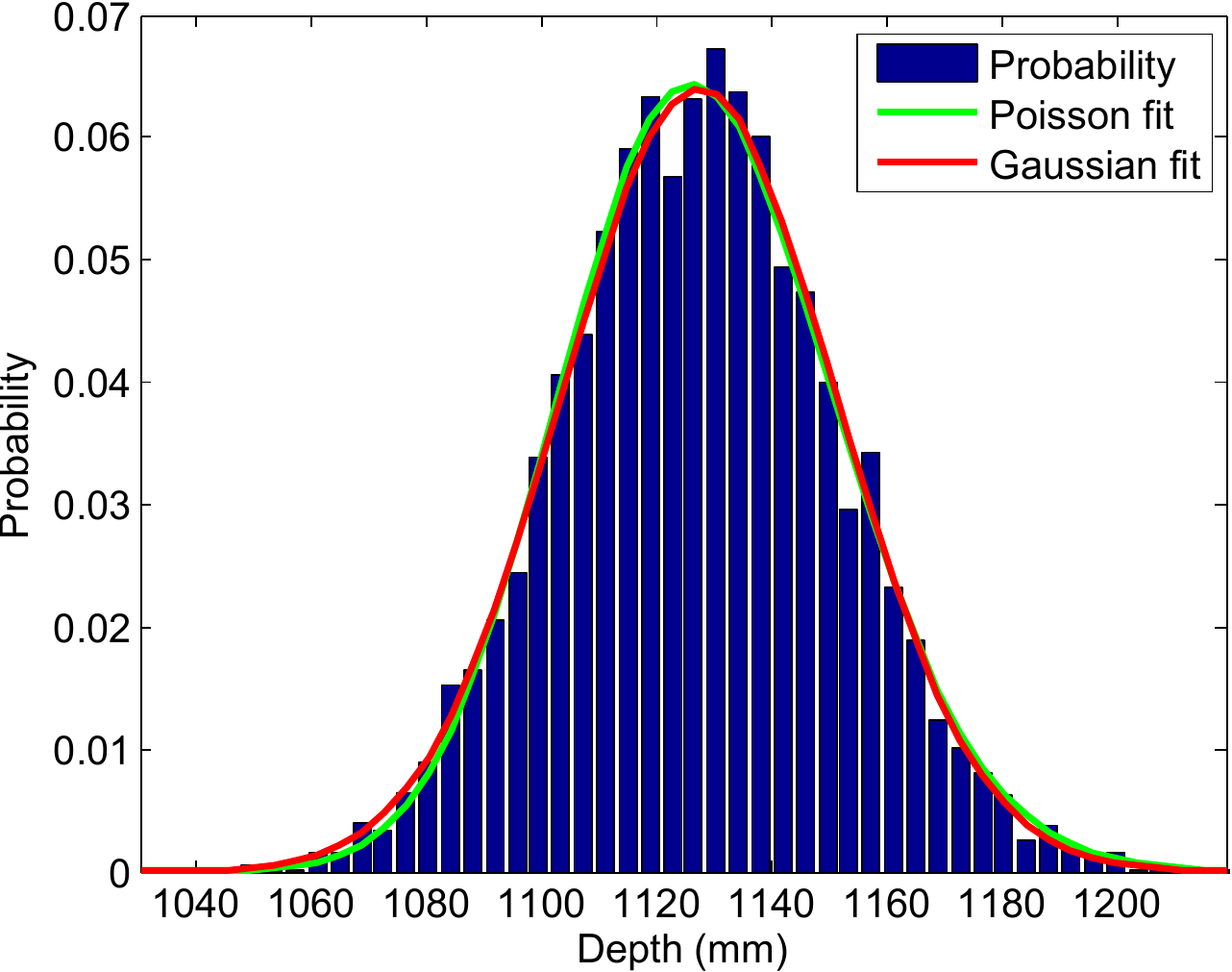}
  \end{tabular}
  \caption{Density distribution of 3 different pixels during
    $5000$~frames for \kintof\ acquiring a planar wall (values in
    $mm$)}
  \label{fig:statistics.gaussfit}
\end{figure}

As most applications require full 3D information, we first estimate
the intrinsic parameters for both devices using standard calibration
techniques based on a planar checkerboard from the OpenCV
library~\cite{bradskiTOL00}; see Tab.~\ref{tab:camera-parameters}. For
both devices, $50$ images of the checkerboard were acquired with
different orientations and distances. For the \kintof, we directly use
the amplitude image delivered by the camera. Whereas for the \kinsl, we
use the NIR image of the depth sensor. Since the dot pattern of the
\kinsl\ may degrade the checkerboard detection quality in the NIR
image, we block the laser illumination and illuminate the checkerboard
with an ambient illumination. 

Furthermore, we want to analyze the noise statistics for the
\kinsl\ and the \kintof. As already stated in
Sec.~\ref{sec:comparison}, the strong quantization applied in the
\kinsl\ makes it hard to derive per-pixel temporal statistic values. In
the literature there are alternative approaches using a mixed
spatiotemporal analysis to derive some kind of noise
statistics~\cite{nguyen12kinectsl-noise}, but this approach is
difficult to compare with pure temporal statistics. Therefore, we
focus on the \kintof's temporal statistics only.

For the temporal statistics we acquired $5000$~frames of the
\kintof\ observing a planar wall at about 1 meter distance. The
OpenKinect driver with deactivated filtering was used to obtain
unchanged range data. Fig.~\ref{fig:statistics.gaussfit} shows the
histograms for a central, an intermediate and a corner pixel of this
time series including fits for a Gaussian and a Poisson
distribution. Both fits were done using MATLAB. We use non linear
least square optimization approaches in order to get the suitable
parameters for the Poisson distribution. Tab.~\ref{tab:noise-fits}
gives the resulting parameters of both fitting for the three pixel
statistics as well as the corresponding RMSE. It can be noted that
corner pixels have a higher variance than pixel at the center area of
the image, which is due to a reduced amplitude of the illumination in
corner regions. We can also deduce that the Poisson distribution and
the Gaussian fitting results in the same fitting quality.

\subsection{Ambient Background Light}
\label{sec:results.ambient}

\paragraph{Goal} This test scenario addresses the influence of ambient
light onto the range measurement of the \kinsl\ and the
\kintof\ cameras. The primary goal for the experiment is to show the
relation between ambient background radiance incident to the Kinect
and the delivered depth range of the sensor. The main focus for this
experiment is thus to measure the incident background radiance with
respect to image regions accurately. 

As both Kinect cameras do have imperfect illumination in the sense,
that pixels in the vicinity on the image receive less active light
than pixels close to the center, a secondary goal is to given some
insight into a possible spatial variation of the influence of ambient
background light.

\begin{figure*}[t!]
  \centering
  \includegraphics[width=.7\textwidth]{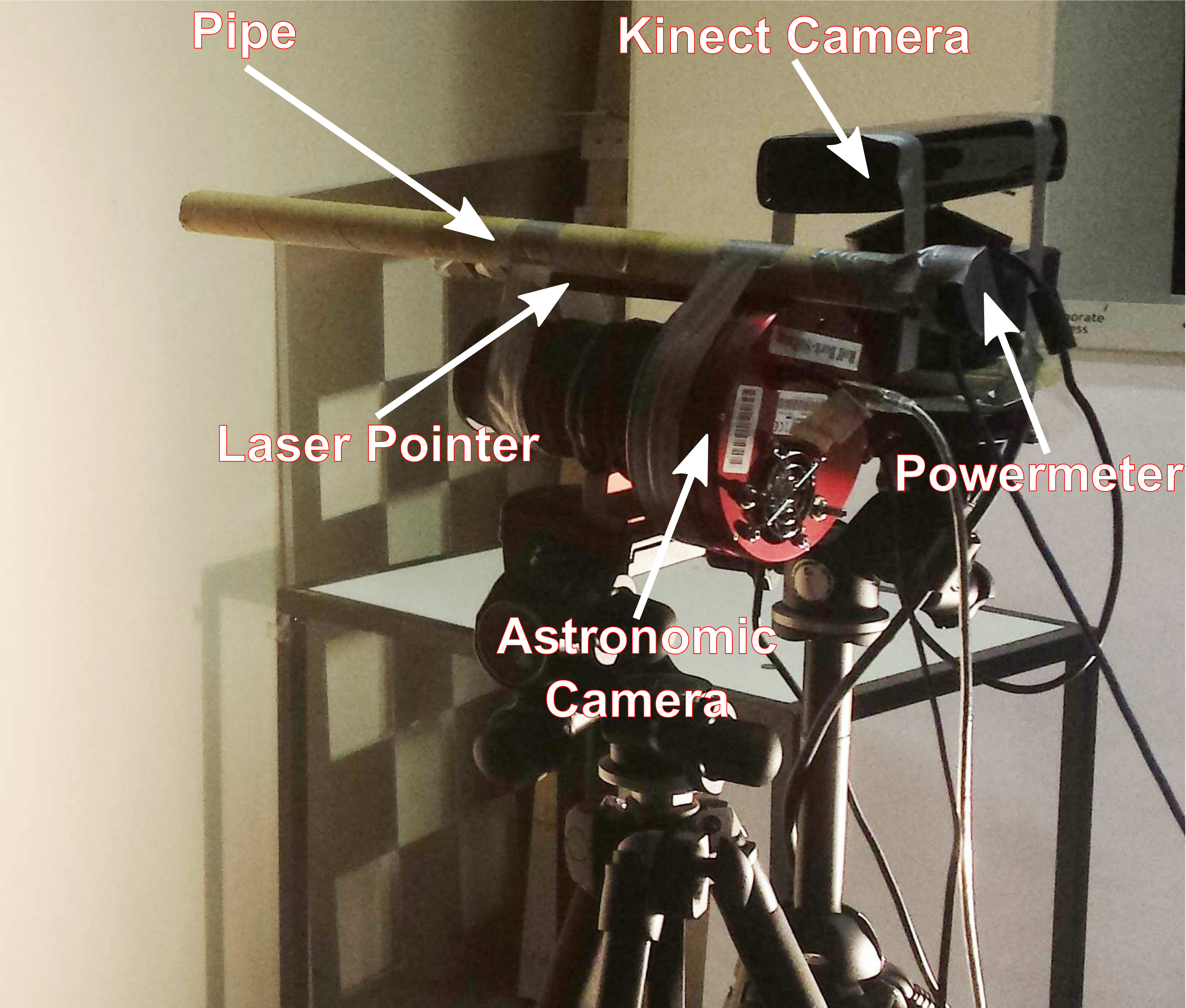} 
  \caption{Ambient background light experiment setup. The laser
    pointer is mainly hidden by the pipe.}
  \label{fig:kinect-signal}
\end{figure*}

\paragraph{Experimental Setup}
The Kinect camera is mounted $1200$mm in front of a approximately
diffuse white wall in an environment where the amount of light can be
controlled using three HALOLINE ECO OSRAM $400$W halogen lamps. The
radiosity on the wall depends on the number of active lamps and their
distance to the wall. We measure the radiant emittance of the surface
resulting from our light sources with a Newport 818-SL powermeter. The
powermeter directly delivers power intensity in W/cm$^2$ and it is calibrated to
$850$nm, which relates to the Kinect's laser diode illumination of
$850$nm. An additional laser pointer allows for the directional
adjustment of the powermeter's pipe to a point on the wall in order to
accurately measure the radiance. To register the point at the wall
with a pixel in the Kinect camera, we temporally attached a small
cuboid as depth marker to the wall.

As both Kinect cameras have an NIR-filter suppressing visible
light\footnote{We did not explicitly measure the NIR filter, as this
  would require to destroy the Kinect camera.}, we equip the
powermeter with an additional Kodak Wratten high-pass filter number
87C. This filter has a $50$\% cutoff at $850$nm. A pipe is mounted to
the powermeter in order to measure the incident radiance for a
specific spatial direction from single point on the wall.

We further add an Atik 4000 astronomy photography camera equipped with
the same NIR filter alongside with a powermeter in order to verify
radiance measurements provided by the powermeter setup. The astronomy
camera measures the radiant emittance in a linear relation to the
number of photons received per pixel, i.e. per observed
direction. In our experiments we found a proper linear relation
between both measurements.

We interrelate the radiance measurement of the powermeter to daylight
condition. Therefore, we acquired a radiant flux density reference
measurement with the powermeter setup without pipe of direct sunlight
on a sunny summer day in central Europe. This results in
$11$mW/cm$^2$. Furthermore, we relate the radiant flux density
measurement to the incident radiance measurement of an indirectly
illuminated diffuse white paper using the powermeter with pipe at
$1.2$m distance resulting in a factor of $1.1\cdot10^3$. As the later
setup is comparable to the evaluation setup for the Kinect cameras, we
can deduce a sun reference incident radiance value of about
$10\mu$W/cm$^2$.

The final radiance measurements are done with the powermeter
setup. The radiance measurements take place when the Kinect camera is
turned off in order to prevent interference with the camera's
illumination unit. We acquired $200$ frames for various light
conditions up to $20\mu$W/cm$^2$.  Since we expect some variation of
the effect for different pixel locations, we measured three points
along the half diagonal, i.e.  a point close to the upper left corner,
the principle point of the range image and one intermediate point in between both
points.

\begin{figure}[t!]
  \centering
  \includegraphics[trim = 7mm 50mm 9mm 91mm,clip,width=.49\textwidth]{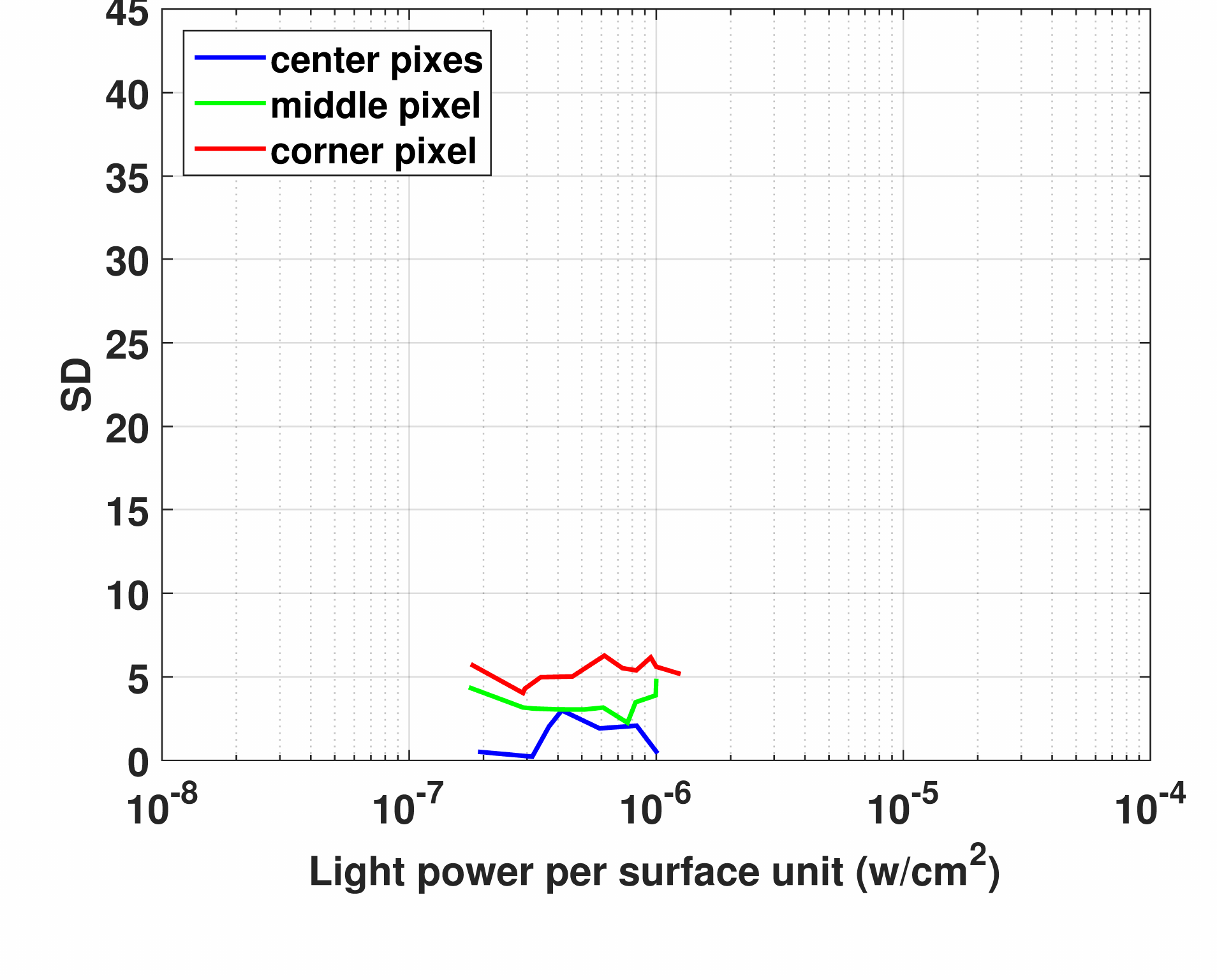}
  \includegraphics[trim = 7mm 50mm 9mm 91mm,clip,width=.49\textwidth]{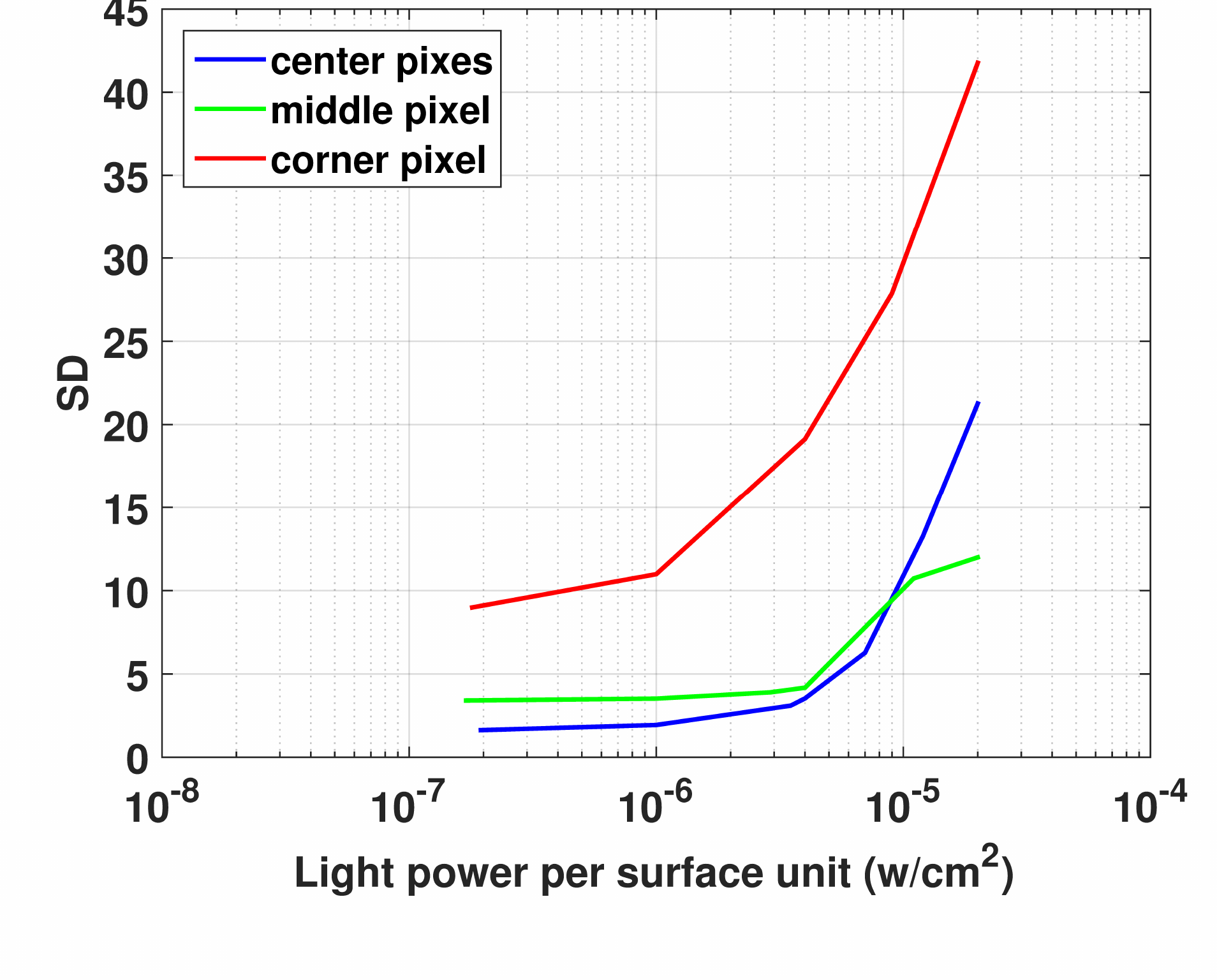}\\
   \includegraphics[trim = 7mm 50mm 9mm 91mm,clip,width=.49\textwidth]{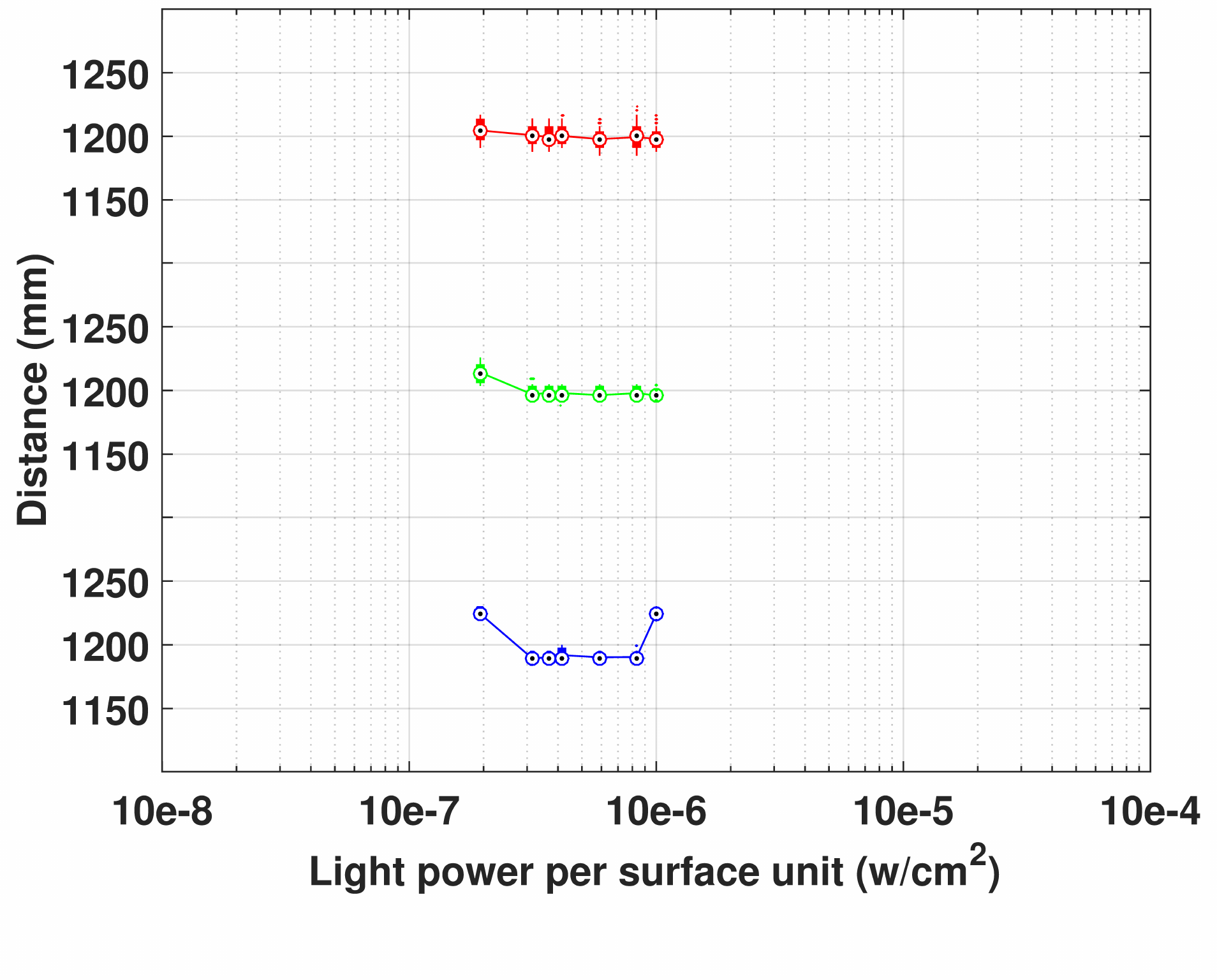}
  \includegraphics[trim = 7mm 50mm 9mm 91mm,clip,width=.49\textwidth]{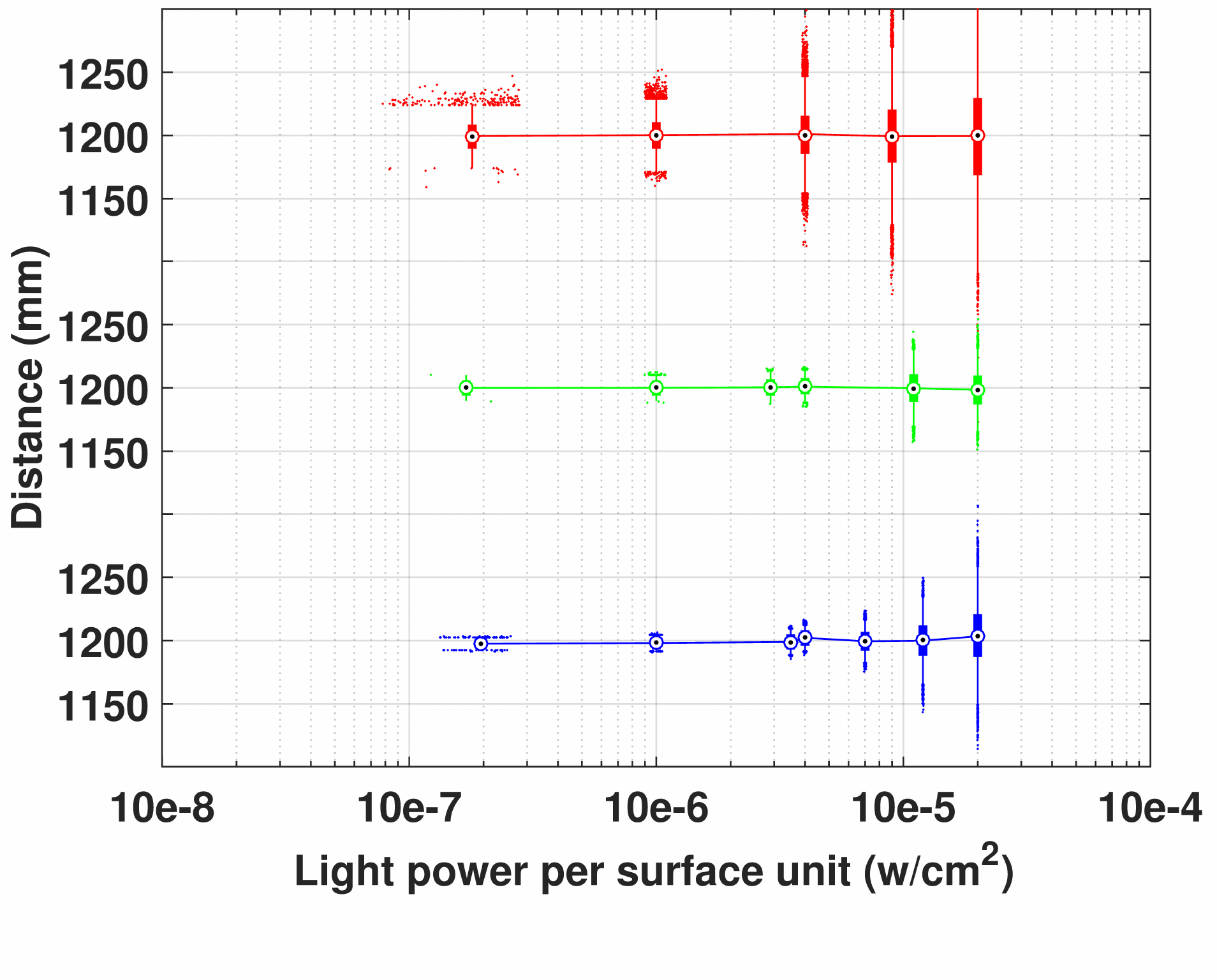}

  \caption{Kinect Comparison for Ambient Background Light. The SD
    (top row) and distance
    statistics (bottom row) for the center, intermediate and corner
    pixel for \kinsl\ (left) and the \kintof\ (right).}
  \label{fig:interference:kComparisons}
\end{figure}

\paragraph{Evaluation and Results}
We apply a variance analysis for the $K=200$ frames of range values
$D_i(u,v),\,i=1,\ldots,K$ delivered by each camera for each of the
pixels $(u,v)$ (center, intermediate, corner) by computing the
\emph{Standard Deviation (SD)} over time
\begin{equation}
  \text{SD} = 
  \sqrt{\frac1{K}\sum\limits_{i=1}^K \left( D_{i}(u,v) - D^{\text{mean}}(u,v)
    \right)^2},\quad
  D^{\text{mean}}(u,v) = \frac1{K}\sum\limits_{i=1}^K D_{i}(u,v),
  \label{eq:SD}
\end{equation}
and plot this as function over the ambient light intensity; see
Fig.~\ref{fig:interference:kComparisons}, top. Additionally,
Fig.~\ref{fig:interference:kComparisons}, bottom, shows explicit
distance values including box plots as function over ambient light.

It can be observed, that the \kinsl\ is not able to handle background
light beyond $1\mu$W, whereas the \kintof\ delivers range data
throughout the full range of ambient light applied in the experiment.

The \kinsl\ delivers more robust depth values than the
\kintof\ throughout the ambient background light range where valid
data is delivered. All observed pixels are below $6$mm SD and the
max. variation from the median is $25$mm for the corner pixel. The SD
and the box plots show, that for the \kinsl\ the depth variation is
hardly effected by the ambient light, as long as valid range data is
delivered. The plots for the different pixels show, that the variation
increases for pixels closer to the image vicinity.

The \kintof, at the other hand, shows the expected raise in the depth
variation for increasing ambient light due to a reduced SNR. Whereas
the center and the intermediate pixel show similar SD below
$6\mu$W as the \kinsl, i.e. below $4$mm, the
box plots reveal a larger number of outliers compared to the
\kinsl. However, the \kintof's corner pixel delivers worse SD and
quantile statistics than the one for the \kinsl. In the range beyond
$10\mu$W ambient light, the variation increases
to some $22,12$ and $42$mm for the center, intermediate and corner
pixel, respectively. The effect that the center pixel gets worse than
the intermediate pixel may be explained by oversaturation effects
solely due to the active illumination of the \kintof.

\begin{figure}[b!]
  \centering 
  \includegraphics[width=.4\textwidth]{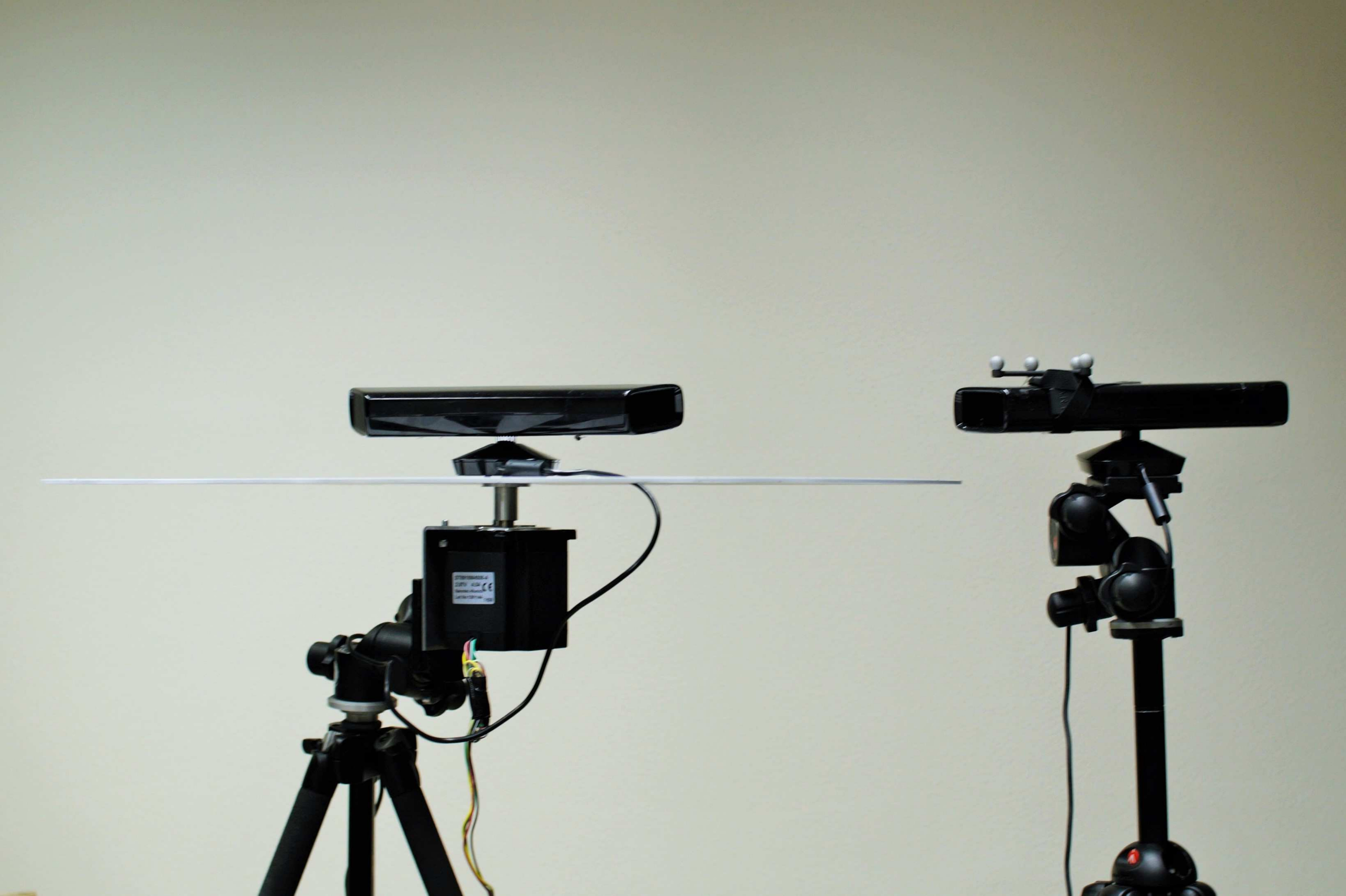}
 
  \caption{Multi-device interference experiment setup }
  \label{fig:interference:setup}
\end{figure}

\begin{figure}[t!]
  \centering

  \includegraphics[width=.49\textwidth]{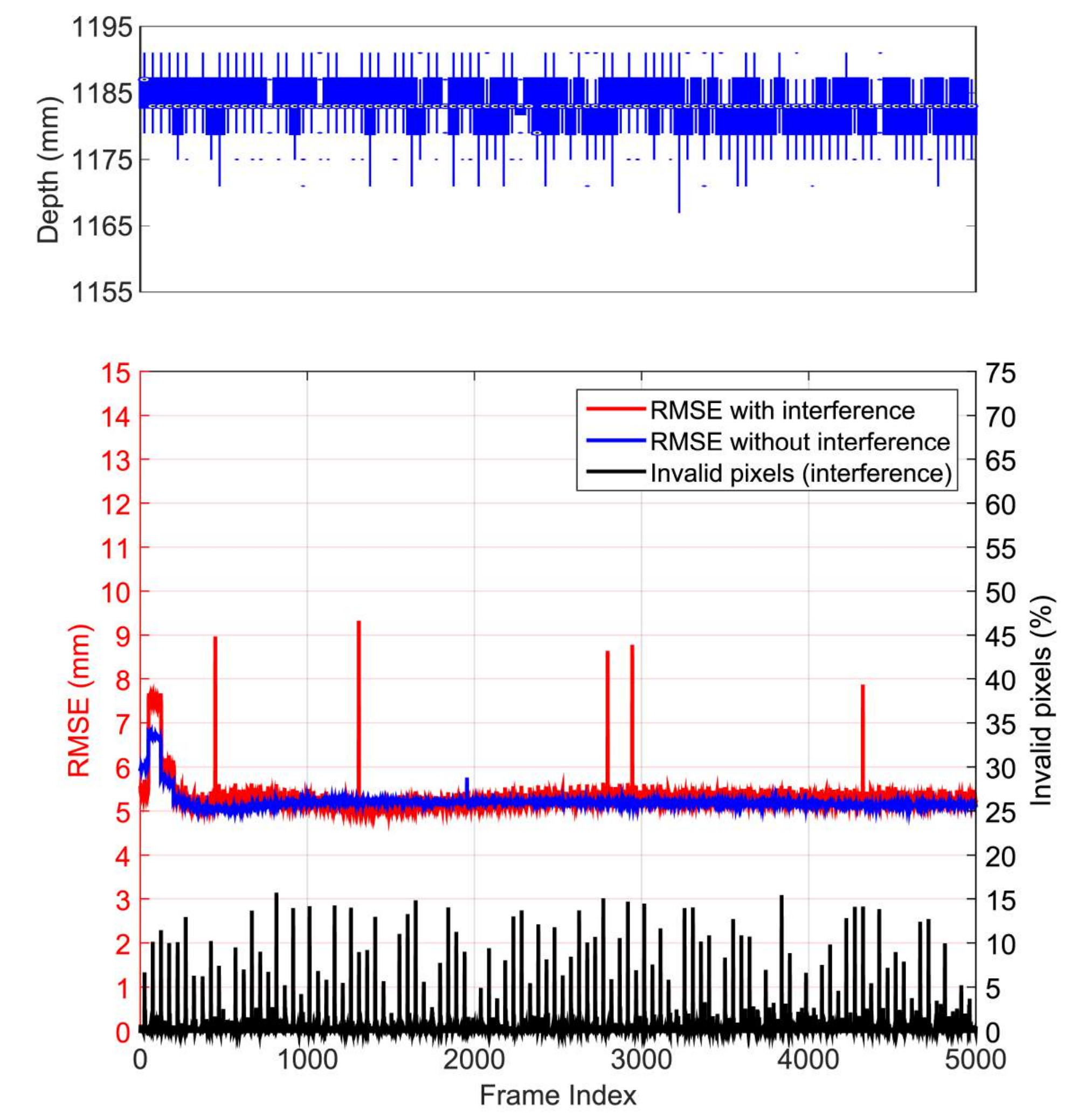}
  \includegraphics[width=.49\textwidth]{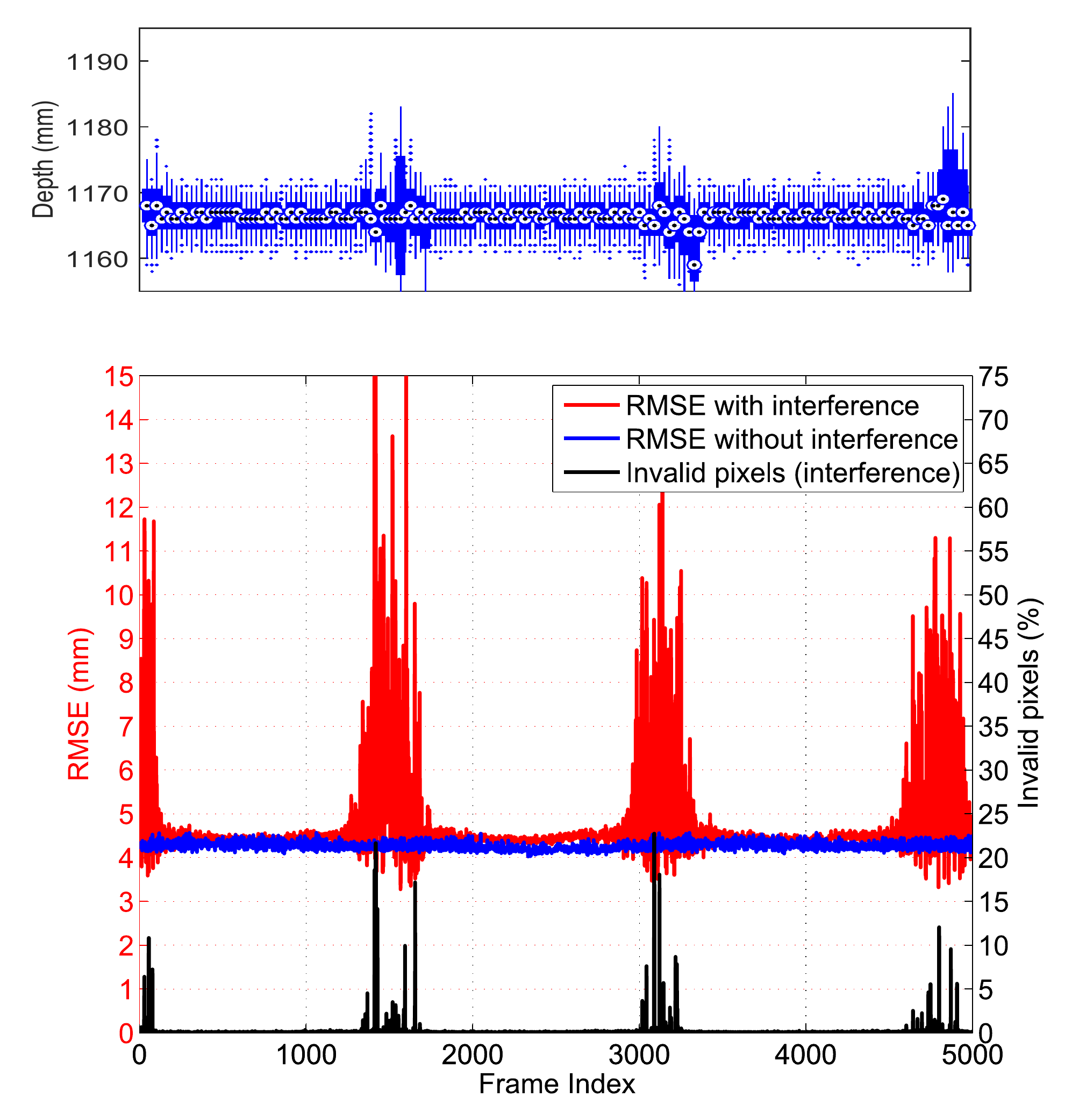}

  \caption{Multi Device Interference: Error for the static scene for
    the \kinsl\ with moving interference device (left) and for the
    static \kintof\ (right) as boxplot statistics for the interference
    situation (top row) and RMSE plot with and without interference
    including invalid pixel counts for the interference setup (bottom
    row).}
  \label{fig:MultiDevice:Static}
\end{figure}

\subsection{Multi-Device Interference}
\label{sec:results:interference}

\paragraph{Goal} This experiment addresses the problem arising from
the interference of the active illumination between several
Kinect-cameras of the same type, when running them in
parallel. Primarily, we want to evaluate the influence of the
interference on the range measurement. Secondarily, we want to gain
some insight into the temporal and spatial distribution of the artifacts.

Note, that in contrast to other ToF-cameras (see
Sec.~\ref{sec:principle.err}) we are not able to modify the modulation
frequencies for the \kintof\ in order to reduce or suppress
multi-device interference artifacts.

\paragraph{Experimental Setup}
The general idea of the experiment is to acquire an approximately
diffuse white planar wall, adding a second Kinect device of the same
kind as interference over a longer period of time. As the \kinsl\ uses
a static structured light pattern, a fully static setup may not
capture the overall interference. As circumventing interference for
the \kinsl\ may not always be possible with a ``Shake'n'Sense''-like
approach~\cite{butlerSNS12}, we investigate the influence of the
camera poses of the two devices on the interference. Thus, for the
\kinsl\ setup, we mount the interfering device on a turntable and
rotate it $\pm10^\circ$ about the vertical axis with $1$RPM in order
to get a variation of the overlay of the SL-patterns of the two
devices. The angular speed is low enough to prevent any motion
artifacts. We also investigated different inter-device distances, but
the resulting impact on the interference was comparable. The
\kintof\ setup the interfering device is always static.   The distance
between the wall and the Kinect was set to $1.2$m and the distance
between the devices is $0.5$m; see
Fig.~\ref{fig:interference:setup}. We do not take the exact orientation of the measuring and the interference devices into account, the measuring and the interfering device, but
both devices observe approximately the same region on the wall.

\begin{figure}[b!]
  \centering 
  \includegraphics[trim = 20mm 10mm 4.7mm 2mm,clip,height=.11\textheight]{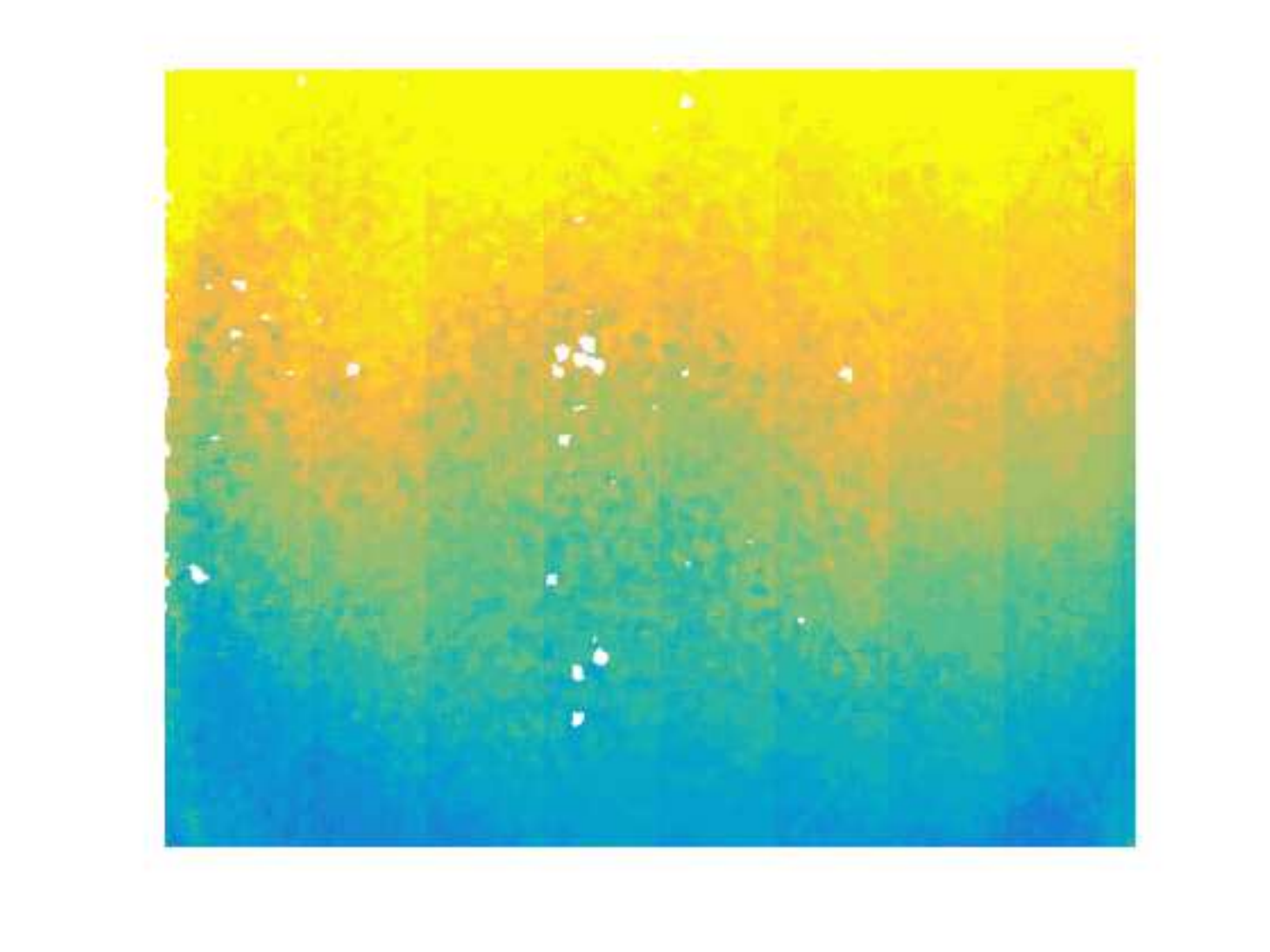}
  \includegraphics[trim = 20mm 10mm 4.7mm 2mm,clip,height=.11\textheight]{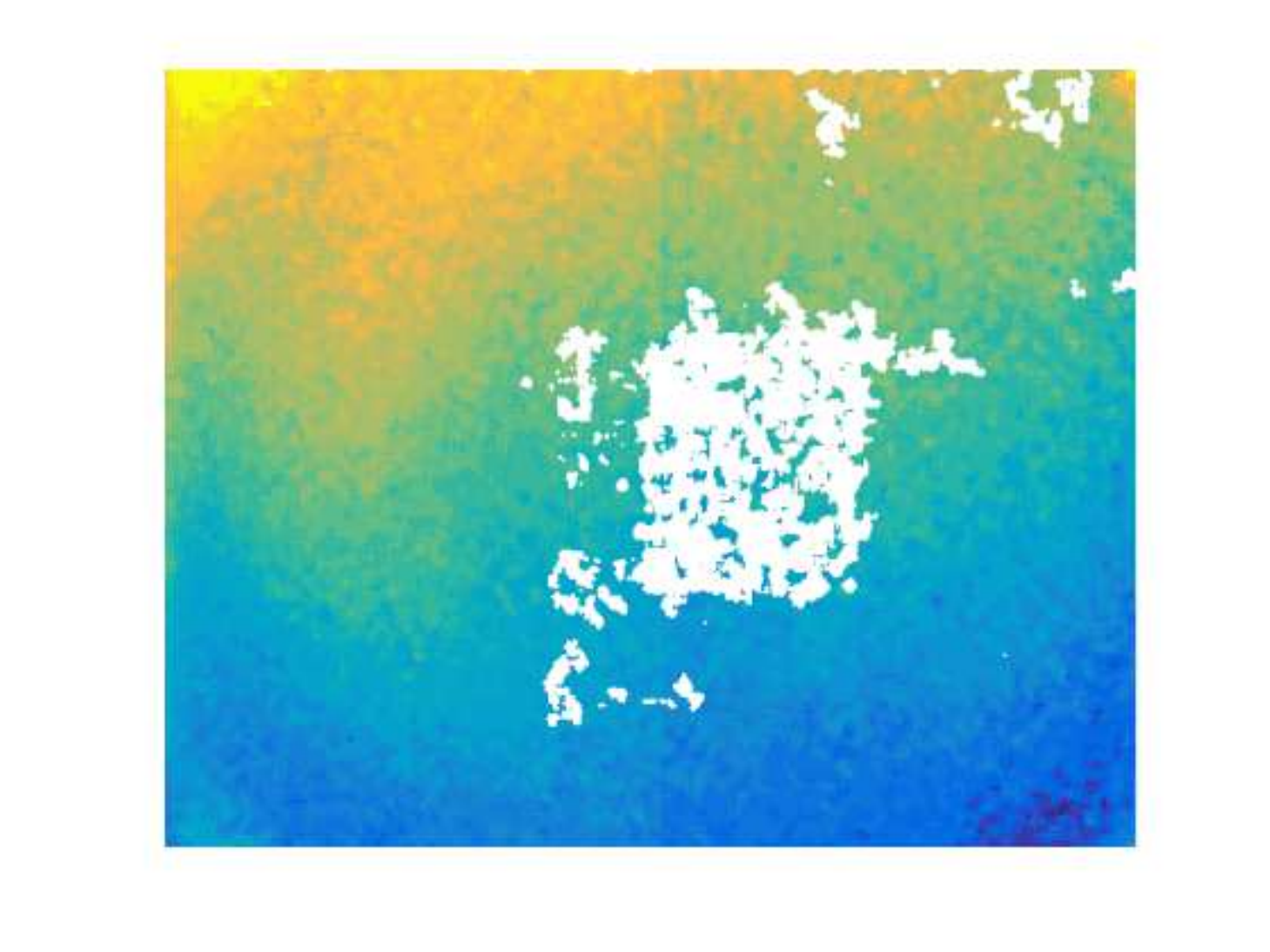}
  \includegraphics[trim = 20mm 10mm 4.7mm 2mm,clip,height=.11\textheight]{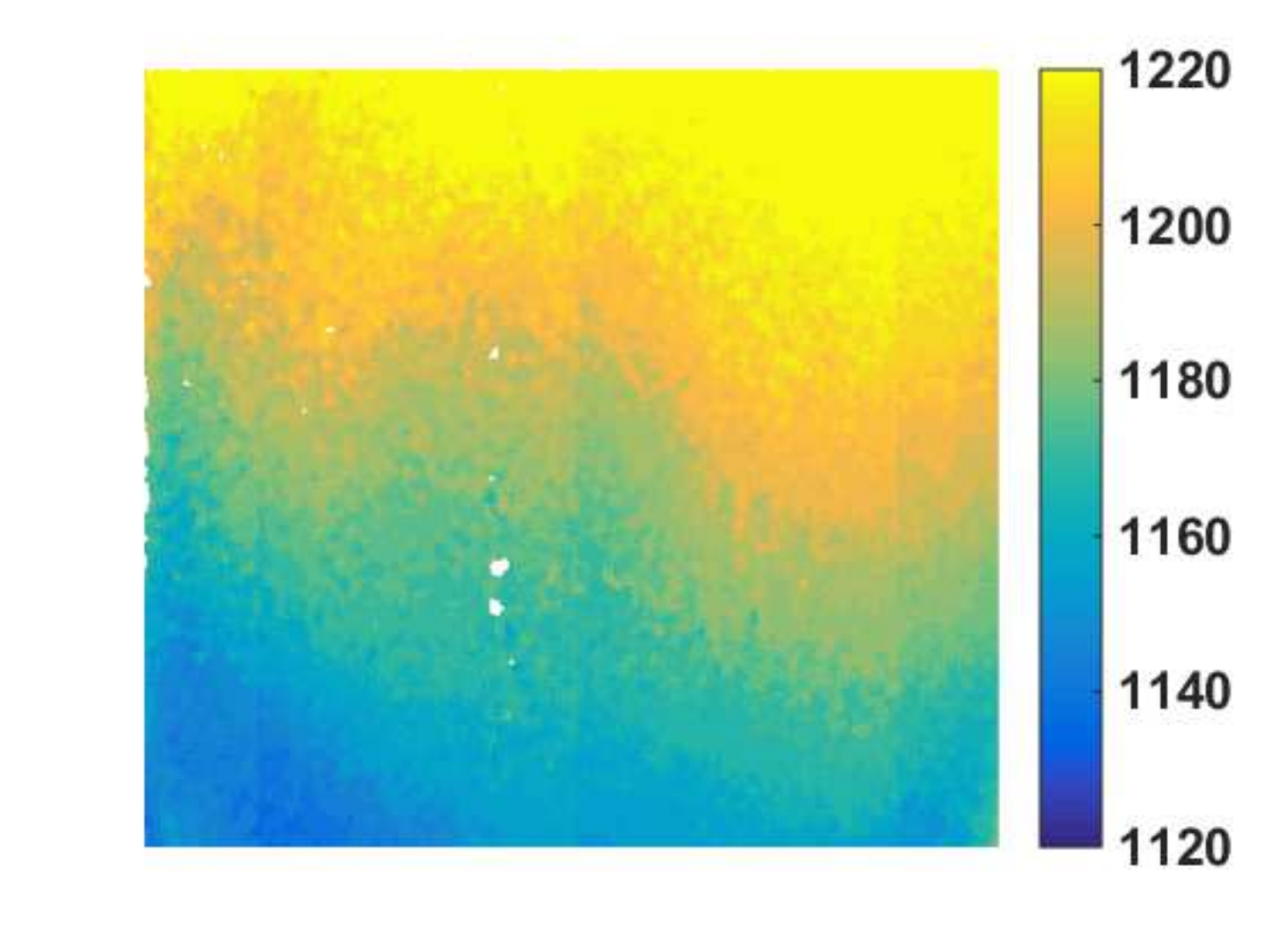}
  \includegraphics[trim = 20mm 25mm 5mm 2mm,clip,  height=.11\textheight]{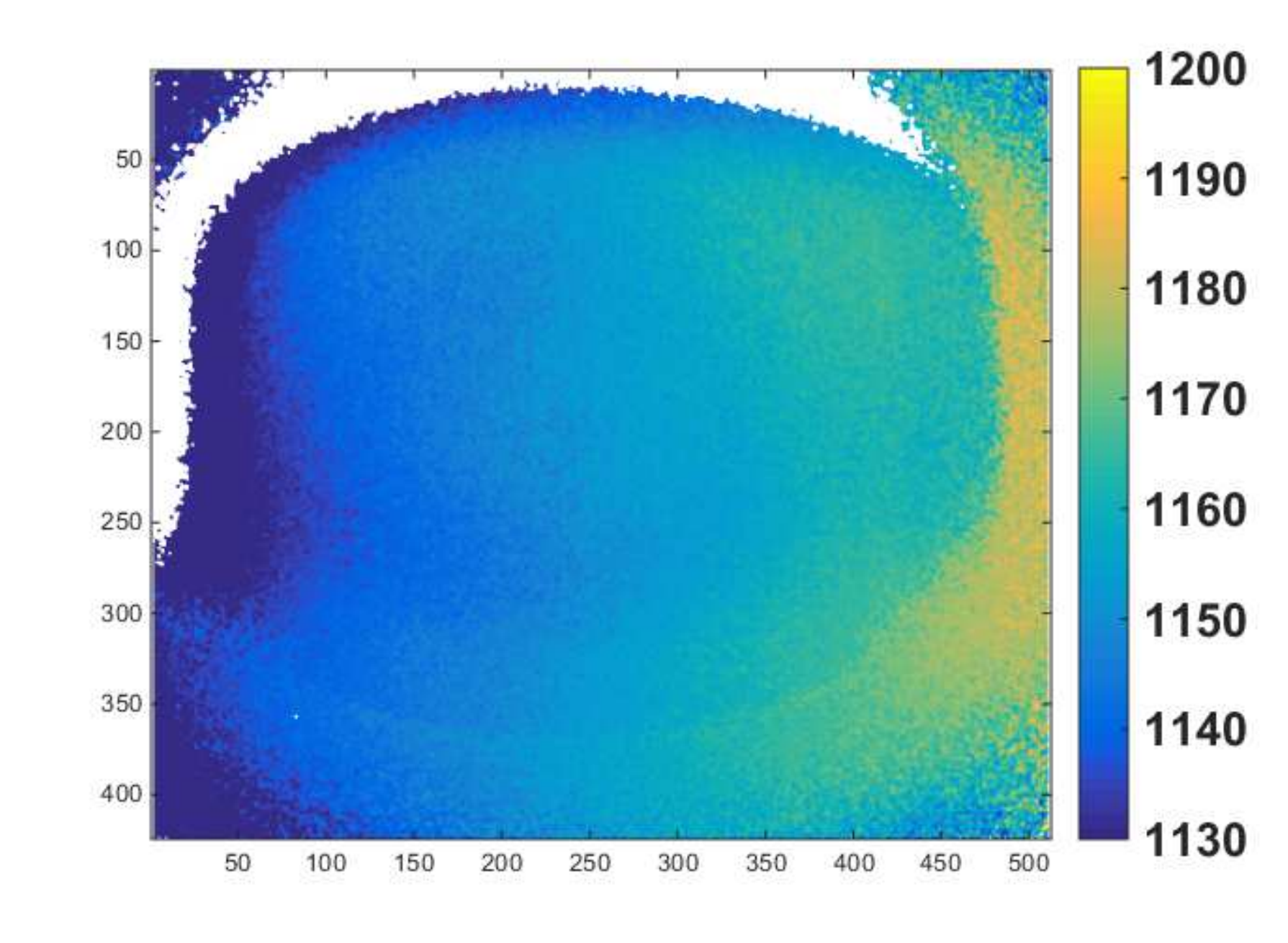} \\
  \caption{Sample range images for the multi-device interference
    setup: The \kinsl\ range image for the initial pose of the
    interfering device, left, and for two pose with a high invalid
    pixels count, mid-left, and a high RMSE, mid-right. A
    \kintof\ range image with high invalid pixel count and high RMSE, right.}
  \label{fig:interference:depth}
\end{figure}

\paragraph{Evaluation and Results}
In order to account for a potential misalignment of the Kinect towards
the wall, we use a RANSAC plane fit to the range data with inactive
interference device. The per-pixel ranges values, deduced from this
plane fit, are considered as reference distances
$D^{\text{ref}}(u,v)$. We compute the deviation for
each frame $D_i$ in respect to the reference distance as
Root-Mean-Square Error (RMSE), i.e.
\begin{equation}
  \text{RMSE} = \sqrt{\frac1{m \cdot n}{\sum\limits_{u=1}^n
      \sum\limits_{v=1}^m \left( D_i(u,v) - D^{\text{ref}}(u,v)
      \right)^2 }}
  \label{eq:RMSE}
\end{equation}
where $n,m$ represent the width and height of the averaged range
image, respectively.

As can be seen in Fig.~\ref{fig:MultiDevice:Static}, the active
frequency pattern of the \kintof\ has a stronger interference than the
structured light pattern for the \kinsl\ for most poses of the
interfering camera. On average, the \kinsl\ shows little interference
effect (RMSE $\textless 5.6$mm), beside some very prominent poses
(RMSE up to $<9.4$mm). Fig.~\ref{fig:interference:depth}, mid-right,
shows a sample range image with a high RMSE. The \kintof\ camera shows
low interference for the majority of the frames (RMSE: $<5$mm), but
extreme interference errors for some $25$\% of the frames (RMSE up to
$19.3$mm) that occur in a sequence which has a nearly constant
repetition rate. This behavior is most likely due to the asynchronous
operation of the two devices. A signal drift over time between the
signals generated in both devices would lead to a repetitive
interference pattern as the one observed.  The range statistics
represented in Fig.~\ref{fig:MultiDevice:Static}, top, shows that the
median in \kinsl\ is not altered by interference, which is mainly due
to the strong quantization applied in the disparity maps. In phases of
maximum interference, the \kintof\ delivers increased drift of the
median, up to $5$mm, and a stronger variation.

Regarding the invalid pixels, the \kinsl\ nearly always delivers
invalid pixels. For the initial pose, we find some $1.5$\% invalid
pixels; see Fig.~\ref{fig:interference:depth}, left. While changing
the pose of the interfering device, we find up to $16.3$\% invalid
pixels; see Fig.~\ref{fig:interference:depth}, mid-left.  The
\kintof\ does not deliver invalid pixels in the non-interfered
periods, but in the interference periods up to $22.7$\% of invalid
pixels have been observed; see Fig.~\ref{fig:interference:depth},
right.

We want to point out that we always observe strong variations within
the first $400$ frames, i.e. the first $13$ seconds after starting the
acquisition with the \kinsl. In this experiment we have an increased
RMSE of up to $6.7$mm without interference and $7.7$mm with
interference.  Therefore, it is advisable to not use this initial
sequence captured with the \kinsl.

\subsection{Device Warm-Up}
\label{sec:results:temperature}

\paragraph{Goal} This test scenario is designed to evaluate the 
drift in range values during the warm-up in standard operation,
i.e. the stability of the range measurements of both Kinects with
respect to the operating time.

\paragraph{Experimental Setup}
We accommodate the device in a room with a constant temperature of
$21^\circ$C which is actively controlled by an air conditioning system
with a variance below $0.1^\circ$C. We start to operate the device
measuring a planar wall at a distance of $1200$mm.  We acquire $200$
frames in a row and drop frames for $15$s ($450$ frames) afterwards
and repeat this until a total time of $120$ minutes. During the
acquisition a digital thermometer (precision $\pm0.1^\circ$C) records
the temperature inside the Kinect devices. The temperature in the
device interior is measured with a flexible sensor tip inserted
through the ventilation holes. Thus, the devices remained intact in
order to keep the original temperature dissipation system.

\begin{figure}[t!]
  \centering

    \includegraphics[width=.99\textwidth]{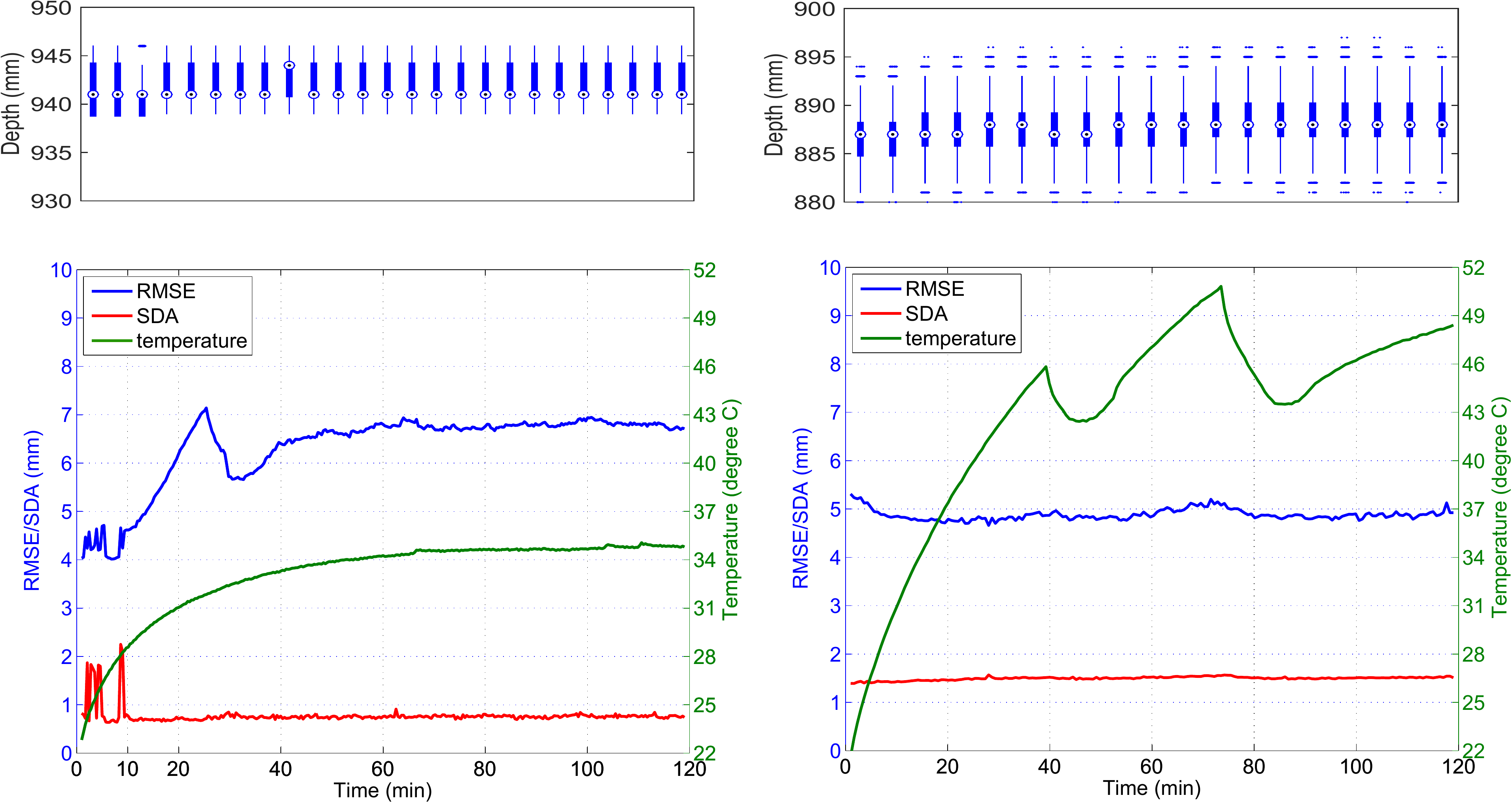}

  \caption{Device Warm Up: Box plot Error mean and temperature versus warm-up
    time (bottom row) for \kinsl\ (left) and \kintof\ (right).}
  \label{fig:warmup:results}
\end{figure}

\begin{figure}[t!]
  \centering
    \includegraphics[trim = 1mm 5mm 5mm 2mm,clip,width=.45\textwidth]{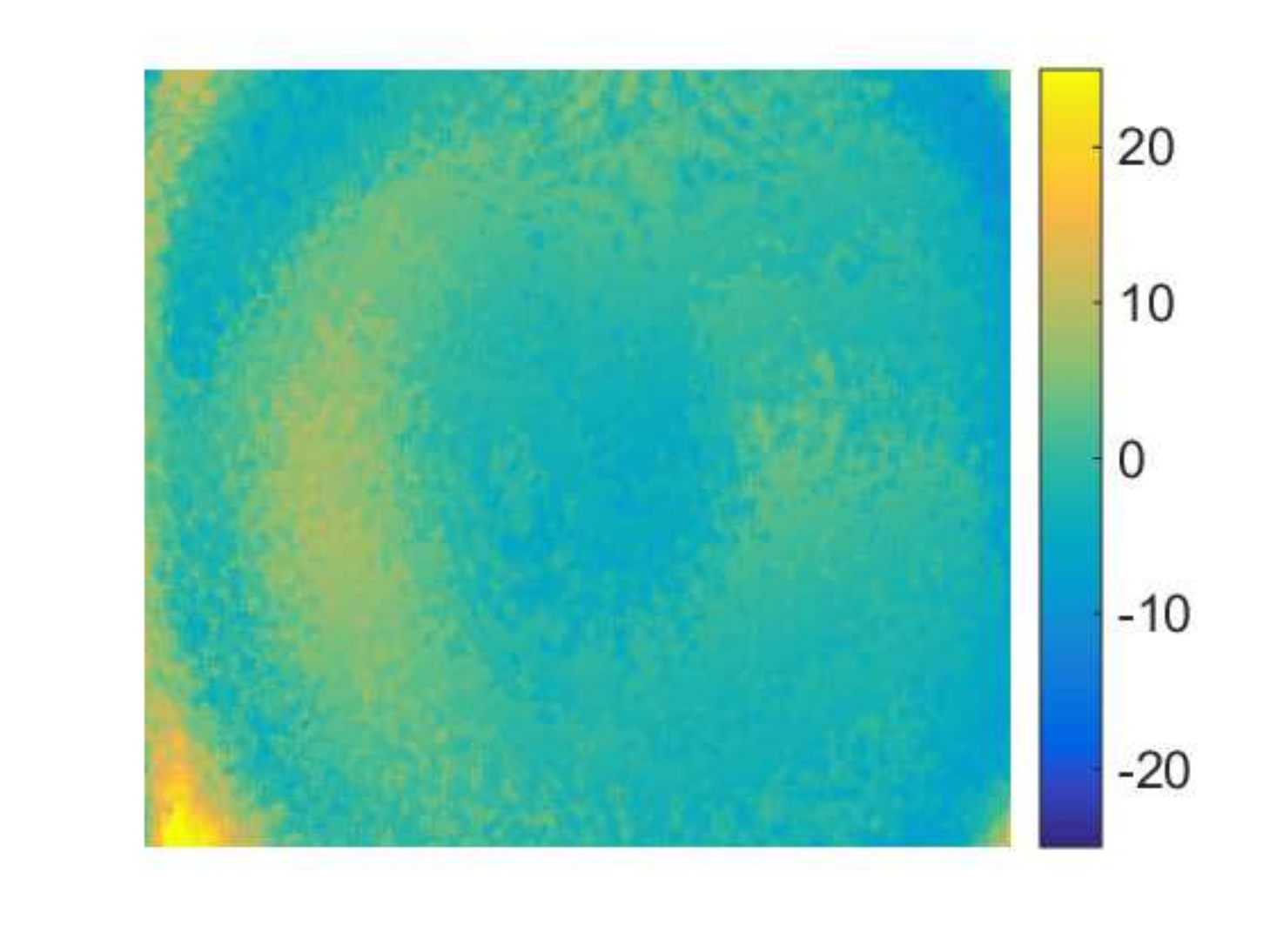}
    \includegraphics[trim = 1mm 5mm 5mm 2mm,clip,width=.45\textwidth]{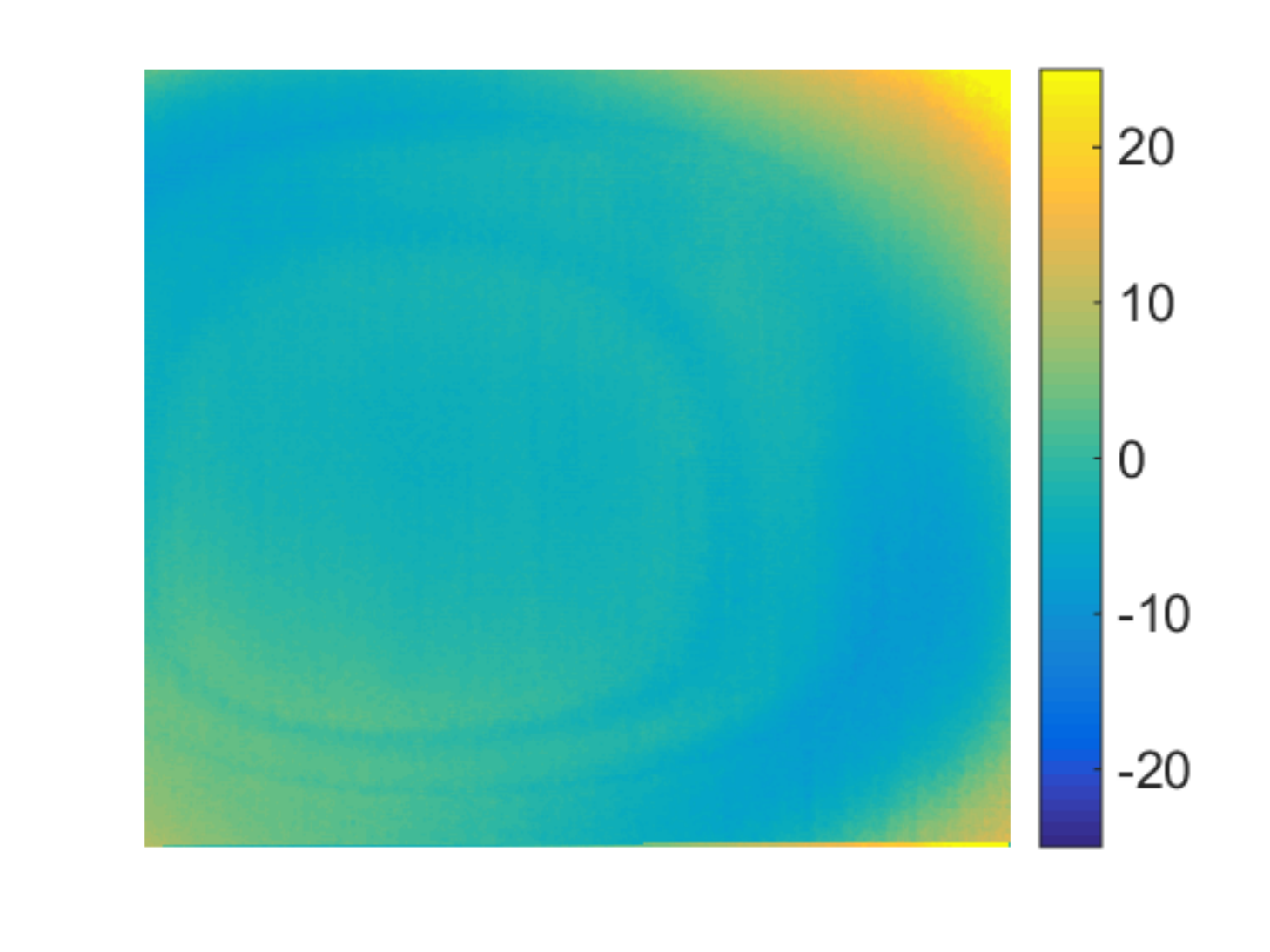}
  \caption{Device Warm Up: depth error at minute 60 for \kinsl\ (left)
    and \kintof\ (right).}
  \label{fig:warmup:range-image}
\end{figure}

\paragraph{Evaluation and Results}
As the variation in the range data is smaller for the cold device than
for the warm device, we make a RANSAC fit to the averaged first steady
sequence of $200$ frames, resulting in a reference depth image
$D^{\text{ref}}$. However, for \kinsl the first steady sequence of 200
frames was captured after $10$ minutes, as we observe a very strong
variation in this initial range of measurements. Nevertheless, since the RANSAC is applied to the whole frame there might be some bias to the reference frame. We calculate the RMSE
for the average of all $200$ frames $D^{\text{mean}}$ in a frame
sequence with respect to the fitted plane as
\begin{equation}
  \text{RMSE} = \sqrt{\frac1{m \cdot n}\sum\limits_{u=1}^m
    \sum\limits_{v=1}^n \left( D^{\text{mean}}(u,v) - D^{\text{ref}}(u,v)
    \right)^2}.
  \label{eq:RMSE-mean}
\end{equation}

Furthermore, we calculate the per-pixel \emph{standard deviation average}
(SDA) for each sequence of $K=200$ frames $D_i$.
\begin{equation}
  \text{SDA} = \frac 1{m \cdot n} \sum\limits_{u=1}^m \sum\limits_{v=1}^n
  \sqrt{\frac1{K}\sum\limits_{i=1}^K \left( D_{i}(u,v) - D^{\text{mean}}(u,v)
    \right)^2}.
  \label{eq:SDA}
\end{equation}

The results for the device warm-up test are shown in
Fig.~\ref{fig:warmup:results}. The fluctuation in the temperature of
the \kintof\ is due to the cooling system, that gets activated and
deactivated depending on the system temperature. For the \kinsl\ there
is only a small temperature difference of 11$^\circ$C after 120
minutes. The results show that \kintof\ has in general less error than
the \kinsl\ and SDA and RMSE are nearly constant over time. The
\kinsl\ has strong error and variation fluctuations in the first
$10$ min of the warm-up phase. After the device has reached its
operation temperature, the distance SDA stays within $1$mm, which is
slightly better than for the \kintof\ which has a distance SDA of
$1.5$mm. However, the distance error is higher for the \kinsl\ (RMSE
$<7.1$mm) than for the \kintof\ (RMSE $<5.3$mm). The distance box
plots in Fig.~\ref{fig:warmup:results}, top, show less variation for
the \kinsl\ than for the \kintof. However, we again point out, that
the homogeneous appearance of the box-plots for the \kinsl\ partially
result from the heavy quantization applied in this device.

Fig.~\ref{fig:warmup:range-image} shows the depth error in respect to
the fitted plane in absolute signed values. The depth images are taken
at minute 60, when both devices are at a stable temperature. As it can
be seen in the depth images the \kintof\ delivers smoother results
with less out of plane errors compared to \kinsl, which is consistent with the RMSE values at minute 60.

\subsection{Rail Depth Tracking}
\label{sec:results:rail}

\paragraph{Goal} This test scenario primarily addresses the quality of the range
data in respect with ground truth distances for a planar wall. The
test involves the linearity including planarity tests as well as the
intensity related error. The latter applies only for the
\kintof\ camera. A secondary goal is to give some clue about the
dependence of the error from the pixel location, therefore we evaluate
the error at a few different image locations.

\paragraph{Experimental Setup}
The setup comprises a motorized linear rail mounted perpendicular to a
white wall, which measures distances between $0.5$m and $5$m at a
step-size of $2$cm. The camera is mounted on the carriage of the rail
facing perpendicular to the wall. As the wall does not cover the full
range image for farther distances, we evaluate planarity and linearity
of the camera only within a region-of-interest including pixels lying
on the white flat wall in the full distance range. The pixel region of
interest for \kinsl\ is (1,1), (630,480) and (74,4), (502,416) for
\kintof. Furthermore, we observe some pixels along a line-of-interest
from the image center to the top-left corner, which are always
covering the wall. We acquire $200$~frames for each distance. For the
evaluation of the intensity-related error, the acquisition is repeated
with a $5\times6$ checkerboard attached to the wall. The checkerboard
consists of $10$ gray-level rectangles on white background, where the
gray-level degrades from $100$\% to $0$\% black. The checkerboard has
been printed using a standard laser printer which delivers
sufficiently proportional reflectivity in the visual and the NIR
range.

In order to re-project the range values into 3D-space, we first
estimated the camera intrinsics using the well known photometric
calibration technique from \cite{zhang00Calib}. Similar approaches
have been applied to the \kinsl\ \cite{macknojia2013KINSLCalib} (where
the laser beam is obstructed and an incandescent lamp is used to
highlight the checkerboard in order to acquire a reliable NIR image of
the calibration rig) and for ToF-cameras~\cite{lindnerLAD06}.

\subsubsection{Linearity -- Evaluation and Results}
\label{sec:results:rail.linearity}

\begin{figure}[t!]
  \centering

  \includegraphics[trim = 13mm 65mm 18mm 73mm,clip,width=.49\textwidth]{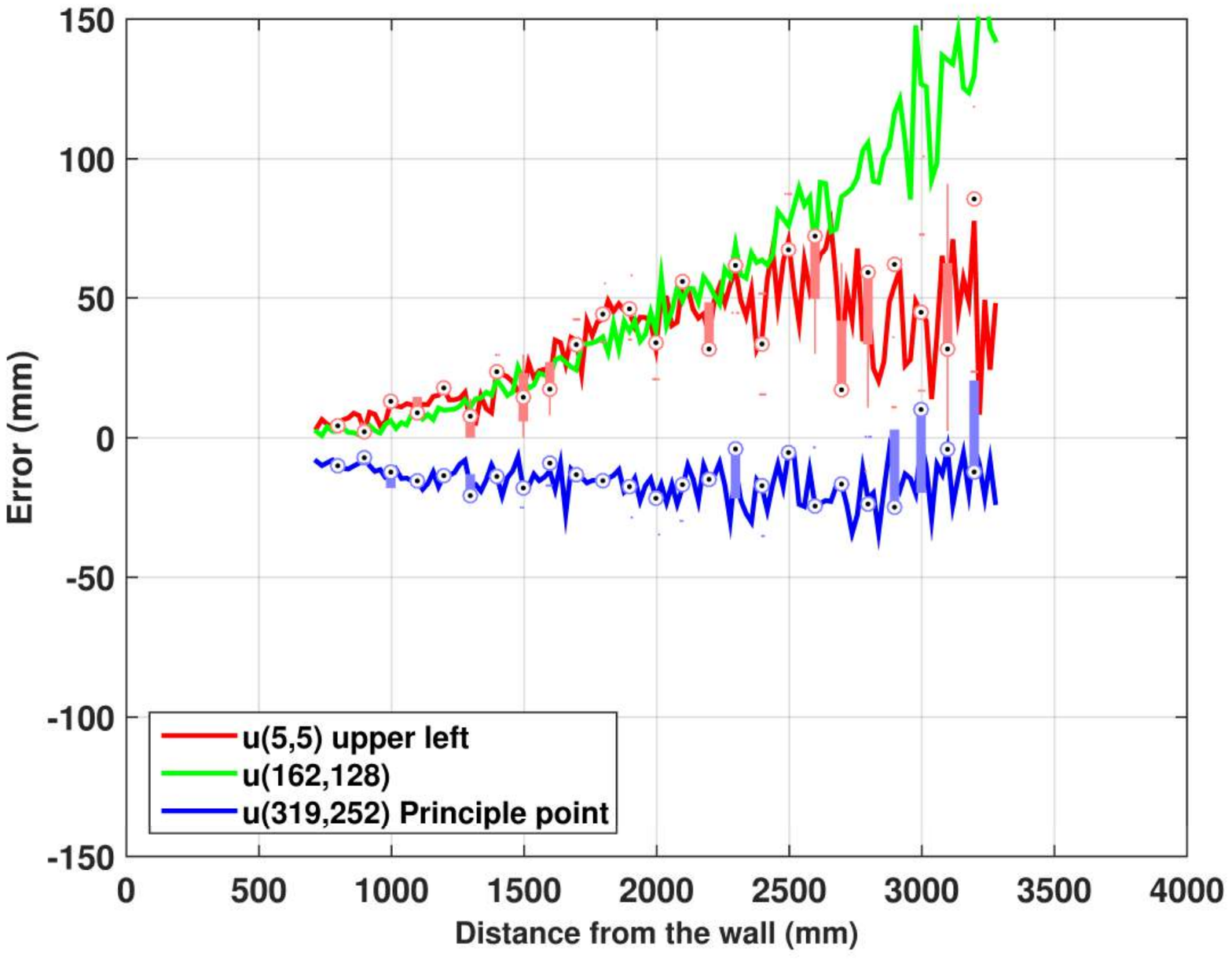}
  \includegraphics[trim = 13mm 65mm 18mm 73mm,clip,width=.49\textwidth]{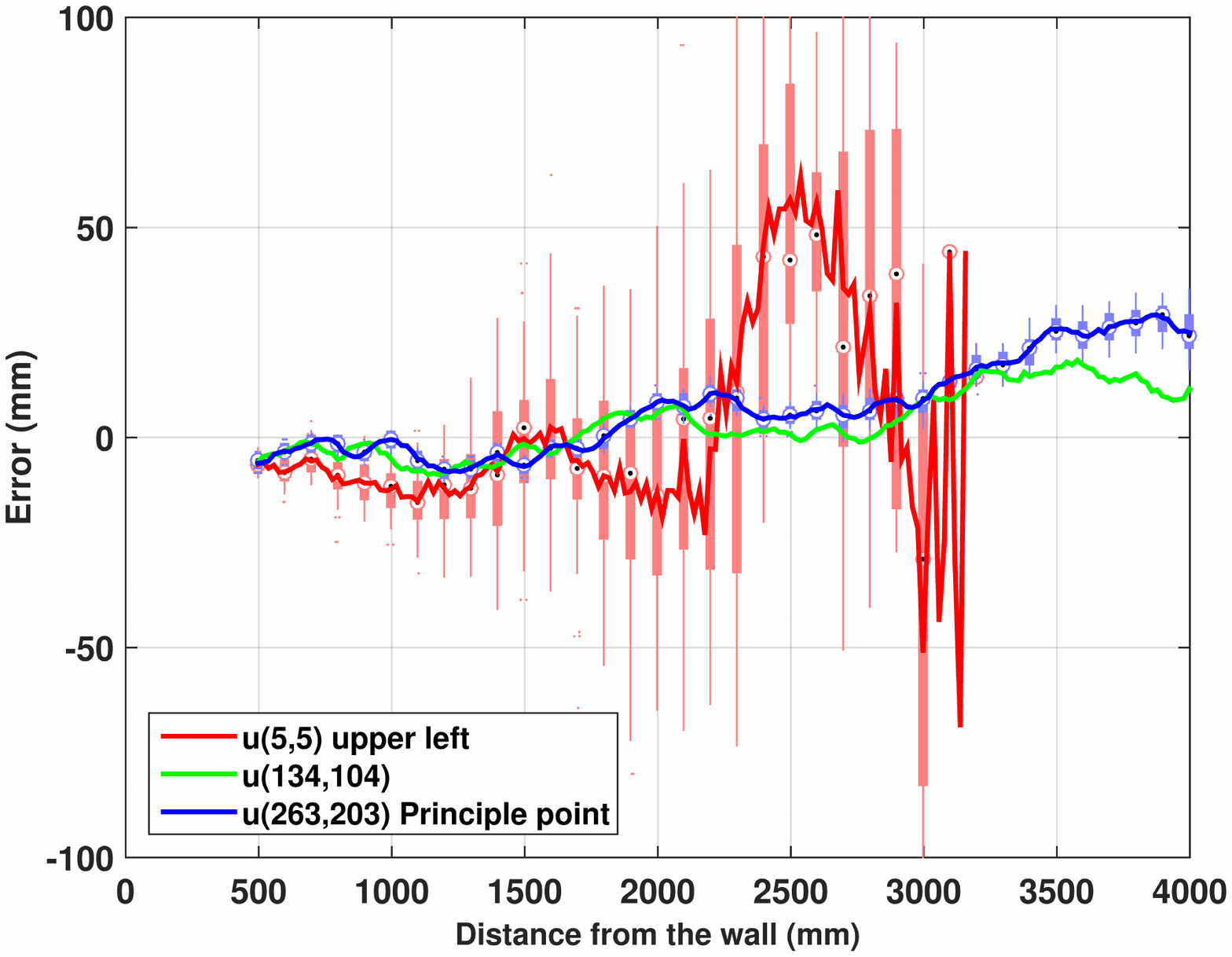}

  \includegraphics[trim = 13mm 65mm 18mm 71mm,clip,width=.49\textwidth]{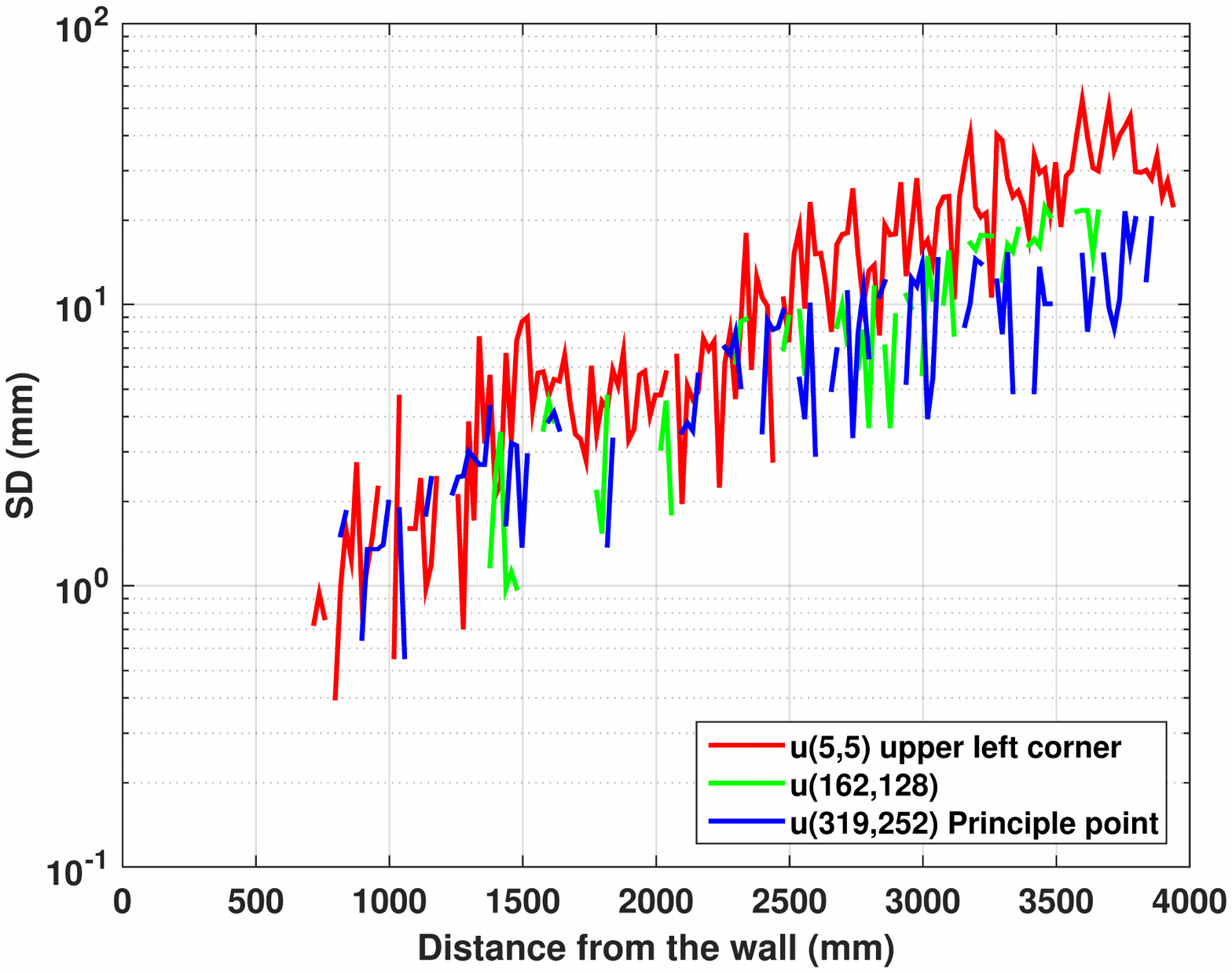}
  \includegraphics[trim = 13mm 65mm 18mm 71mm,clip,width=.49\textwidth]{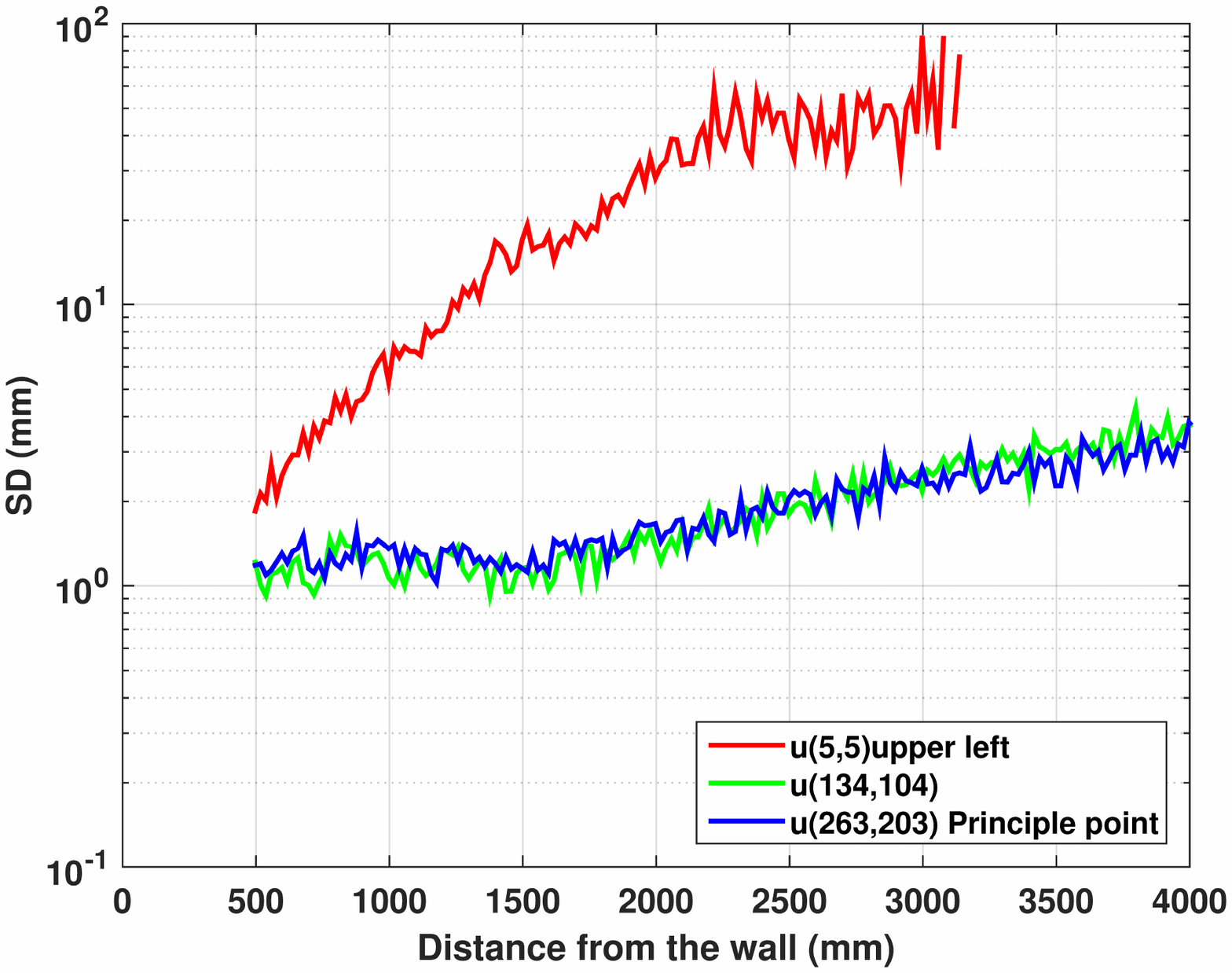}

  \caption{Linearity Error for four points along the line of interest:
    Distance error and SD of for \kinsl\ (left) and
    \kintof\ (right). The corner pixel of the \kintof\ delivers
    invalid depth after about $3100$mm, therefore no depth values are
    given for this range.}
  \label{fig:linearity:results}
\end{figure}

The evaluation of the linearity requires a proper measurement of the
ground truth distances for the range images acquired with the rail
system. As a perfect orthogonal alignment of the camera towards the
wall can not be guaranteed, we propose to bypass this problem using
photometric methods.  Having a complete lens calibration of both
camera systems (i.e. depth and color intrinsic and distortion
parameters) and the extrinsic transformation between the High-Res
color camera and the depth camera, the precise 3D camera position of
the depth camera can be obtained using a simple black-white
checkerboard reference fixed to the planar wall and which we acquire
at $10$ different rail positions. The corresponding 3D positions of
the depth camera relative to the reference wall is done using the
standard method  \cite{zhang00Calib}. A 3D
line was fitted to these 3D positions using a RANSAC statistical
approach which gives the robust orientation of the linear
rail. Finally, knowing the precise displacement of each measurement of
the linear rail (we use a $2$cm step size), the 3D position of the
camera can robustly be estimated and thus a precise ground-truth of the
wall be generated using the lens parameters of the depth camera.
Having the ground truth distance $D_d^{gt}(u,v)$ for a given
camera-to-wall distance $d$ for each pixel, we calculate the
\emph{signed error (SE)} for the average of all depth measurements
$D_d^{\text{mean}}(u,v)$ at the rail system, thus suppressing sensor
noise
\begin{equation}
  \text{SE}_d = \frac1{k \cdot l}{\sum\limits_{u=1}^k \sum\limits_{v=1}^l \left(
    D_d^{\text{mean}}(u,v) - D_d^{gt}(u,v) \right) },
  \label{eq:SE}
\end{equation}
within the region-of-interest consisting of $k\times l$ pixel. For
some pixels along the line-of-interest we also evaluate the individually signed
linearity errors.

Furthermore we calculate the variance for each pixel in the
region-of-interest using the $200$~frames $D_{i,d}$ taken for each
camera-to-wall distance $d$ in order to retrieve the standard
deviation average SDA according to Eq.~\ref{eq:SDA}.

Fig.~\ref{fig:linearity:results}, top, shows the signed linearity
errors for both Kinects for the selected pixels with some box plots
superimposed. As can be seen in Fig.~\ref{fig:linearity:results}
right, the \kintof\ delivers more precise range data than the \kinsl,
if the corner pixel is not taken into account. In the proposed work
range of the \kinsl\ below $3$m the error lies in the range of
$[-34,-1.5]$mm for the best (central) pixel and of $[2.5,76]$mm for
the worst (peripheral) pixel. The SD for the \kinsl\ below $3$m is
very similar for all pixels and is below $30$mm. Above $3$m the
distance error of the \kinsl\ strongly increases for peripheral
pixels. Even though there seem to be some fluctuations in the distance
error for the \kintof, this effect is much smaller and much less
regular than the ``wiggling''-error observed so far for
ToF-cameras~\cite{lindnerLAD06, lindnerTOF10}.  For pixels not in the
extreme periphery the absolute per-pixel distance error and the SD
lies in the range of $[129,-34]$mm and $[0.4,14]$mm, respectively. For
the corner pixel the SD range increases to $[0.4,28]$mm.

\subsubsection{Planarity -- Evaluation and Results}
\label{sec:results:rail.planarity}

The region-of-interest in each range image acquired using the rail
lies on a planar wall, so the resulting range measurements should ideally result in a plane. Similar to Khoshelham and
Elberink~\cite{khoshelhamAAR12}, we apply a RANSAC plane fitting
method to avoid outliers and calculate the standard deviation of the
points from the fitted plane as planarity error.

\begin{figure}[t!]
  \centering
  \includegraphics[trim = 14mm 68mm 18mm 74mm,clip,width=.55\textwidth] {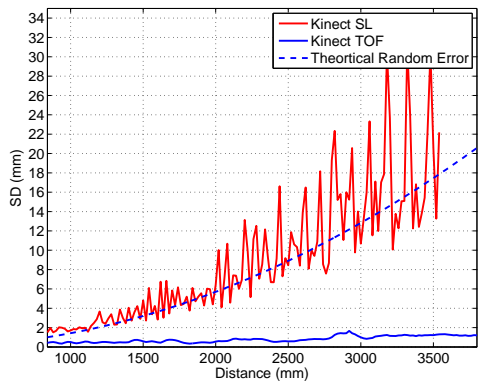}

  \caption{Planarity Error. The standard deviation of the pixels
    within the region-of-interest for the \kinsl\ (red) and the
    \kintof\ (blue). Additionally, the theoretical random error
    deduced by Khoshelham and Elberink~\cite{khoshelhamAAR12} for the 
    \kinsl\ is shown.}
  \label{fig:Planarity:Results}
\end{figure}

Fig.~\ref{fig:Planarity:Results} shows the planarity error as SD
for both Kinect and the theoretical random error deduced by Khoshelham
and Elberink~\cite{khoshelhamAAR12} for the \kinsl. The
\kinsl\ delivers much stronger out-of-plane errors than the \kintof,
which stays below $1.65$mm for the whole range of $4$m. The curve for
the \kinsl\ is roughly within the expected range. Compared to
Khoshelham and Elberink~\cite{khoshelhamAAR12} we observe an
additional fluctuation which can be explained by the decreasing depth
resolution of the \kinsl\ which leads to a significant depth
quantization for increasing distances. 

\begin{figure}[b!]
  \centering
  \includegraphics[trim = 0mm 9mm 0mm 0mm,clip,width=.85\textwidth]{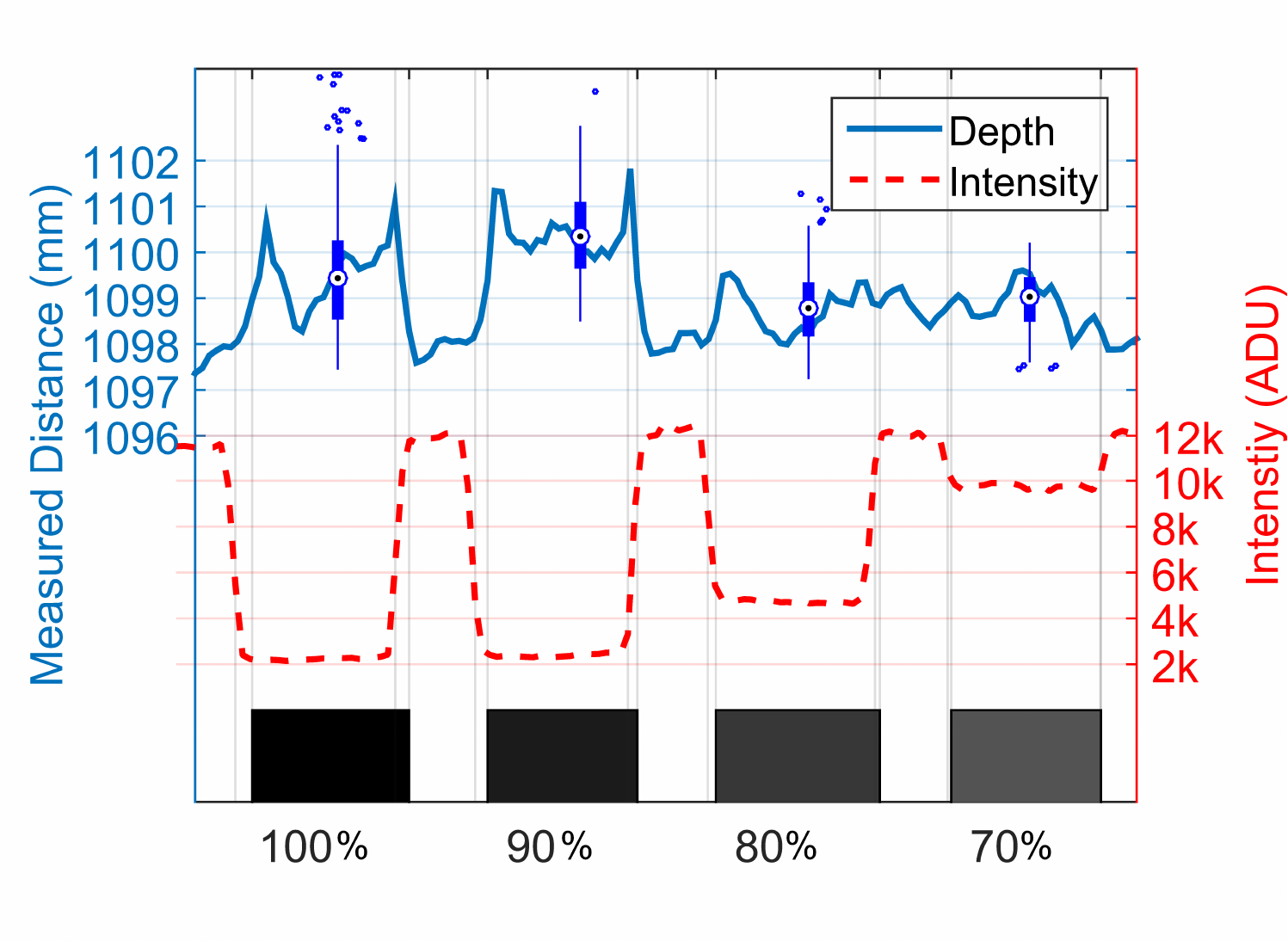}

  \caption{Intensity Related Error for \kintof. Measured depth and
    intensity versus the actual intensity of the checker board 
    at about 1 meter distance. The intensity is given in
    \emph{arbitrary digital units (adu)} as delivered by the \kintof.}
  \label{fig:intensity:results}
\end{figure}

\subsubsection{Intensity (\kintof\ only)}
\label{sec:results:rail.intensity}

Similar to Lindner and Kolb~\cite{lindnerTOF10} we evaluate the planar
checkerboard with varying gray-levels at $1$m distance. For this
scenario we select horizontal pixel lines across the gray-level
rectangles and directly plot the distance values for several distances
to the wall. Fig.~\ref{fig:intensity:results} shows, that the
\kintof\ delivers very stable results and the range error for the
darkest rectangle is max. $3$mm, compared to the white reference
distance. Compared to earlier ToF-camera prototypes, for which range
errors up to $50$mm have been observed~\cite{lindnerTOF10}, this is a
significant improvement of quality.

\subsection{Semitransparent Liquid}
\label{sec:results:milk}

\begin{figure}[t!]
  \centering
  \includegraphics[trim = 12mm 58mm 9mm 70mm,clip,width=.49\textwidth]{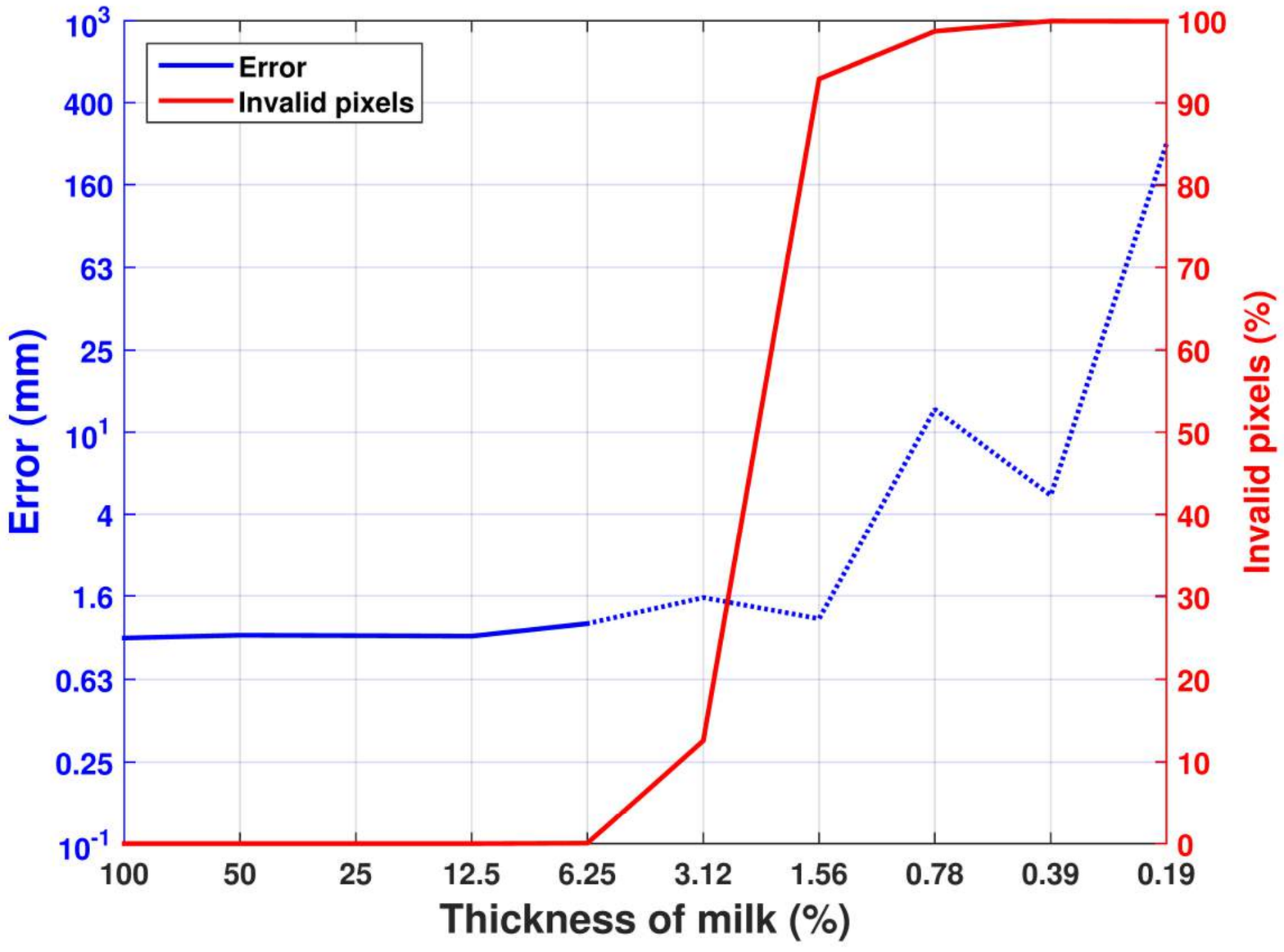}
  \includegraphics[trim = 12mm 58mm 9mm 70mm,clip,width=.49\textwidth]{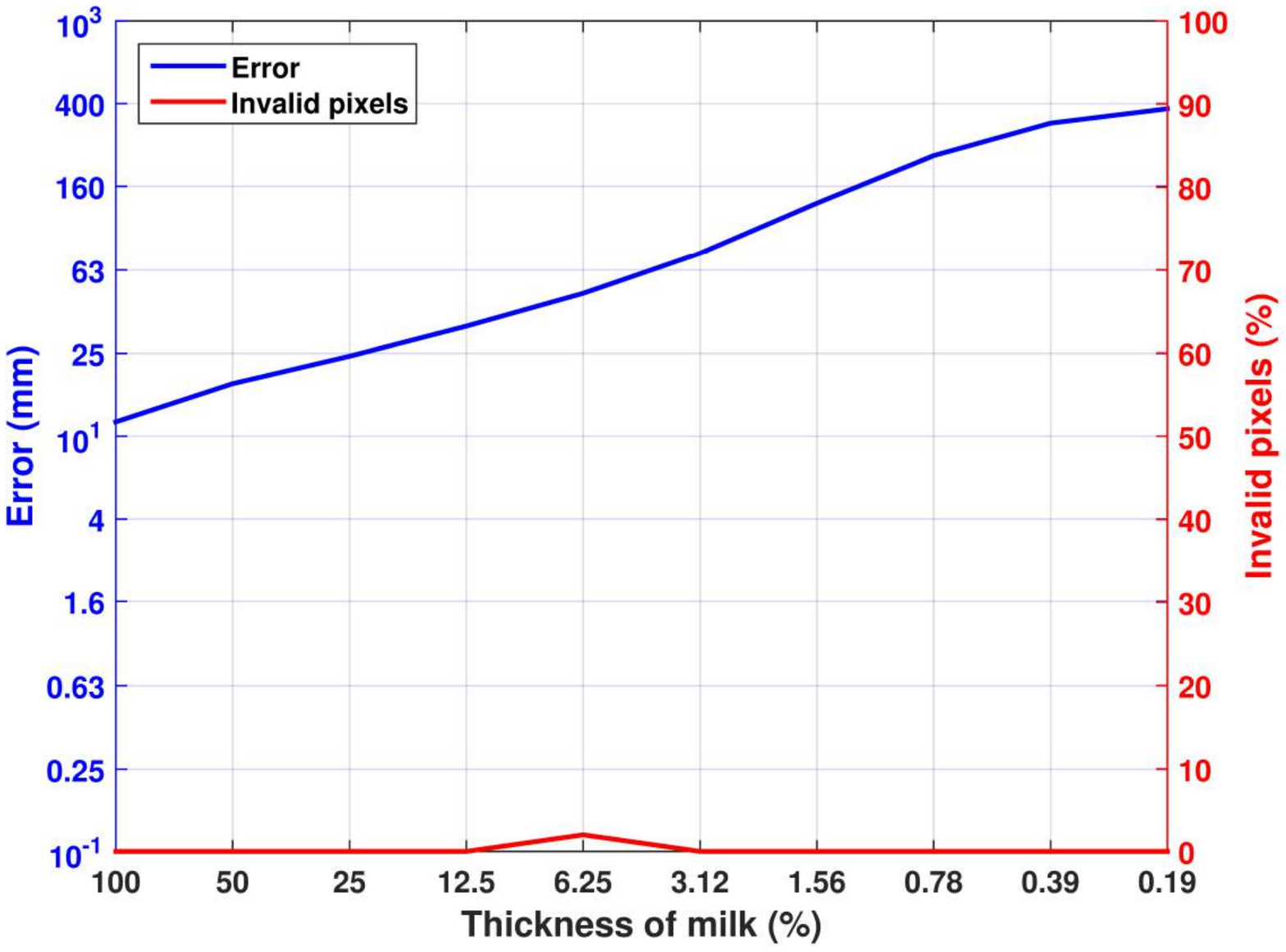}\\
  
  \includegraphics[width=.9\textwidth]{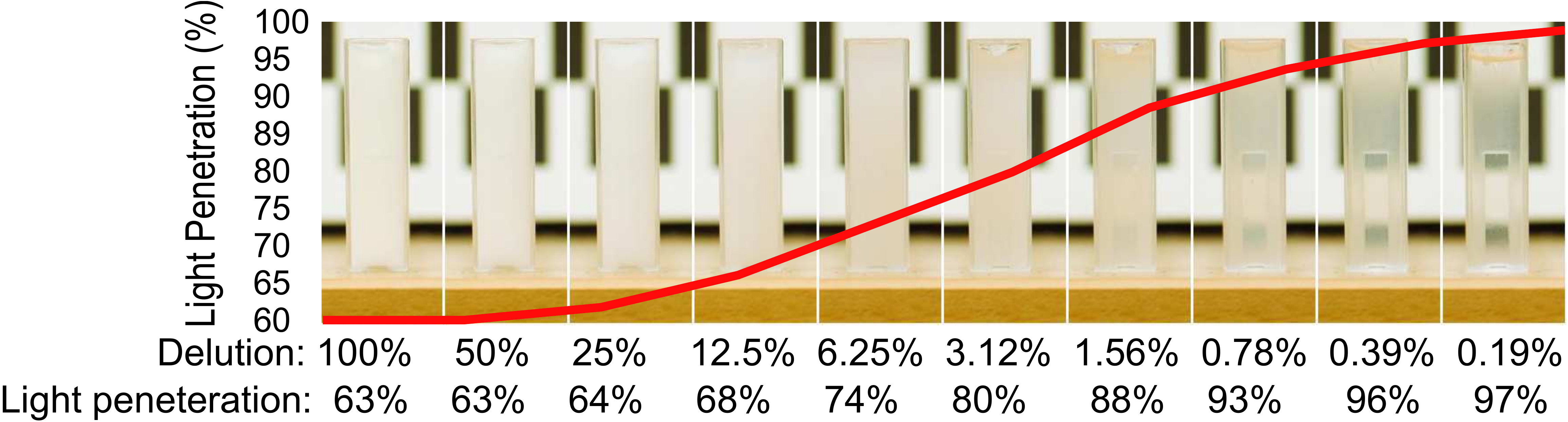}\\

  \caption{Semitransparent Liquid. Depth error and amount invalid
    pixels versus the transparency of the liquid for the
    \kinsl\ (left) and the \kintof\ (right). Samples in standard cuvette are shown in bottom row.}
  \label{fig:Milk:results}
\end{figure}

\paragraph{Goal} This test scenario is designed in order to evaluate
the effects of translucent, i.e. semitransparent and scattering
material on the quality of the acquired object geometry.

\paragraph{Experimental Setup} 
Similar to Hansard\etal\cite{hansard13time} we use a sequence of
semitransparent liquids, i.e.  a plastic cylinder filled with diluted
milk. The cylinder has an inner and outer diameter of $77$ and $79$mm,
respectively. By diluting the milk with the same amount of water in
each step, we get sequence of $10$ objects with an amount of
$2^{-k},\,k=0,\ldots,9$, i.e. $100$\%,\ldots,$0.19$\% milk. We acquire
$200$ frames for each setup. The cylinders are acquired from frontal
view at a distance of $1.2$m. 

Fig.~\ref{fig:Milk:results}, bottom, the visual appearance of the milk
probes is shown. The diluted milk is filled in cuvettes of $1$cm
square cross section and placed in front of a checkerboard in order to
demonstrate the degree transparency in the visual range.

In order to provide a quantitative transparency degree, we measured the light penetration through the cuvettes at $850$nm. The measured intesity through a cuvette filled with water was the reference $$(I_0)$$ and each sample was divided by the reference. The Kodak Wratten $850$nm filter explained at ~\ref{sec:results.ambient} was applied to filter visible light.
$$
Penetration _{@850nm}=\frac{I}{I_0} \times 100,
$$ 

\paragraph{Evaluation and Results}
Same as Hansard\etal\cite{hansard13time} we directly measure the
signed error in depth for a manually segmented region in the range
image with respect to the mean image $d^{\text{mean}}$ as a function
of transparency by comparing against a reference measurement with a
non-transparent cylinder of the same size; see
Eq.~\eqref{eq:SE}. Furthermore, we plot the number of invalid pixels.

As can be seen in Fig.\ref{fig:Milk:results}, top-left,
\kinsl\ performs very well for liquid samples with more than $3.12$\%
milk, with almost no invalid pixels and a signed error in the range of
$[1,1.5]$mm, which is around the thickness of the plastic
cylinder. However, for the samples with concentration of milk below
$3.12$\%, the number of invalid pixels increases dramatically to above
$90$\% and the depth error of the remaining valid pixels is increasing
as well.  For the same experiments the \kintof\ shows a positive
distance error between $12$ and $378$mm, but does not mark any
measurements as invalid, i.e. the number of invalid pixels is
negligible; see Fig.\ref{fig:Milk:results}, top-right. However, for the
samples thinner than $3.1$\%, the number of invalid pixels increases
dramatically to above $90$\% for the \kinsl\, while \kintof\ still
delivers valid pixels with rising error up to $400$mm for $0.2$\%
milk.  In conclusion, \kinsl\ performs good for thicker
semitransparent liquids and indicates failure for the thinner
cases. On the other hand, using the \kintof\ is much harder, as the
device does not indicate the pixel's invalidity even for a large
amount of distance error.

\subsection{Reflective Board}
\label{sec:results:reflect}

\paragraph{Goal} 
This test evaluates the impact of strongly reflecting objects which
potentially result in erroneous depth measurements mainly due to
multi-path effects. Beside the reflectivity as such, the multi-path
effect strongly depends on the orientation of the reflective object
towards other bright objects in the scene and the camera. Therefore,
we are mainly interested in the relation between the angular
orientation and measured depth error.

\begin{figure}[t!]
  \centering
  \includegraphics[trim = 20mm 20mm 20mm 20mm,clip,height=.2\textheight]{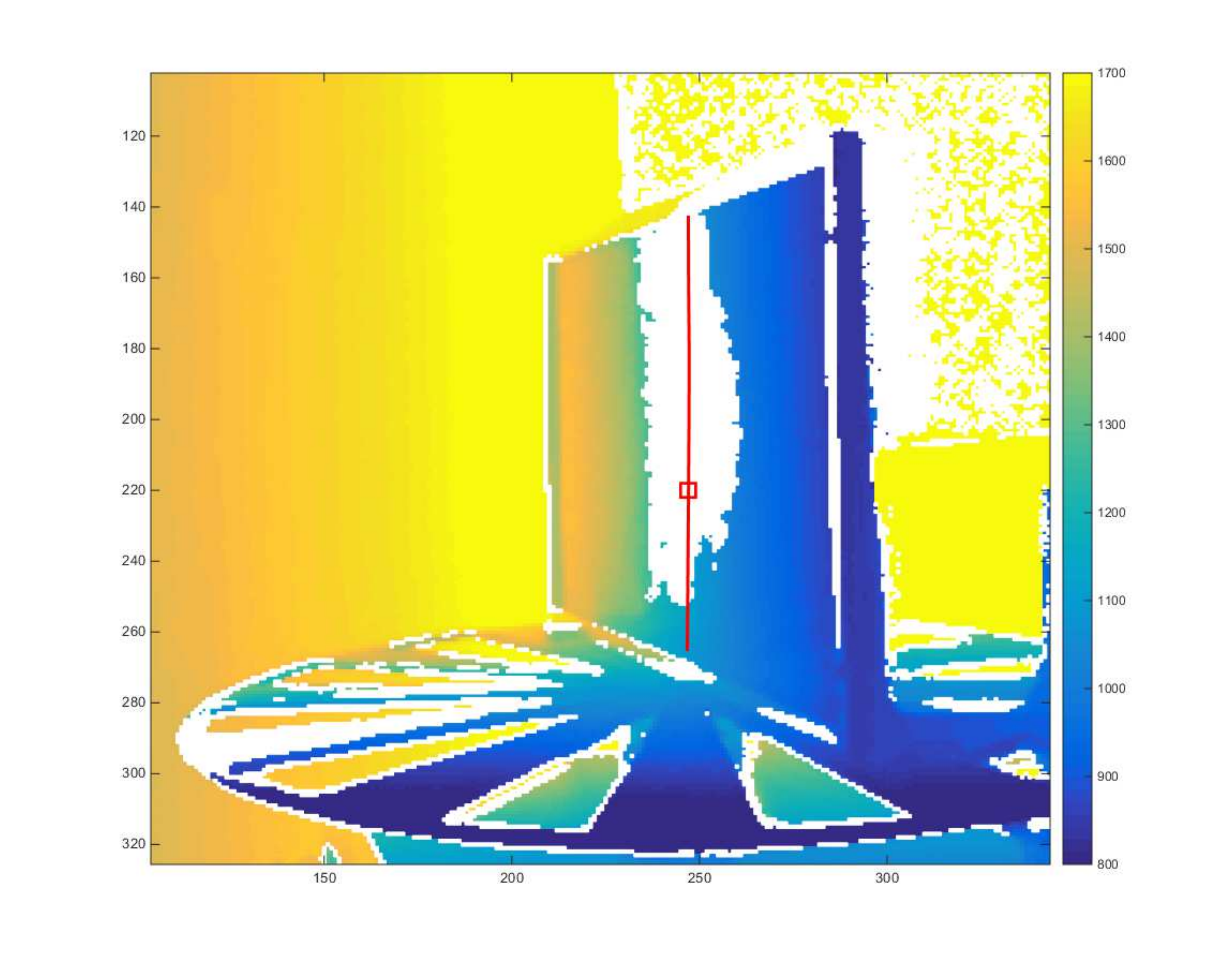} 
  \includegraphics[height=.2\textheight]{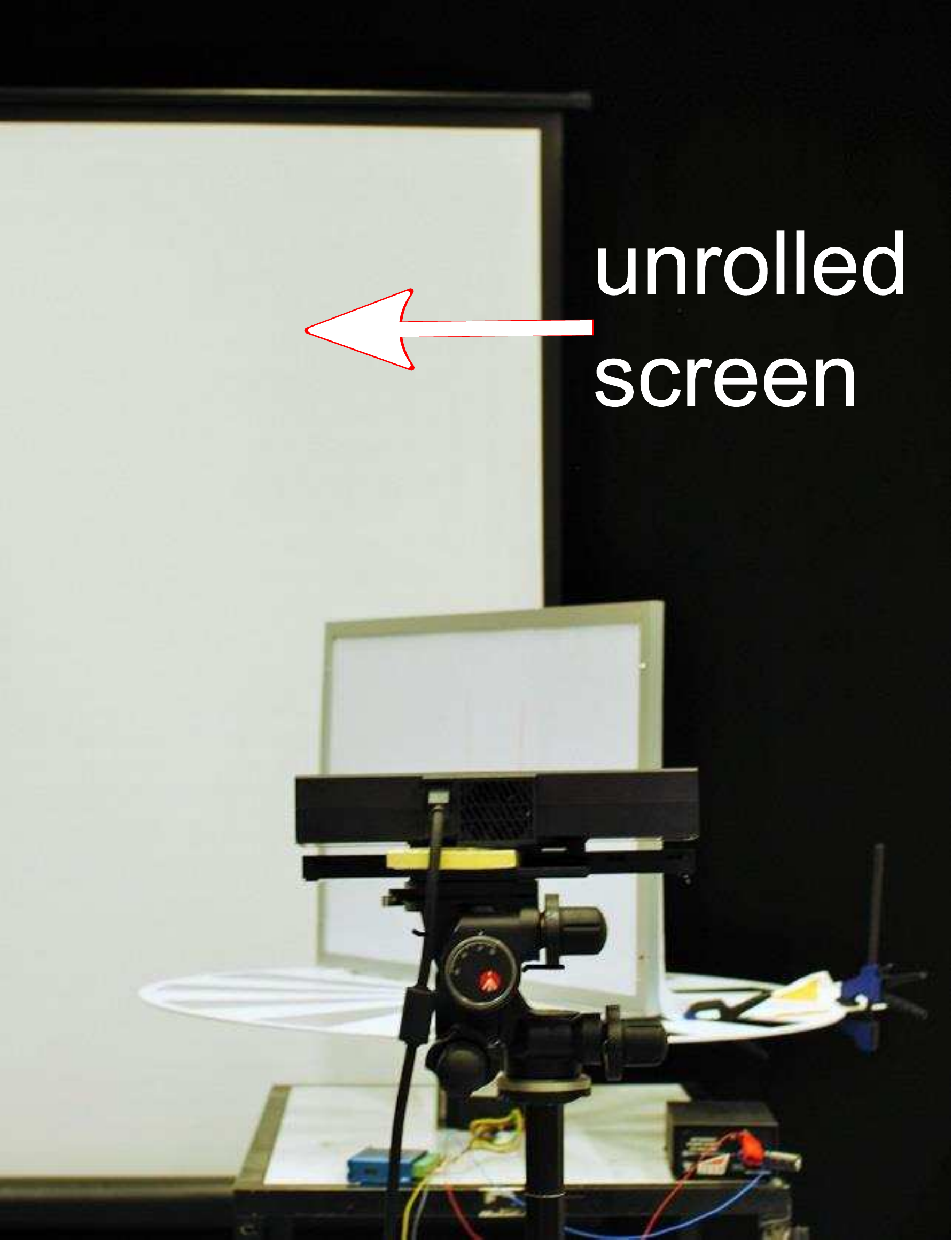}
  \includegraphics[height=.2\textheight]{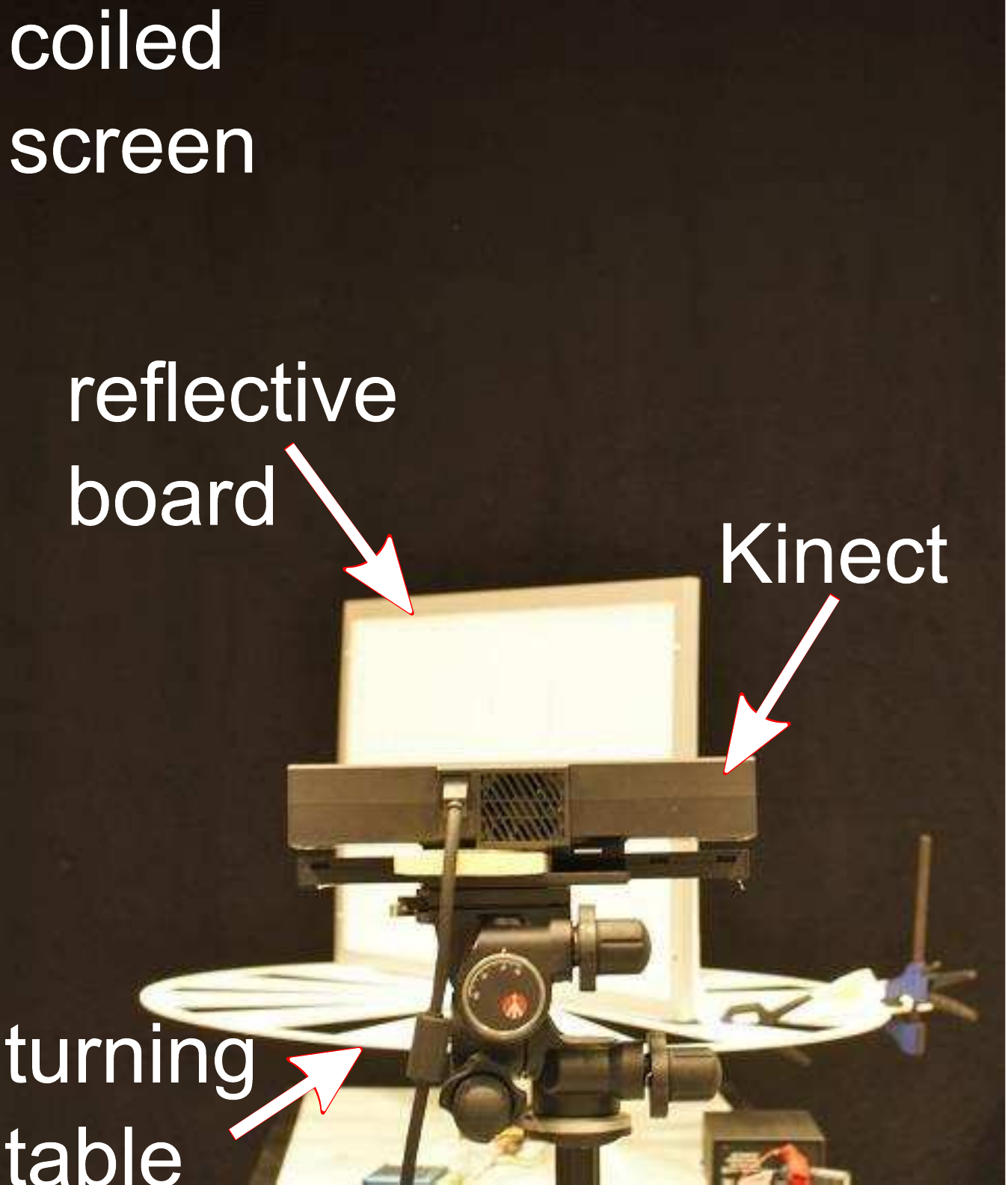}
 
  \caption{Reflective Board. Sample range image with imposed
    pivot-line (left) and a photo of the setup with unrolled screen
    (middle) and coiled curtain (right).}
  \label{fig:multipath:setup}
\end{figure}

\paragraph{Experimental Setup} 
We use a common whiteboard of $60\times40$cm size as reflective object
and place it vertically on a turning table in front of a white
projector screen at a distance of $170$cm from the camera. The
projector screen can be rolled up and behind that there is a
non-reflecting black curtain in order to make a non-multi-path
reference measurements; see Fig.~\ref{fig:multipath:setup}, middle and
right. The rotating vertical board is placed in front of the Kinect
camera so that the board rotates around the pivot line which
intersects the center of the rotating table. The points lying on the
pivot line remain at the same distance to the camera. The rotation
starts from $0^{\circ}$ to $90^{\circ}$ with resolution of
$0^{\circ}15'$. The specific multi-path effect depends on the board
angle. For each step we acquire $20$ frames.

\paragraph{Evaluation and Results} 

For each acquired pair of range images, i.e. for a given fixed angle,
with a coiled and unrolled screen, we select a vertical $4\times100$
pixel region of interest around the rotation pivot on the
whiteboard. For each pixel in the vertical region of interest we assume
a constant distance to the camera and a constant multi-path
situation. Within this region we compute the RMSE with respect to the
reference measurement at $90^\circ$ and the SD of the measurement itself as a
function of the incident angle. Furthermore, we calculate the relative
number of invalid pixels.

\begin{figure}[t!]
  \centering
  \begin{tabular}{cc}
    \includegraphics[trim = 14mm 65mm 4mm
      71mm,clip,width=.49\textwidth]{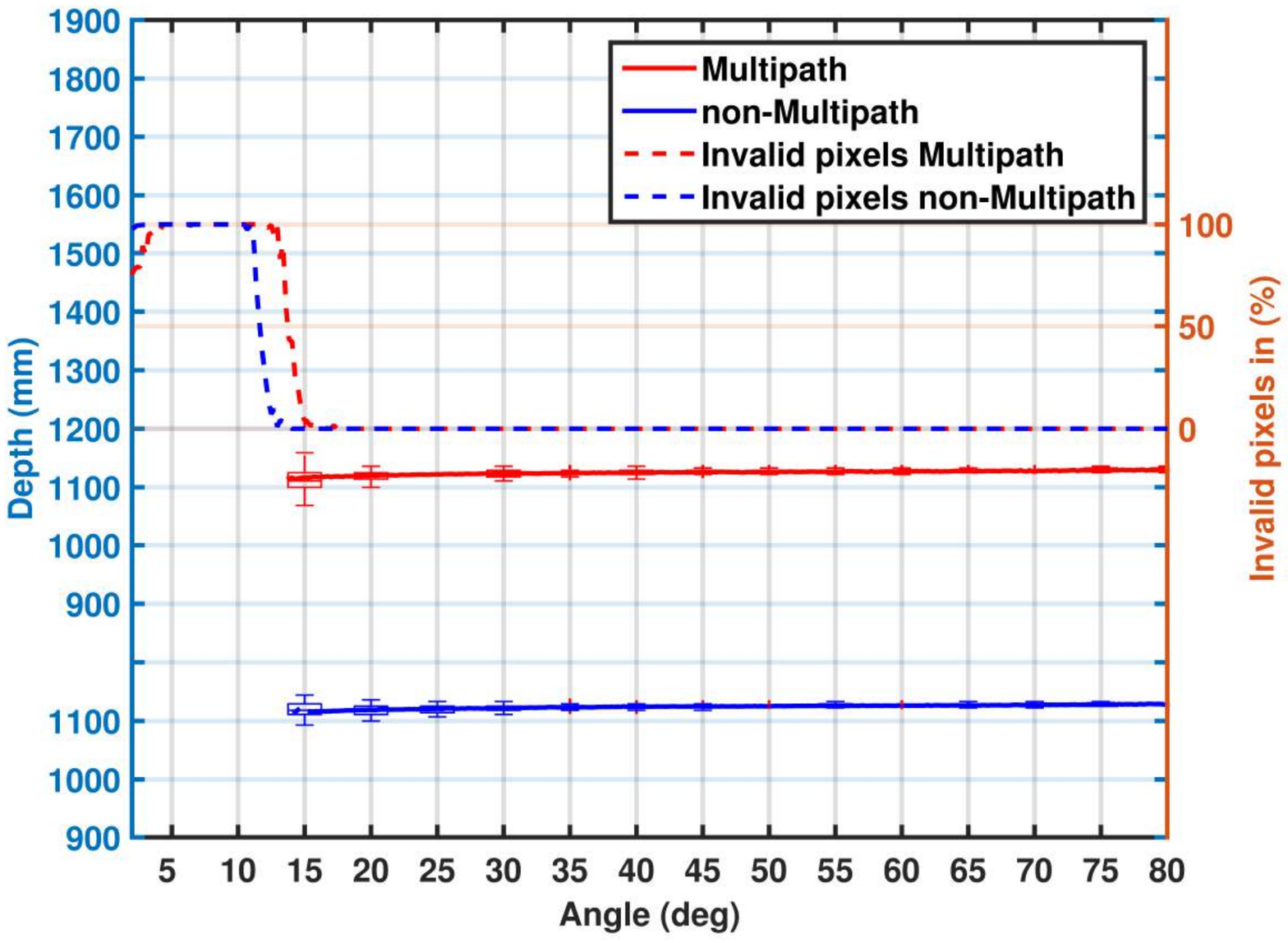} &
    \includegraphics[trim = 14mm 65mm 4mm
      71mm,clip,width=.49\textwidth]{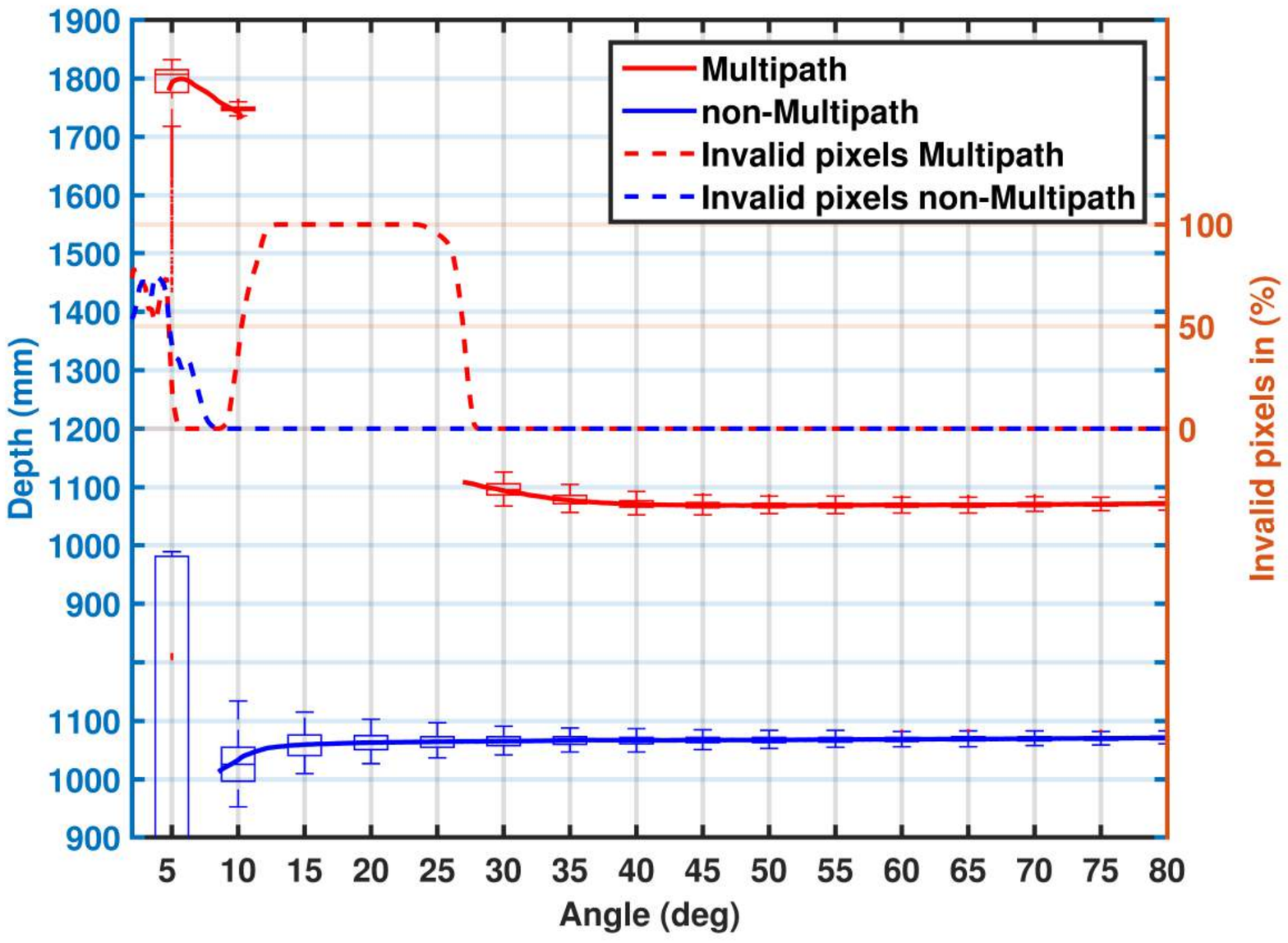} \\
    \small \kinsl & \small \kintof
  \end{tabular}
  \caption{Reflective Board.  The SD and the amount of invalid pixels
    versus the angle of incidence for the non-multi-path and the
    multi-path situation for the \kinsl\ (left) and the
    \kintof\ (right).}
  \label{fig:multipath:results}
\end{figure}

In Fig.~\ref{fig:multipath:results} the RMSE and the SD for all
acquisition angles for the \kinsl\ (left) and the \kintof\ (right) are
plotted. Additionally, we plot the amount of invalid pixel. As
expected, the \kinsl\ has much less problem with this indirect
lighting setup, since the structured light principle does not get
confused by diffuse scattered light. However, the \kinsl\ has also
limitations for low angles and delivers a higher invalid depth for low incident angles. Even though the measurement principle
should not get affected by this. One simple explanation would be, that
too little light is getting reflected to the camera, however, this would
also be true for the reference measurement with coiled
screen.

 For incident angles
below $10^\circ$ up to $100$\% of the pixels are marked invalid. For
angles above $15^\circ$, the \kinsl\ yields nearly no invalid pixels
and the depth error is close to zero. The \kintof, on the other hand,
has a lot more problems with the superposition of the indirect
illumination, i.e. the multi-path situation. Apparently, the
\kintof\ is able to detect some of the corrupted pixel, but at angles
below $10^\circ$ which get affected by a low incident angle, are not
classified as invalid, resulting in extremely range errors up to
$800$mm. For incident angles between $10^\circ$ and $30^\circ$ the
\kintof\ delivers up to $100$\% invalid pixel. Similar as for the
\kinsl, the \kintof\ range values are again more reliable for angles
above $35^\circ$, i.e. no invalid pixels are delivered with a depth
error below $50$mm.

\subsection{Turning Siemens Star}
\label{sec:results:dynamic}

\begin{figure}[t!]
  \centering
  \includegraphics[height=.24\textheight]{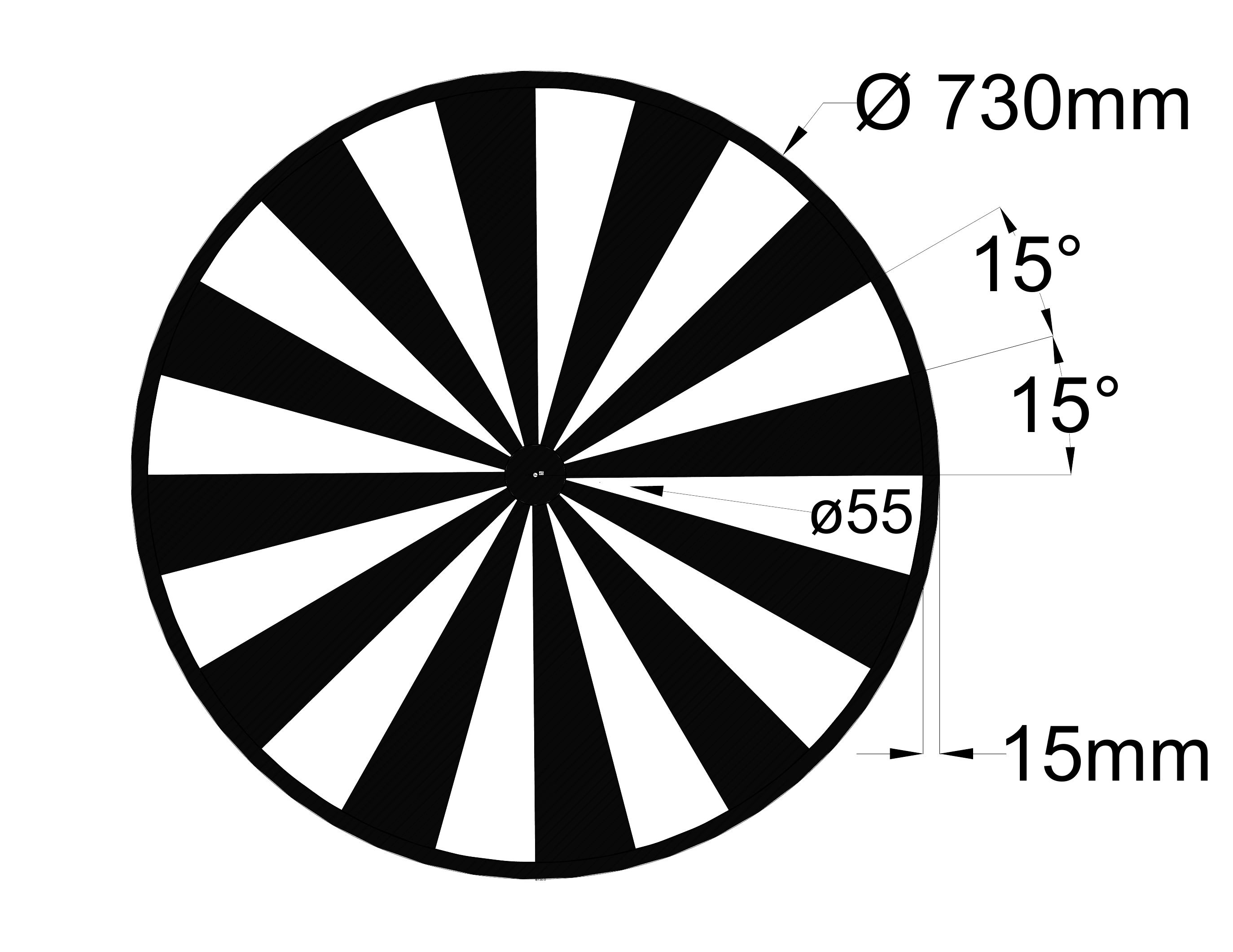}
  \includegraphics[width=.45\textwidth]{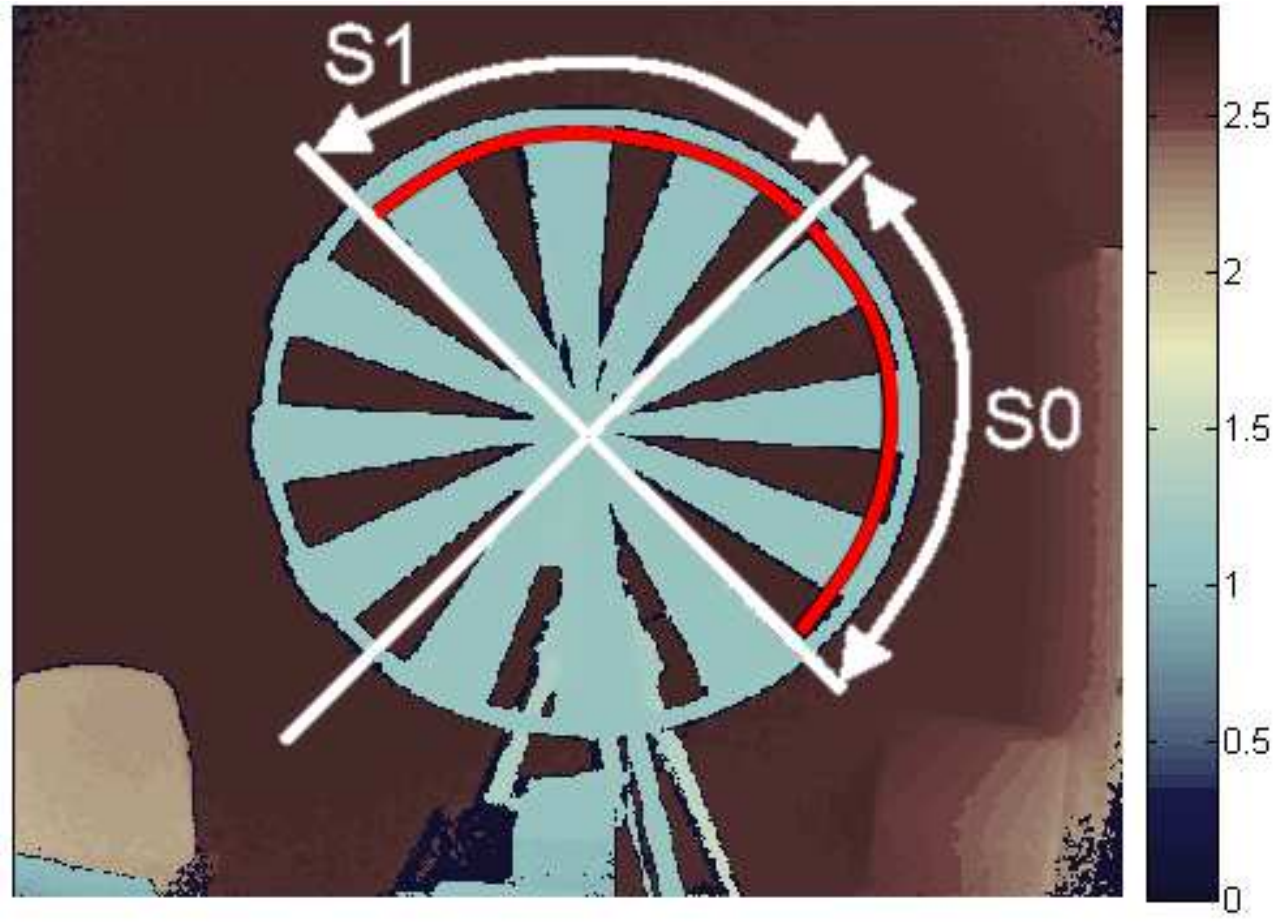}
  \caption{The ``Siemens Star''. Mechanical details (left) and the
    segments as well as the pixels used for the evaluation are marked
    in red (right).}
  \label{fig:MotionArtifacts:segmentation}
\end{figure}

\paragraph{Goal}
The performed test targeted at measuring the amount of flying pixels,
i.e.pixels that cover an inhomogeneous region in terms of depth and
thus do not deliver proper depth values, for static and dynamic
scenes. Both Kinect cameras mark unreliable pixels as ``invalid'',
which also applies for the flying pixels.

\paragraph{Experimental setup}
Similar to Lottner\etal\cite{lottner11scanning} we manufactured a 3D
Siemens star. However, we mount it to a stepping motor that actuates
the star in a controlled fashion in front of a planar background wall
at $1.8$m distance. The geometrical dimensions of the Siemens star are shown in
Fig.~\ref{fig:MotionArtifacts:segmentation}, left.  We apply different
angular velocity while capturing range data with any of the Kinect
cameras. As the both Kinects have different intrinsic parameters and
different range image resolution, there are different options for the
geometric setup of the measurement. We opted for a setup where each
sensor has the same ``pixel coverage'' on the star, thus the cameras
have different distances to the star during the acquisition, but as
both cameras have approximately the same temporal resolution,
i.e. frame rate, pixel coverage for the lateral motion is comparable
for both Kinect.

We acquire range data for the static and the dynamic wheel. We have
chosen nine velocity steps between $0,11,22,\ldots,100$RPM. In our
setup $100$RPM relates to $35$ pixel swept in the most outer circle
(red arc in Fig~\ref{fig:MotionArtifacts:segmentation}, right) by the
wheel within one range image, i.e. in $1/30$s.

\begin{figure}[t!]
  \centering
  \begin{tabular}{cc}
    \includegraphics[trim = 28mm 65mm 9mm 71mm ,width=.5\textwidth]{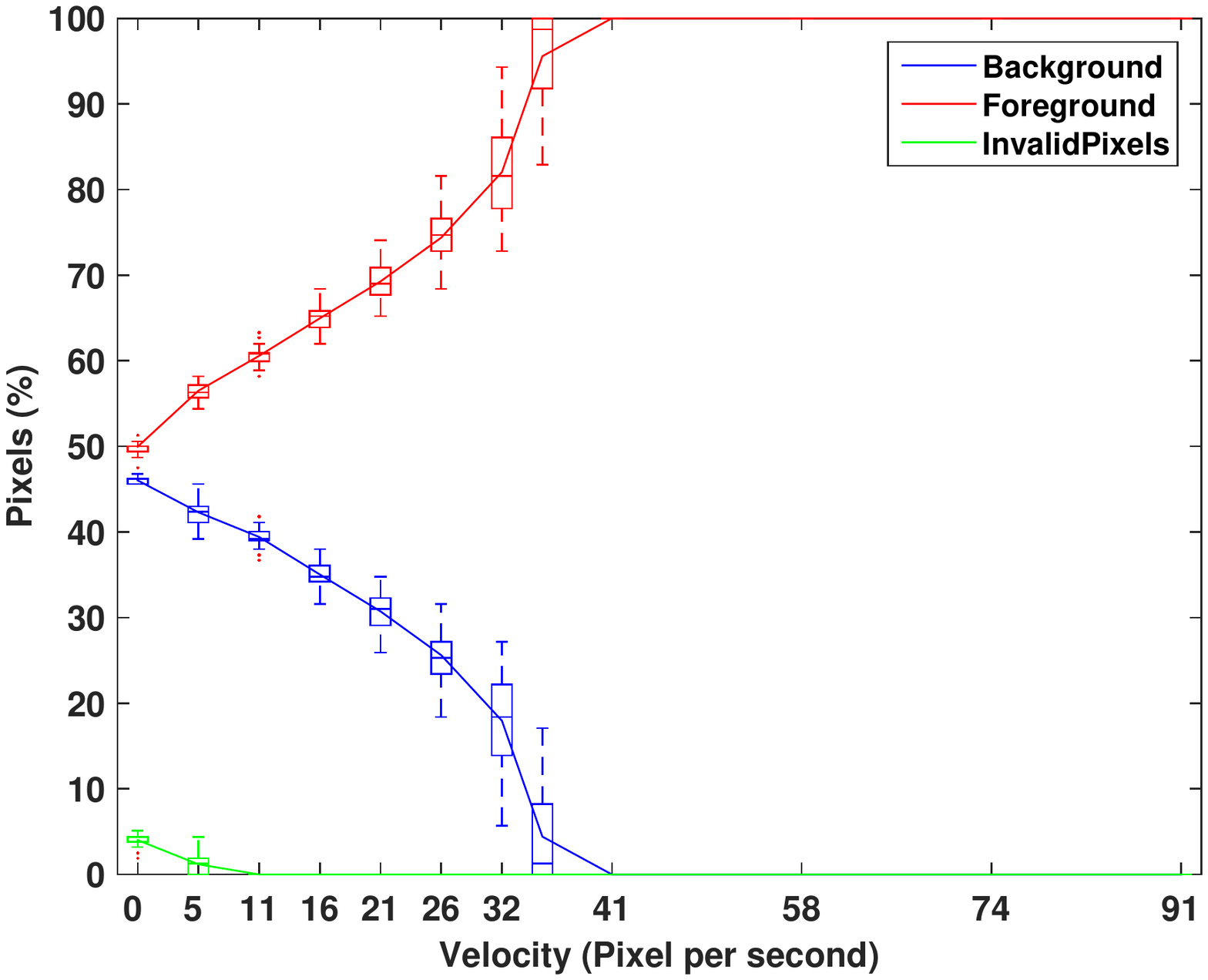} &
    \includegraphics[trim = 28mm 65mm 9mm 71mm,width=.5\textwidth]{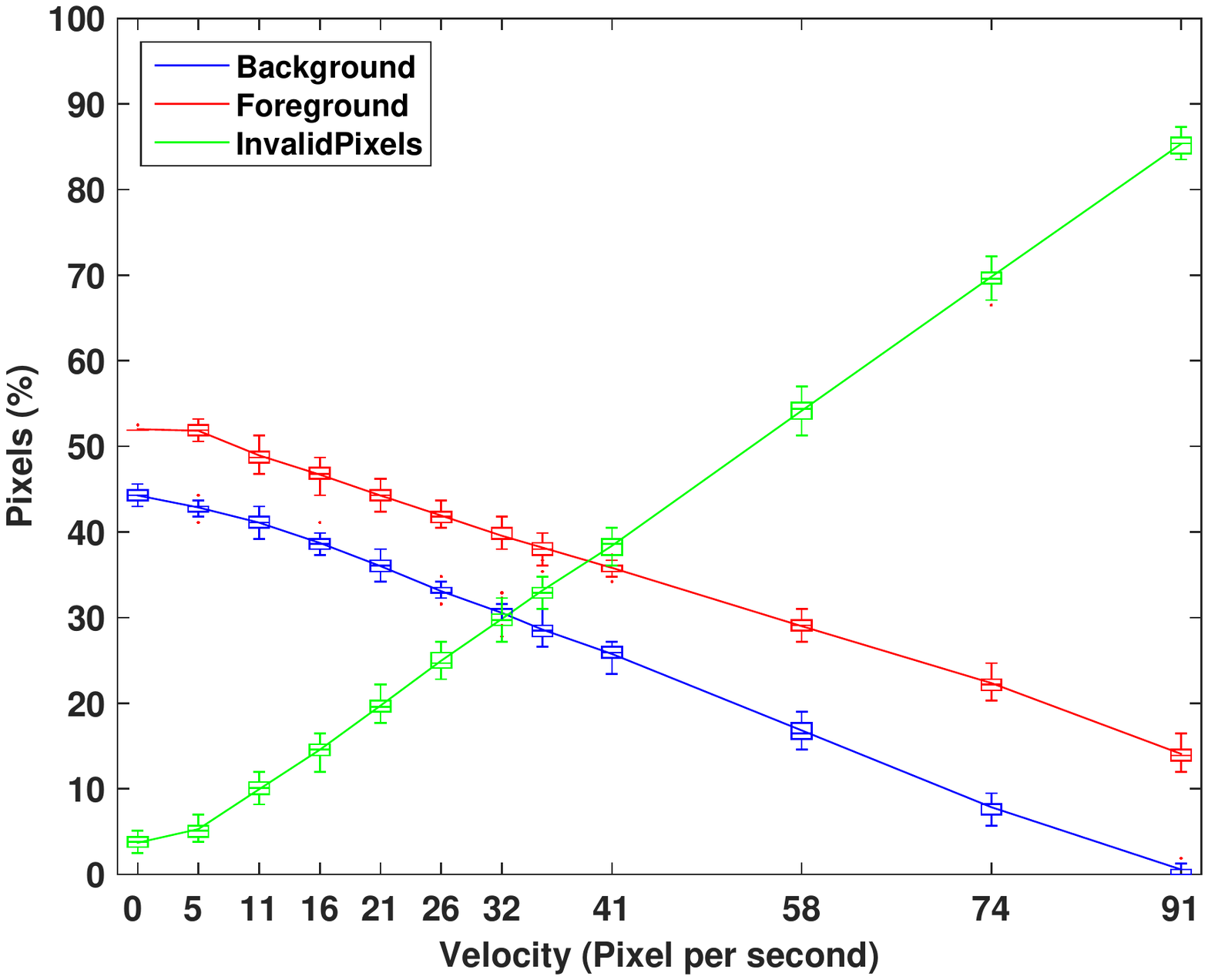}\\
    
    \includegraphics[trim = 28mm 65mm 9mm 71mm,width=.5\textwidth]{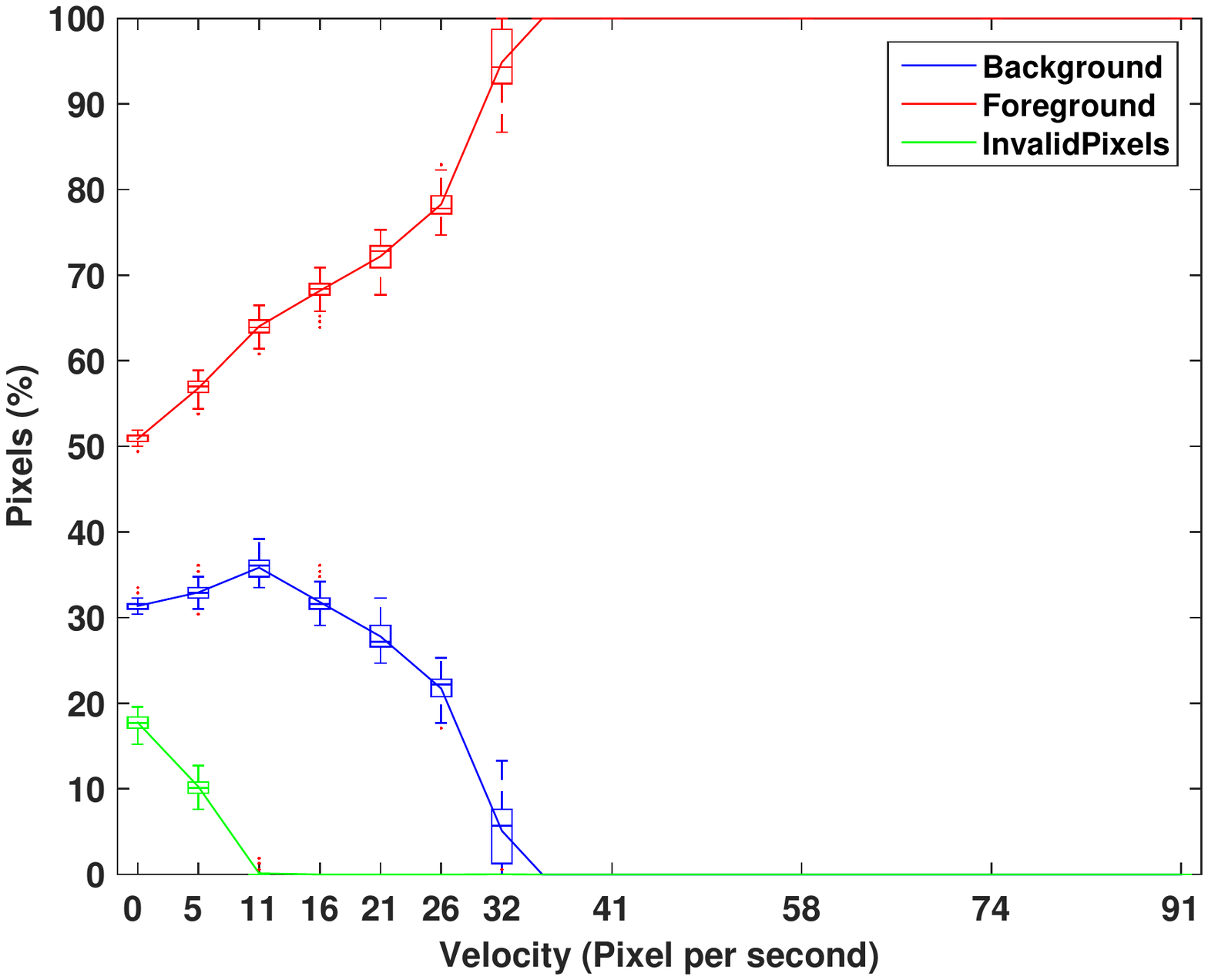} &
    \includegraphics[trim = 28mm 65mm 9mm 71mm, width=.5\textwidth]{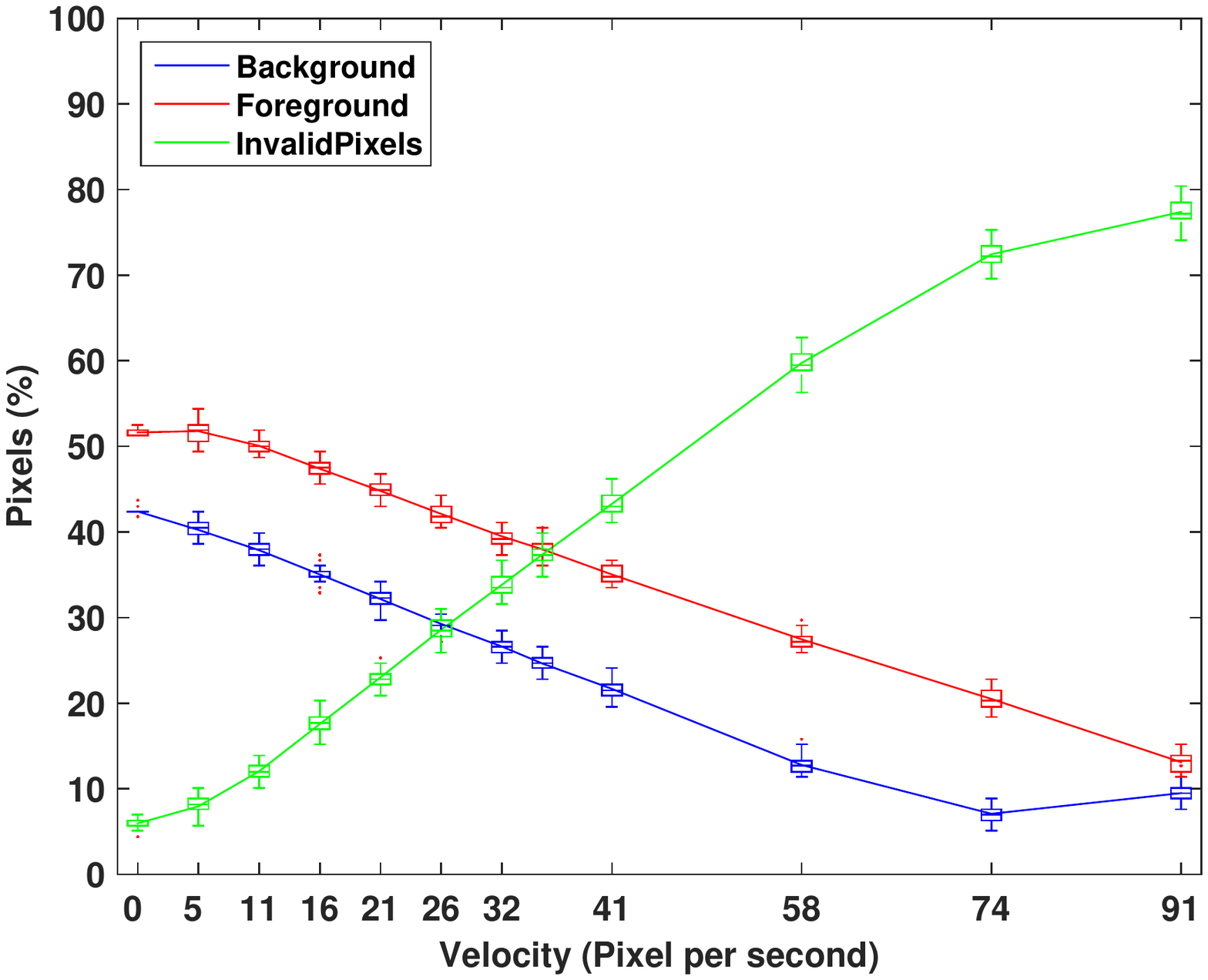}\\
    
  \end{tabular}
  \caption{Analysis for the Turning Siemens Star. The minimal, mean
    and maximal relative numbers of foreground, background and invalid
    pixels plotted over the angular velocity for the \kinsl\ (left) and the
    \kintof\ (right) for the segments ``S0'' (top) and ``S1'' (bottom).}
  \label{fig:MotionArtifacts:results}
\end{figure}

\paragraph{Evaluation and Results}

In the evaluation we account for the fact, that the Kinect's
illumination units are mounted horizontally for both cameras, leading
to different shadowing effects at vertical and horizontal edges. Thus
we expected varying results between regions with predominantly
vertical and horizontal edges and performed the analysis separately
for the two wheel quarters, one at the right (``S0'') and one at the
top (``S1''). For the evaluation we use pixels at a circular arc at
the outer part of the wheel illustrated by the red arc; see in
Fig~\ref{fig:MotionArtifacts:segmentation}, right.

In an ideal case, along the arcs there should be 50\% foreground
and 50\% background pixels. Therefore we simply calculate the minimal,
mean and maximal relative numbers of foreground, background and
invalid pixels for the different speed values.

Fig.~\ref{fig:MotionArtifacts:results} shows the results of the
foreground-background analysis for both cameras and both segments. One
first insight is, that the classification results are very stable,
namely the \kintof\ shows very little variation in its results.
Comparing the classification results for the static scene, i.e. the
flying pixels, the foreground classification is nearly perfect,
i.e. $52.5$\% for the \kinsl\ and $53.2$\% for the \kintof\ for
segment ``S0''. The amount of invalid pixel for this segment is
$7.0$\% for the \kinsl\ and $10.9$\% for the \kintof.

For increasing speed, it is apparent that the \kinsl\ delivers less
invalid pixels and more (false) foreground pixels , resulting in $100$\%
foreground pixels for $100$RPM. The behavior of the \kintof\ is much
more reliable. As one would expect, the number of invalid pixels
increases for higher speed and the number of foreground and background
pixel decreases in a comparable way. However, there are always more
foreground than background pixel. This effect can be explained by the
shadowing which only applies to the background, i.e. the holes in the
Siemens star.


As expected, there are differences between the two arc segments for
both Kinect. In general, the results for the top segment ``S1'' are
worse for both devices, as shadowing effects are stronger for vertical
edges. For the \kinsl\ mainly the number of invalid pixels is higher
for lower speed, which is counter-intuitive. For the \kintof\ the
differences between the two segments are less prominent.

\begin{figure}[t!]
  \centering
  \includegraphics[trim = 23mm 78mm 32mm 96mm,clip,width=.49\textwidth]{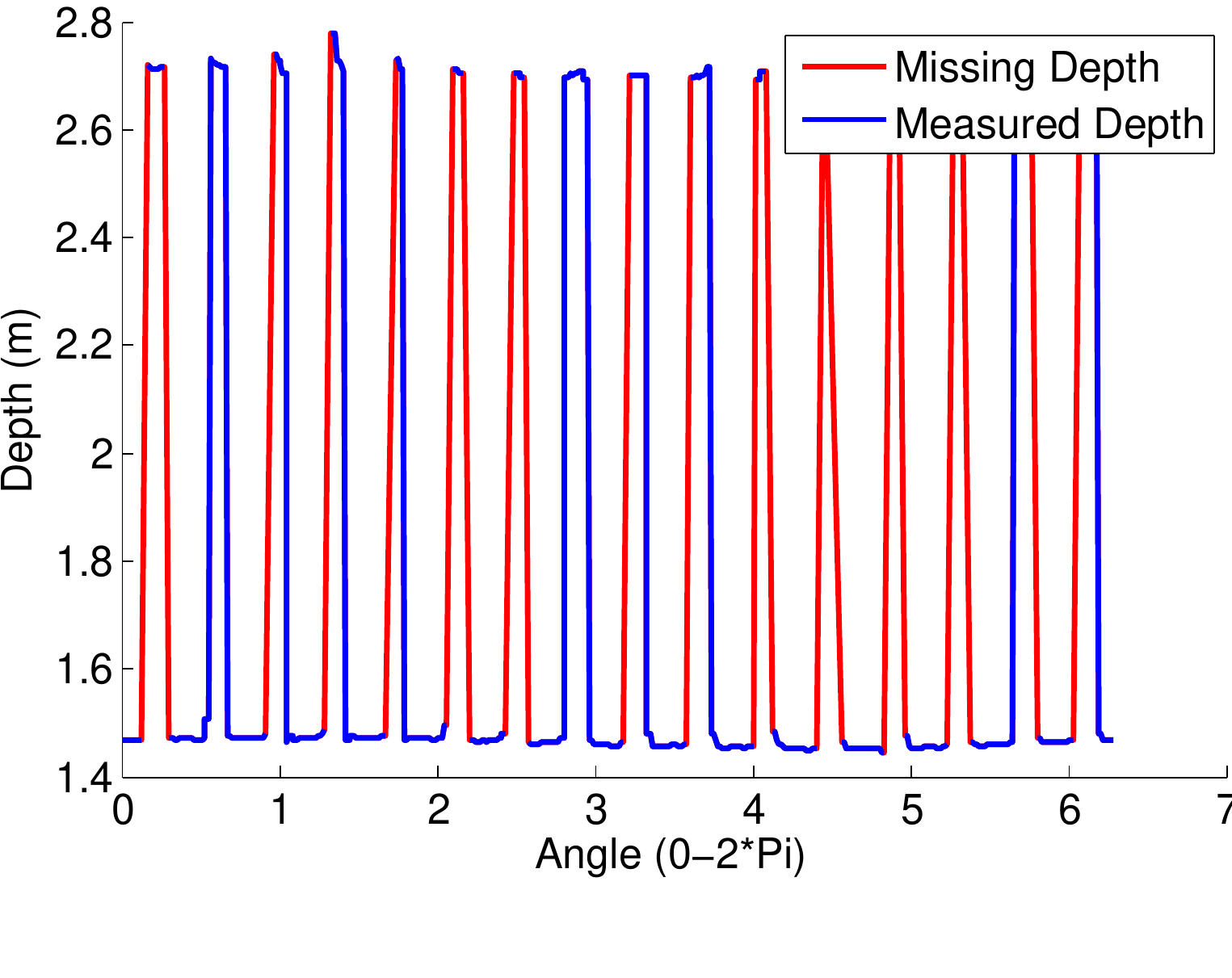}
  \includegraphics[trim = 23mm 78mm 32mm 96mm,clip,width=.49\textwidth]{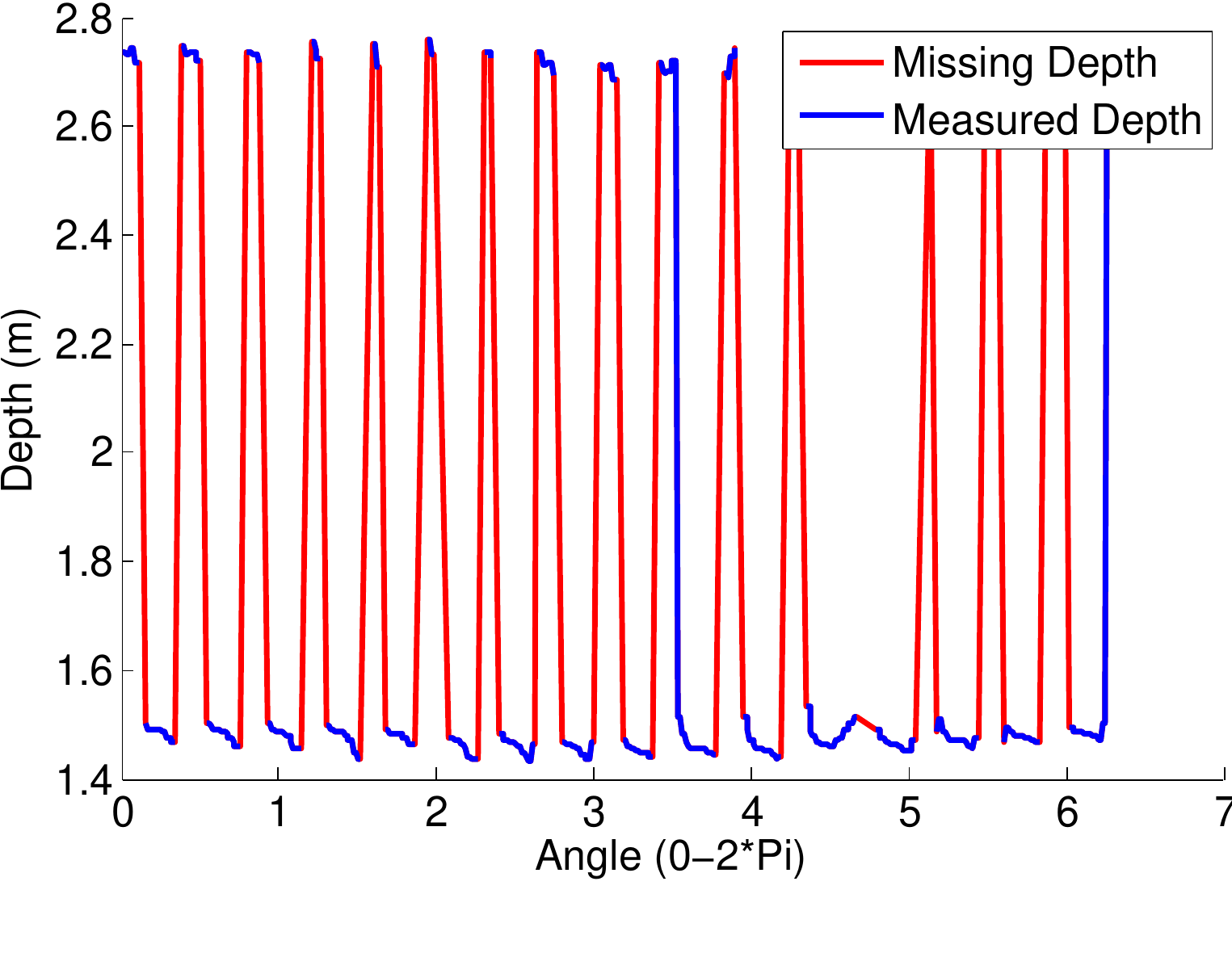}
  \caption{Range Profile for the Turning Siemens Star for the angular
    speed of $0$RPM (left) and $60$RPM (right) measured by \kintof. Red indicates invalid
    pixels. Note, that the range between $4.3-5$~radiant is the lower
    part of the wheel where the tripod distorts the background
    region.}
  \label{fig:MotionArtifacts:profile}
\end{figure}

Beside the classification for the pixels along the arc,
Fig.~\ref{fig:MotionArtifacts:profile} shows the range profile for
$0$RPM and $60$RPM for the full outer pixel circle. This profile
plot shows an additional range distortion effect at the edges of the
foreground parts. Here, additional ``overshooting'' effects occur,
which are due to motion artifacts apparent to ToF cameras; see
Sec.~\ref{sec:principle.err}.

\section{Conclusion}
\label{sec:conclusion}

This paper presents an in-depth comparison between the two versions of
the Kinect range sensor, i.e. the \kinsl, which is based on the
Structured Light principle, and the new Time-of-Flight variant
\kintof. We present a framework for evaluating Structured Light and
Time-of-Flight cameras, such as the two Kinect variants, for which
we give detailed insight here. Our evaluation framework consists of
seven experimental setups that cover the full range of known artifacts
for these kinds of range cameras.

\begin{table}
  \begin{tabular}{|>{\small}p{.5\textwidth}|>{\small}p{.5\textwidth}|} \hline
    \hfill\bfseries\kinsl\hfill~ & 
    \hfill\bfseries\kintof\hfill~\\ \hline\hline

    \multicolumn{2}{|c|}{Ambient Background Light
      (Sec.~\ref{sec:results.ambient})} \\ \hline 
    Below $1\mu$W: IP=0\%, SD$\textless 6$mm\newline
    Above $1\mu$mW: IP=$ 100$\%
    & 
    Below $1\mu$mW: IP=0\%, SD$\textless 11$mm\newline
    Below $10\mu$mW: IP=0\%, SD$\textless 30$mm\newline
    At $20\mu$mW: IP=0\%, SD$\textless 42$mm
    \\ \hline 

    \multicolumn{2}{|c|}{Multi-Device Interference
      (Sec.~\ref{sec:results:interference})} \\ \hline 
    IP:  $\textless 16.3$\% (evenly distributed)\newline
    Frame $\textless 400$: 
    RMSE w/o interf. $\textless 6.7$mm,    
    RMSE w interf. $\textless 7.7$mm \newline
    Rest of frames:
    RMSE w/o interf.$\textless 5.8$mm,    
    RMSE w interf.$\textless 9.4$mm
    & 
    IP:  $\textless 22.7$\% (repetitively blocked) \newline
    RMSE w/o interf. $\textless 4.6$mm \newline
    RMSE w interf. $\textless 19.3$mm,
    
    \\ \hline
    \multicolumn{2}{|c|}{Temperature Drift
      (Sec.~\ref{sec:results:temperature})} \\ \hline
    Before 10':  RMSE $\in[4;4.6]$mm,
    SDA $\in[0.6;1.8]$mm \newline
    After 10':
    RMSE rising from $4.0$ to $7.1$mm,
    SDA $\in[0.6;1.0]$mm
    & 
    RMSE $\in[4.6;5.3]$mm \&  
    SDA $\in[1.3;1.5]$mm\\ \hline

    \multicolumn{2}{|c|}{Linearity Error
      (Sec.~\ref{sec:results:rail.linearity}); Pt\#1
          (center)\ldots Pt\#4 (corner); \kinsl\ below $3$m}\\ \hline
    SE1$\in[-34;-1.5]$,
    SE2$\in[-6.5;62]$mm, 
    SE3$\in[-4 ;129]$mm, 
    SE4$\in[2.5;76]$mm \newline
    SD1-3$\in[0.4;14]$mm, SD4$\in[0.4;28]$mm 
    &    
    SE1$\in[-8;29]$mm 
    SE2$\in[-8;17]$mm,
    SE3$\in[-8;37]$mm,
    SE4$\in[-69;62]$mm\newline    
    SD1-3$\in[0.8;6.8]$mm,
    SD4$\in[1.8;90]$mm
    \\ \hline

    \multicolumn{2}{|c|}{Systematic Error: Planarity
      (Sec.~\ref{sec:results:rail.planarity})} \\ \hline
    SD($\le 1.5$m)$\in[1.2;4.8]$mm, \newline
    SD($\le 2.5$m)$\in[2.7;16.6]$mm,\newline
    SD($\le 3.5$m)$\in[7.5;30.9]$mm  &
    SD$\textless 1.65$mm
    \\ \hline

    \multicolumn{2}{|c|}{Intensity Related Error
      (Sec.~\ref{sec:results:rail.intensity})} \\ \hline 
    --not applicable--& 
    distance error $\textless 3$mm@~1m distance
    \\ \hline

    \multicolumn{2}{|c|}{Semitransparent Media \& Scattering
      (Sec.~\ref{sec:results:milk})} \\ \hline
    Light Penetration $\ge80$\%: \\
    SE$\in[1;1.5]$mm,
    IP$\textless 5$\%
    & 
    SE$\in[17.89;378.1]$mm,
    IP$\textless 2$\%
    \\ \hline

  \end{tabular}
\end{table}
\begin{table}
  \begin{tabular}{|>{\small}p{.5\textwidth}|>{\small}p{.5\textwidth}|} \hline
    \hfill\bfseries\kinsl\hfill~ & 
    \hfill\bfseries\kintof\hfill~\\ \hline\hline

    \multicolumn{2}{|c|}{Multipath Effect
      (Sec.~\ref{sec:results:reflect})} \\ \hline
    Incid. angle $\textless 13^\circ$: 
    IP$\textgreater 90$\%,
    \newline
    Incid. angle $\textgreater 20^\circ$: 
    IP$\textless 1$\%,  
    Err. $\textless 5$mm
    &
    Incid. angle $\textless 10^\circ$:
    IP$\textless 70$\%, 
    Err.$\textgreater 600$mm\newline
    Incid. angle $\in[10^\circ,30^\circ]$:
    IP$\approx 100$\%\newline
    Incid. angle $\textgreater30^\circ$:
    IP$\textless 2$\%, 
    Err. $\textless 4$mm
    \\ \hline

    \multicolumn{2}{|c|}{Depth Inhomogeneity and Dynamic Scenery
      (Sec.~\ref{sec:results:dynamic})} \\ \hline
    Seg. ``S0'':
    FG from $52.5$ to $99.48$\%, SD$\textless 5.9$mm,
    BG from $43.0$ to $0.5$\%, SD$\textless 6.0$mm,
    IP from $7$ to $0$\%, SD$\textless 0.9$mm \newline
    Seg. ``S1'':
    FG from $52.33$ to $100$\%, SD$\textless 5.0$mm,
    BG from $19.4$ to $0$\%, SD$\textless 3.5$mm,
    IP from $28.1$ to $0$\%, SD$\textless 3.1$mm
    &
    Seg. ``S0'':
    FG from $53.2$ to $44.8$\%, SD$\textless 1.3$mm,
    BG from $35.9$ to $19.5$\%, SD$\textless 1.5$mm,
    IP from $10.9$ to $35.6$\%, SD$\textless 2.1$mm\newline
    Seg. ``S1'':
    FG from $57$ to $41.11$\%, SD$\textless 1.4$mm,
    BG from $36$ to $18.3$\%, SD$\textless 1.6$mm,
    IP from $6.8$ to $40.4$\%, SD$\textless 1.5$mm
    \\ \hline
  \end{tabular}
  \caption{Summarizing the major Kinect characteristics. IP=''invalid
    pixel'', SD=''standard deviation'', RMSE=''root mean square
    error'', SDA=''standard deviation average'', SE=''signed error'',
    BG=''background'', FG=''foreground''.}
  \label{tab:conclusion}
\end{table}

\begin{table}
  \centering
  \includegraphics[trim = 15mm 195mm 15mm 0mm,clip,width=1\textwidth]{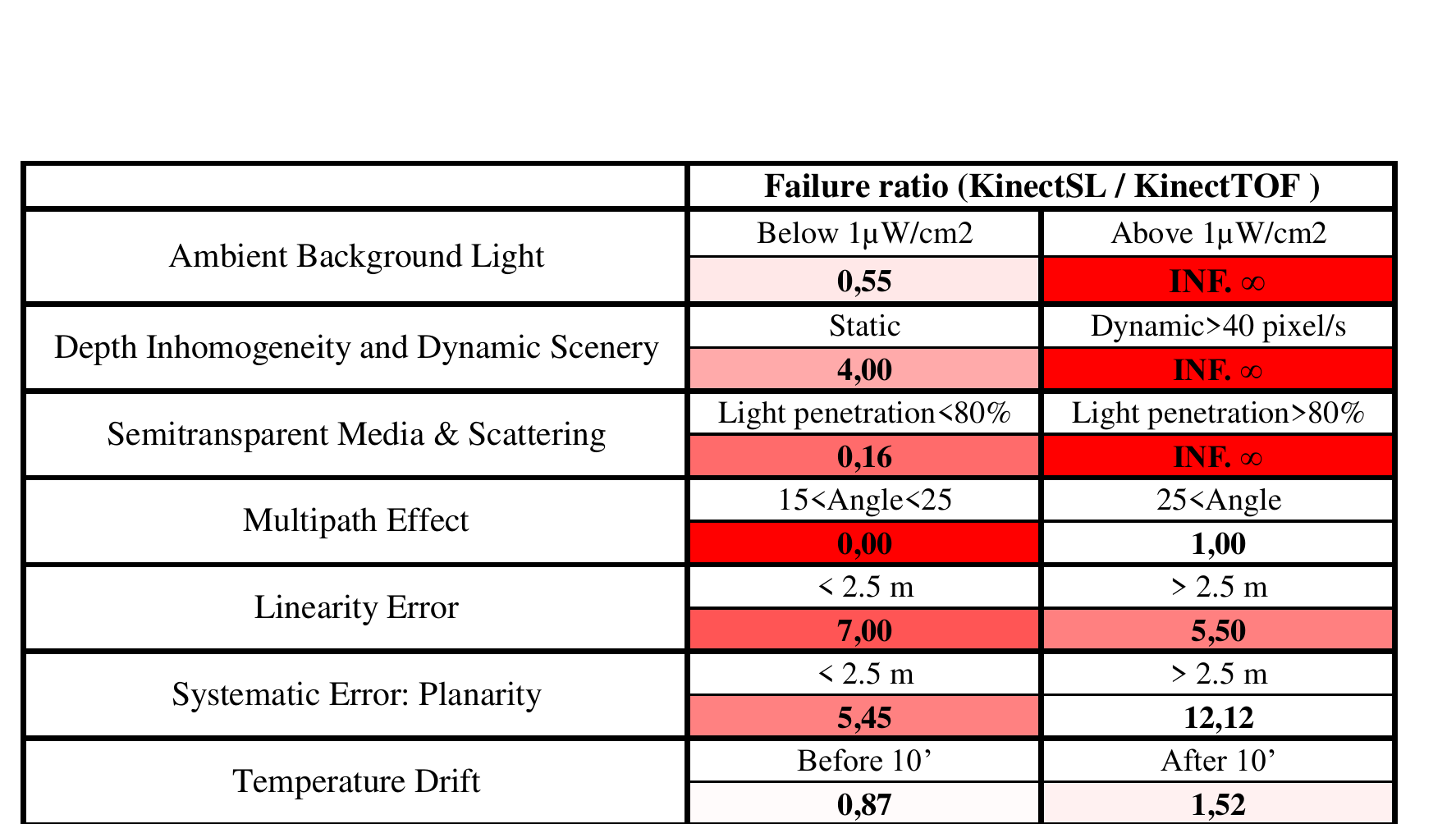} 
  \caption{Device failure ratios for two application modes for the major error sources discussed in this paper. A ratio of 1 indicates equal behaviour of both devices, values close to 0 and infinty indicate high relatively failure for \kinsl\ and \kintof\, respectively. The color intensity indicates deviations from 1, i.e. cases where both devices behave differently.}
  \label{tab:conclusion2}
\end{table}
\subsection*{Device Selection Hints}
Since device selection is highly application dependent, Tab.~\ref{tab:conclusion2} sets up some rules of thumb to help users to make a more profound decision on which device to select depending on their application circumstances. The table compares device performance in different conditions with respect to the main error sources discussed above. For each source, we define two modes of operation, i.e. two ranges of application parameters. For each mode, we state a failure weight which is the performance ratio for both Kinect cameras. The individual failure values are deduced by combining error values and the number of valid pixels for each mode. Based on the individual failure values we compute the ratio between the \kinsl and the \kintof failure, whereby a ratio close to 1 means that both devices perform quite similarly, whereas values close to 0 or close to infinity indicate, that \kinsl and \kintof have relatively high failure rates, respectively. For a specific application scenario, the user selects relevant error sources and by multiplying the failure ratios, the overall failure ratio is computed. If this final ratio is smaller than 1, \kinsl would be the best choice, otherwise \kintof is preferable. Of course, this is only a very coarse but quick guideline resulting in a first suggestion. The user should in any case have a further look at the details for the error sources that are most relevant to the specific application.
 
Note, that we dropped the error sources ``Intensity Related Error'' and ``Multi-Device Interference" from Tab.~\ref{tab:conclusion}, because the ``Intensity Related Error" applies only to \kintof and has, compared to prior ToF devices only very little impact. Furthermore, if ``Multi-Device Interference" is essential to the application, further actions need to be applied, such as using ``Shake`n'Sense" in case of \kinsl \cite{butlerSNS12} or different modulation frequencies in case of the  \kintof.

Example 1:
User A requires a depth sensing device for indoor scene reconstruction where the scene has static semi reflective surfaces at high angles:

  \text{Failure ratio} = $0.55\times 4\times 0.16\times 0\times 7 \times 5.45 \times 0.87$ = $0$ 

Therefore \kintof would absolutely fail in this application.

Example 2:
User B requires face gesture recognition at 1.2 meter distance in indoor office conditions:
 
    \text{Failure ratio} = $0.55\times 4\times 0.16\times 1\times 7 \times 5.45 \times 0.87$ = $11.68$

As the failure ratio is more than 1, user B should choose \kintof\ for his application.

\subsection*{Open Science}
We have prepared a website to
make the following material publicly available:
\begin{enumerate}
\item A documented version of the evaluation scripts for all experiments written in Matlab.
\item Further technical details for setting up the required test
  scenarios, e.g. a CAD file for the Siemens star, intensity and calibration checker board.
\end{enumerate} 

The website is available for use by other researchers
at:\\ \href{http://www.cg.informatik.uni-siegen.de/data/KinectRangeSensing/}
{http://www.cg.informatik.uni-siegen.de/data/KinectRangeSensing/.}
 
\subsection*{Acknowledgment} 
The authors would like to thank Microsoft Inc. for making the
prototype of the \kintof-cameras available via the Kinect For Windows
Developer Preview Program (K4W DPP) and Dr. Rainer Bornemann from the
Center for Sensor Systems of Northrhine-Westphalia (ZESS), Siegen, for
the reference measurements of the illumination signal for the
\kintof\ camera and for the support in measuring the ambient
illumination.

\bibliography{main}

\end{document}

%% file: images/pmd-principle.tex
\begin{picture}(0,0)%
\includegraphics{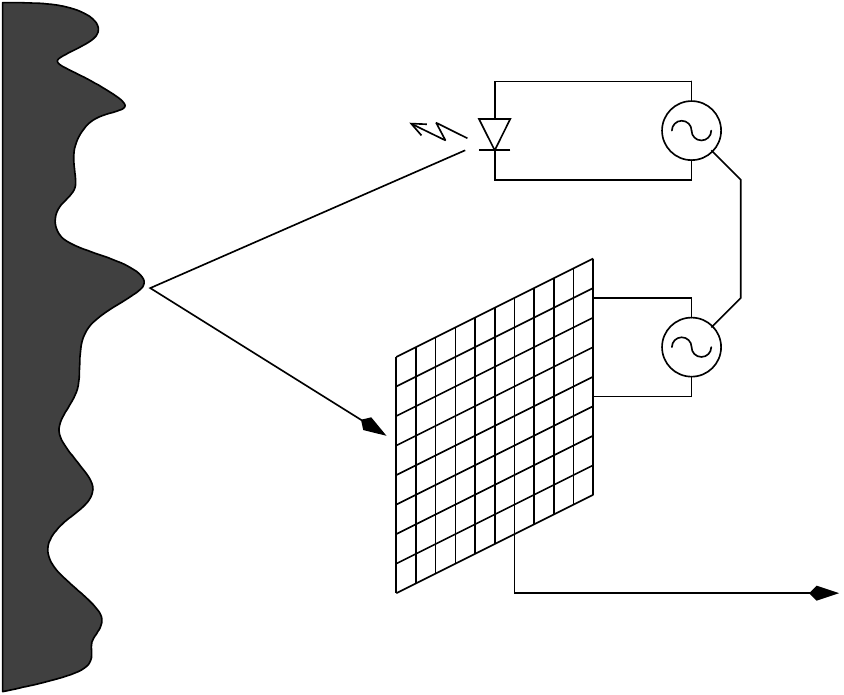}%
\end{picture}%
\setlength{\unitlength}{4144sp}%
\begingroup\makeatletter\ifx\SetFigFont\undefined%
\gdef\SetFigFont#1#2#3#4#5{%
  \reset@font\fontsize{#1}{#2pt}%
  \fontfamily{#3}\fontseries{#4}\fontshape{#5}%
  \selectfont}%
\fi\endgroup%
\begin{picture}(3849,3174)(4489,-5923)
\put(8011,-3391){\makebox(0,0)[b]{\smash{{\SetFigFont{12}{14.4}{\familydefault}{\mddefault}{\updefault}{\color[rgb]{0,0,0}$g^{\text{ill}}$}%
}}}}
\put(8011,-4381){\makebox(0,0)[b]{\smash{{\SetFigFont{12}{14.4}{\familydefault}{\mddefault}{\updefault}{\color[rgb]{0,0,0}$g^{\text{ref}}$}%
}}}}
\put(8101,-3886){\makebox(0,0)[b]{\smash{{\SetFigFont{12}{14.4}{\familydefault}{\mddefault}{\updefault}{\color[rgb]{0,0,0}sync}%
}}}}
\put(7606,-5416){\makebox(0,0)[b]{\smash{{\SetFigFont{12}{14.4}{\familydefault}{\mddefault}{\updefault}{\color[rgb]{0,0,0}data readout}%
}}}}
\put(5851,-4291){\makebox(0,0)[b]{\smash{{\SetFigFont{12}{14.4}{\familydefault}{\mddefault}{\updefault}{\color[rgb]{0,0,0}$s^{\text{ill}}$}%
}}}}
\put(6526,-3211){\makebox(0,0)[rb]{\smash{{\SetFigFont{12}{14.4}{\familydefault}{\mddefault}{\updefault}{\color[rgb]{0,0,0}IR light source}%
}}}}
\put(6526,-2986){\makebox(0,0)[rb]{\smash{{\SetFigFont{12}{14.4}{\familydefault}{\mddefault}{\updefault}{\color[rgb]{0,0,0}incoherent}%
}}}}
\put(6076,-5236){\makebox(0,0)[rb]{\smash{{\SetFigFont{12}{14.4}{\familydefault}{\mddefault}{\updefault}{\color[rgb]{0,0,0}mixer chip}%
}}}}
\end{picture}%
 

%% file: main.bbl
\begin{thebibliography}{57}
\expandafter\ifx\csname natexlab\endcsname\relax\def\natexlab#1{#1}\fi
\providecommand{\url}[1]{\texttt{#1}}
\providecommand{\href}[2]{#2}
\providecommand{\path}[1]{#1}
\providecommand{\DOIprefix}{doi:}
\providecommand{\ArXivprefix}{arXiv:}
\providecommand{\URLprefix}{URL: }
\providecommand{\Pubmedprefix}{pmid:}
\providecommand{\doi}[1]{\href{http://dx.doi.org/#1}{\path{#1}}}
\providecommand{\Pubmed}[1]{\href{pmid:#1}{\path{#1}}}
\providecommand{\bibinfo}[2]{#2}
\ifx\xfnm\relax \def\xfnm[#1]{\unskip,\space#1}\fi
\bibitem[{Bauer et~al.(2013)Bauer, Seitel, Hofmann, Blum, Wasza, Balda,
  Meinzer, Navab, Hornegger \& Maier-Hein}]{bauer13range-in-healtcare}
\bibinfo{author}{Bauer, S.}, \bibinfo{author}{Seitel, A.},
  \bibinfo{author}{Hofmann, H.}, \bibinfo{author}{Blum, T.},
  \bibinfo{author}{Wasza, J.}, \bibinfo{author}{Balda, M.},
  \bibinfo{author}{Meinzer, H.-P.}, \bibinfo{author}{Navab, N.},
  \bibinfo{author}{Hornegger, J.}, \& \bibinfo{author}{Maier-Hein, L.}
  (\bibinfo{year}{2013}).
\newblock \bibinfo{title}{Real-time range imaging in health care: A survey}.
\newblock In \bibinfo{editor}{M.~Grzegorzek}, \bibinfo{editor}{C.~Theobalt},
  \bibinfo{editor}{R.~Koch}, \& \bibinfo{editor}{A.~Kolb} (Eds.), {\it
  \bibinfo{booktitle}{Time-of-Flight and Depth Imaging. Sensors, Algorithms,
  and Applications}\/} (pp. \bibinfo{pages}{228--254}).
\newblock \bibinfo{publisher}{Springer} volume \bibinfo{volume}{8200} of {\it
  \bibinfo{series}{Lecture Notes in Computer Science}\/}.
\bibitem[{Beder et~al.(2007)Beder, Bartczak \& Koch}]{beder:2007:CompTOFSV}
\bibinfo{author}{Beder, C.}, \bibinfo{author}{Bartczak, B.}, \&
  \bibinfo{author}{Koch, R.} (\bibinfo{year}{2007}).
\newblock \bibinfo{title}{A comparison of pmd-cameras and stereo-vision for the
  task of surface reconstruction using patchlets}.
\newblock In {\it \bibinfo{booktitle}{Proc. IEEE Int. Conf. on Computer Vision
  and Pattern Recognition, 2007.}\/} (pp. \bibinfo{pages}{1--8}).
\newblock \bibinfo{organization}{IEEE}.
\bibitem[{Berger et~al.(2013)Berger, Meister, Nair \&
  Kondermann}]{berger2013state}
\bibinfo{author}{Berger, K.}, \bibinfo{author}{Meister, S.},
  \bibinfo{author}{Nair, R.}, \& \bibinfo{author}{Kondermann, D.}
  (\bibinfo{year}{2013}).
\newblock \bibinfo{title}{A state of the art report on kinect sensor setups in
  computer vision}.
\newblock In \bibinfo{editor}{M.~Grzegorzek}, \bibinfo{editor}{C.~Theobalt},
  \bibinfo{editor}{R.~Koch}, \& \bibinfo{editor}{A.~Kolb} (Eds.), {\it
  \bibinfo{booktitle}{{Time-of-Flight} and Depth Imaging. Sensors, Algorithms,
  and Applications}\/} (pp. \bibinfo{pages}{257--272}).
\newblock \bibinfo{publisher}{Springer} volume \bibinfo{volume}{8200} of {\it
  \bibinfo{series}{Lecture Notes in Computer Science}\/}.
\bibitem[{Berger et~al.(2011)Berger, Ruhl, Br{\"{u}}mmer, Schr{\"{o}}der,
  Scholz \& Magnor}]{bergerMMC11}
\bibinfo{author}{Berger, K.}, \bibinfo{author}{Ruhl, K.},
  \bibinfo{author}{Br{\"{u}}mmer, C.}, \bibinfo{author}{Schr{\"{o}}der, Y.},
  \bibinfo{author}{Scholz, A.}, \& \bibinfo{author}{Magnor, M.}
  (\bibinfo{year}{2011}).
\newblock \bibinfo{title}{Markerless motion capture using multiple color-depth
  sensors}.
\newblock In {\it \bibinfo{booktitle}{Proc. Vision, Modeling and Visualization
  {(VMV)} 2011}\/} (pp. \bibinfo{pages}{317--324}).
\bibitem[{Blake et~al.(2015)Blake, Echtler \& Kerl}]{blake15openKinectToF}
\bibinfo{author}{Blake, J.}, \bibinfo{author}{Echtler, F.}, \&
  \bibinfo{author}{Kerl, C.} (\bibinfo{year}{2015}).
\newblock \bibinfo{title}{{OpenKinect}: Open source drivers for the kinect for
  windows v2 device}.
\newblock
  \bibinfo{howpublished}{\url{https://github.com/OpenKinect/libfreenect2}}.
\newblock \bibinfo{note}{Last visited: Jan.~26th, 2015}.
\bibitem[{Bradski(2000)}]{bradskiTOL00}
\bibinfo{author}{Bradski, G.~R.} (\bibinfo{year}{2000}).
\newblock \bibinfo{title}{{The OpenCV Library}}.
\newblock {\it \bibinfo{journal}{Dr. Dobb's Journal of Software Tools}\/}, .
\bibitem[{Butler et~al.(2012)Butler, Izadi, Hilliges, Molyneaux, Hodges \&
  Kim}]{butlerSNS12}
\bibinfo{author}{Butler, A.}, \bibinfo{author}{Izadi, S.},
  \bibinfo{author}{Hilliges, O.}, \bibinfo{author}{Molyneaux, D.},
  \bibinfo{author}{Hodges, S.}, \& \bibinfo{author}{Kim, D.}
  (\bibinfo{year}{2012}).
\newblock \bibinfo{title}{Shake'n'sense: Reducing structured light interference
  when multiple depth cameras overlap}.
\newblock In {\it \bibinfo{booktitle}{Proc. Human Factors in Computing Systems
  ({ACM CHI})}\/}.
\newblock \bibinfo{address}{New York, NY, USA}: \bibinfo{publisher}{ACM}.
\bibitem[{Dorrington et~al.(2011)Dorrington, Godbaz, Cree, Payne \&
  Streeter}]{dorringtonSTR11}
\bibinfo{author}{Dorrington, A.}, \bibinfo{author}{Godbaz, J.},
  \bibinfo{author}{Cree, M.}, \bibinfo{author}{Payne, A.}, \&
  \bibinfo{author}{Streeter, L.} (\bibinfo{year}{2011}).
\newblock \bibinfo{title}{Separating true range measurements from multi-path
  and scattering interference in commercial range cameras}.
\newblock In {\it \bibinfo{booktitle}{Proc. IS\&T/SPIE Electronic Imaging}\/}
  (pp. \bibinfo{pages}{786404--786404}).
\bibitem[{Droeschel et~al.(2010)Droeschel, Holz \& Behnke}]{droeschel2010multi}
\bibinfo{author}{Droeschel, D.}, \bibinfo{author}{Holz, D.}, \&
  \bibinfo{author}{Behnke, S.} (\bibinfo{year}{2010}).
\newblock \bibinfo{title}{Multi-frequency phase unwrapping for time-of-flight
  cameras}.
\newblock In {\it \bibinfo{booktitle}{Intelligent Robots and Systems (IROS),
  2010 IEEE/RSJ International Conference on}\/} (pp.
  \bibinfo{pages}{1463--1469}).
\newblock \bibinfo{organization}{IEEE}.
\bibitem[{El-laithy et~al.(2012)El-laithy, Huang \& Yeh}]{el2012study}
\bibinfo{author}{El-laithy, R.}, \bibinfo{author}{Huang, J.}, \&
  \bibinfo{author}{Yeh, M.} (\bibinfo{year}{2012}).
\newblock \bibinfo{title}{Study on the use of microsoft kinect for robotics
  applications}.
\newblock In {\it \bibinfo{booktitle}{Proc. IEEE Symp. on Position Location and
  Navigation (PLANS)}\/} (pp. \bibinfo{pages}{1280--1288}).
\bibitem[{Evangelidis et~al.(2015)Evangelidis, Hansard \&
  Horaud}]{evangelidis2015fusion}
\bibinfo{author}{Evangelidis, G.}, \bibinfo{author}{Hansard, M.}, \&
  \bibinfo{author}{Horaud, R.} (\bibinfo{year}{2015}).
\newblock \bibinfo{title}{Fusion of range and stereo data for high-resolution
  scene-modeling}, .
\bibitem[{Falie \& Buzuloiu(2008)}]{falieDEC08}
\bibinfo{author}{Falie, D.}, \& \bibinfo{author}{Buzuloiu, V.}
  (\bibinfo{year}{2008}).
\newblock \bibinfo{title}{Distance errors correction for the time of flight
  ({ToF}) cameras}.
\newblock In {\it \bibinfo{booktitle}{Proc. European Conf. on Circuits and
  Systems for Communications}\/} (pp. \bibinfo{pages}{193--196}).
\bibitem[{Fechteler et~al.(2007)Fechteler, Eisert \&
  Rurainsky}]{fechtelerFAH07}
\bibinfo{author}{Fechteler, P.}, \bibinfo{author}{Eisert, P.}, \&
  \bibinfo{author}{Rurainsky, J.} (\bibinfo{year}{2007}).
\newblock \bibinfo{title}{Fast and high resolution {3D} face scanning}.
\newblock In {\it \bibinfo{booktitle}{{International Conference on Image
  Processing (ICIP)}}\/} (pp. \bibinfo{pages}{81--84}).
\newblock volume~\bibinfo{volume}{3}.
\bibitem[{Fiedler \& M{\"u}ller(2013)}]{fiedler2013impact}
\bibinfo{author}{Fiedler, D.}, \& \bibinfo{author}{M{\"u}ller, H.}
  (\bibinfo{year}{2013}).
\newblock \bibinfo{title}{Impact of thermal and environmental conditions on the
  kinect sensor}.
\newblock In {\it \bibinfo{booktitle}{Advances in Depth Image Analysis and
  Applications}\/} (pp. \bibinfo{pages}{21--31}).
\newblock \bibinfo{publisher}{Springer}.
\bibitem[{Gallo et~al.(2011)Gallo, Placitelli \& Ciampi}]{gallo2011controller}
\bibinfo{author}{Gallo, L.}, \bibinfo{author}{Placitelli, A.~P.}, \&
  \bibinfo{author}{Ciampi, M.} (\bibinfo{year}{2011}).
\newblock \bibinfo{title}{Controller-free exploration of medical image data:
  Experiencing the kinect}.
\newblock In {\it \bibinfo{booktitle}{Proc. IEEE Int. Symp. Computer-Based
  Medical Systems (CBMS)}\/} (pp. \bibinfo{pages}{1--6}).
\newblock \bibinfo{organization}{IEEE}.
\bibitem[{Hall-Holt \& Rusinkiewicz(2001)}]{hallholtSBC01}
\bibinfo{author}{Hall-Holt, O.}, \& \bibinfo{author}{Rusinkiewicz, S.}
  (\bibinfo{year}{2001}).
\newblock \bibinfo{title}{Stripe boundary codes for real-time structured-light
  range scanning of moving objects}.
\newblock In {\it \bibinfo{booktitle}{Proc. {IEEE} Int. Conf. on Computer
  Vision (ICCV)}\/} (pp. \bibinfo{pages}{359--366}).
\newblock volume~\bibinfo{volume}{2}.
\bibitem[{Han et~al.(2013)Han, Shao, Xu \& Shotton}]{han2013enhanced}
\bibinfo{author}{Han, J.}, \bibinfo{author}{Shao, L.}, \bibinfo{author}{Xu,
  D.}, \& \bibinfo{author}{Shotton, J.} (\bibinfo{year}{2013}).
\newblock \bibinfo{title}{Enhanced computer vision with microsoft {Kinect}
  sensor: A review}.
\newblock {\it \bibinfo{journal}{IEEE Transactions on Cybernetics}\/},  {\it
  \bibinfo{volume}{43}\/}, \bibinfo{pages}{1318--1334}.
\bibitem[{Hansard et~al.(2013)Hansard, Lee, Choi \& Horaud}]{hansard13time}
\bibinfo{author}{Hansard, M.}, \bibinfo{author}{Lee, S.},
  \bibinfo{author}{Choi, O.}, \& \bibinfo{author}{Horaud, R.}
  (\bibinfo{year}{2013}).
\newblock {\it \bibinfo{title}{Time-of-flight cameras: Principles, Methods, and
  Applications}\/}.
\newblock \bibinfo{publisher}{Springer}.
\bibitem[{Herrera~C. et~al.(2012)Herrera~C., Kannala \&
  Heikkil{\"a}}]{herreraJDA12}
\bibinfo{author}{Herrera~C., D.}, \bibinfo{author}{Kannala, J.}, \&
  \bibinfo{author}{Heikkil{\"a}, J.} (\bibinfo{year}{2012}).
\newblock \bibinfo{title}{Joint depth and color camera calibration with
  distortion correction}.
\newblock {\it \bibinfo{journal}{IEEE Trans. Pattern Anal. Mach. Intell.}\/},
  {\it \bibinfo{volume}{34}\/}, \bibinfo{pages}{2058--2064}.
\bibitem[{H\"ogg et~al.(2013)H\"ogg, Lefloch \& Kolb}]{hoeggRTM13}
\bibinfo{author}{H\"ogg, T.}, \bibinfo{author}{Lefloch, D.}, \&
  \bibinfo{author}{Kolb, A.} (\bibinfo{year}{2013}).
\newblock \bibinfo{title}{Real-time motion artifact compensation for {PMD-ToF}
  images}.
\newblock In {\it \bibinfo{booktitle}{Proc. Workshop Imaging New Modalities,
  German Conference of Pattern Recognition (GCPR)}\/} (pp.
  \bibinfo{pages}{273--288}).
\newblock \bibinfo{publisher}{Springer} volume \bibinfo{volume}{8200} of {\it
  \bibinfo{series}{LNCS}\/}.
\bibitem[{Ihrke et~al.(2010)Ihrke, Kutulakos, Lensch, Magnor \&
  Heidrich}]{ihrke2010transparent}
\bibinfo{author}{Ihrke, I.}, \bibinfo{author}{Kutulakos, K.~N.},
  \bibinfo{author}{Lensch, H.}, \bibinfo{author}{Magnor, M.}, \&
  \bibinfo{author}{Heidrich, W.} (\bibinfo{year}{2010}).
\newblock \bibinfo{title}{Transparent and specular object reconstruction}.
\newblock In {\it \bibinfo{booktitle}{Computer Graphics Forum}\/} (pp.
  \bibinfo{pages}{2400--2426}).
\newblock volume~\bibinfo{volume}{29}.
\bibitem[{Kadambi et~al.(2013)Kadambi, Whyte, Bhandari, Streeter, Barsi,
  Dorrington \& Raskar}]{kadambi2013coded}
\bibinfo{author}{Kadambi, A.}, \bibinfo{author}{Whyte, R.},
  \bibinfo{author}{Bhandari, A.}, \bibinfo{author}{Streeter, L.},
  \bibinfo{author}{Barsi, C.}, \bibinfo{author}{Dorrington, A.}, \&
  \bibinfo{author}{Raskar, R.} (\bibinfo{year}{2013}).
\newblock \bibinfo{title}{Coded time of flight cameras: sparse deconvolution to
  address multipath interference and recover time profiles}.
\newblock {\it \bibinfo{journal}{ACM Transactions on Graphics (TOG)}\/},  {\it
  \bibinfo{volume}{32}\/}, \bibinfo{pages}{167}.
\bibitem[{Kahlmann et~al.(2007)Kahlmann, Remondino \&
  Guillaume}]{kahlmannRIT07}
\bibinfo{author}{Kahlmann, T.}, \bibinfo{author}{Remondino, F.}, \&
  \bibinfo{author}{Guillaume, S.} (\bibinfo{year}{2007}).
\newblock \bibinfo{title}{Range imaging technology: new developments and
  applications for people identification and tracking}.
\newblock In {\it \bibinfo{booktitle}{Proc. of Videometrics IX - SPIE-IS\&T
  Electronic Imaging}\/}.
\newblock volume \bibinfo{volume}{6491}.
\newblock \bibinfo{note}{DOI: 10.1117/12.702512}.
\bibitem[{Kahlmann et~al.(2006)Kahlmann, Remondino \&
  Ingensand}]{kahlmannCFI06}
\bibinfo{author}{Kahlmann, T.}, \bibinfo{author}{Remondino, F.}, \&
  \bibinfo{author}{Ingensand, H.} (\bibinfo{year}{2006}).
\newblock \bibinfo{title}{Calibration for increased accuracy of the range
  imaging camera {SwissRanger}\texttrademark}.
\newblock {\it \bibinfo{journal}{Image Engineering and Vision Metrology
  (IEVM)}\/},  {\it \bibinfo{volume}{36}\/}, \bibinfo{pages}{136--141}.
\bibitem[{Keller et~al.(2013)Keller, Lefloch, Lambers, Izadi, Weyrich \&
  Kolb}]{keller:2013:PBF}
\bibinfo{author}{Keller, M.}, \bibinfo{author}{Lefloch, D.},
  \bibinfo{author}{Lambers, M.}, \bibinfo{author}{Izadi, S.},
  \bibinfo{author}{Weyrich, T.}, \& \bibinfo{author}{Kolb, A.}
  (\bibinfo{year}{2013}).
\newblock \bibinfo{title}{Real-time {3D} reconstruction in dynamic scenes using
  point-based fusion}.
\newblock In {\it \bibinfo{booktitle}{Proc. of Joint 3DIM/3DPVT Conference
  (3DV)}\/} (p.~\bibinfo{pages}{8}).
\bibitem[{Khoshelham \& Elberink(2012)}]{khoshelhamAAR12}
\bibinfo{author}{Khoshelham, K.}, \& \bibinfo{author}{Elberink, S.~O.}
  (\bibinfo{year}{2012}).
\newblock \bibinfo{title}{Accuracy and resolution of kinect depth data for
  indoor mapping applications}.
\newblock {\it \bibinfo{journal}{Sensors}\/},  {\it \bibinfo{volume}{12}\/},
  \bibinfo{pages}{1437--1454}. \URLprefix
  \url{http://www.mdpi.com/1424-8220/12/2/1437}.
  \DOIprefix\doi{10.3390/s120201437}.
\bibitem[{Kim et~al.(2009)Kim, Theobalt, Diebel, Kosecka, Miscusik \&
  Thrun}]{kim2009multi}
\bibinfo{author}{Kim, Y.~M.}, \bibinfo{author}{Theobalt, C.},
  \bibinfo{author}{Diebel, J.}, \bibinfo{author}{Kosecka, J.},
  \bibinfo{author}{Miscusik, B.}, \& \bibinfo{author}{Thrun, S.}
  (\bibinfo{year}{2009}).
\newblock \bibinfo{title}{Multi-view image and tof sensor fusion for dense 3d
  reconstruction}.
\newblock In {\it \bibinfo{booktitle}{Proc. Int. IEEE Conf. on Computer Vision
  Workshops (ICCV Workshops)}\/} (pp. \bibinfo{pages}{1542--1549}).
\newblock \bibinfo{organization}{IEEE}.
\bibitem[{Kolb et~al.(2010)Kolb, Barth, Koch \& Larsen}]{kolbTOF10}
\bibinfo{author}{Kolb, A.}, \bibinfo{author}{Barth, E.}, \bibinfo{author}{Koch,
  R.}, \& \bibinfo{author}{Larsen, R.} (\bibinfo{year}{2010}).
\newblock \bibinfo{title}{Time-of-flight cameras in computer graphics}.
\newblock {\it \bibinfo{journal}{J. Computer Graphics Forum}\/},  {\it
  \bibinfo{volume}{29}\/}, \bibinfo{pages}{141--159}.
\bibitem[{Konolige \& Mihelich(2012)}]{Konolige12openKinectSL}
\bibinfo{author}{Konolige, K.}, \& \bibinfo{author}{Mihelich, P.}
  (\bibinfo{year}{2012}).
\newblock \bibinfo{title}{{OpenKinect}: Ros' technical description of kinect
  calibration}.
\newblock
  \bibinfo{howpublished}{\url{http://wiki.ros.org/kinect_calibration/technical}}.
\newblock \bibinfo{note}{Last edited: Dec.~27th, 2012}.
\bibitem[{Kuhnert \& Stommel(2006)}]{kuhnert06fusion}
\bibinfo{author}{Kuhnert, K.}, \& \bibinfo{author}{Stommel, M.}
  (\bibinfo{year}{2006}).
\newblock \bibinfo{title}{Fusion of stereo-camera and {PMD}-camera data for
  real-time suited precise {3D} environment reconstruction}.
\newblock In {\it \bibinfo{booktitle}{Intelligent Robots and Systems (IROS)}\/}
  (pp. \bibinfo{pages}{4780--4785}).
\bibitem[{Langmann et~al.(2012)Langmann, Hartmann \&
  Loffeld}]{langmann:2012:DepthComparison}
\bibinfo{author}{Langmann, B.}, \bibinfo{author}{Hartmann, K.}, \&
  \bibinfo{author}{Loffeld, O.} (\bibinfo{year}{2012}).
\newblock \bibinfo{title}{Depth camera technology comparison and performance
  evaluation.}
\newblock In {\it \bibinfo{booktitle}{International Conference on Pattern
  Recognition Applications and Methods (ICPRAM)}\/} (pp.
  \bibinfo{pages}{438--444}).
\bibitem[{Lefloch et~al.(2013)Lefloch, Nair, Lenzen, Sch\"afer, Streeter, Cree,
  Koch \& Kolb}]{lefloch13foundation}
\bibinfo{author}{Lefloch, D.}, \bibinfo{author}{Nair, R.},
  \bibinfo{author}{Lenzen, F.}, \bibinfo{author}{Sch\"afer, H.},
  \bibinfo{author}{Streeter, L.}, \bibinfo{author}{Cree, M.},
  \bibinfo{author}{Koch, R.}, \& \bibinfo{author}{Kolb, A.}
  (\bibinfo{year}{2013}).
\newblock \bibinfo{title}{Technical foundation and calibration methods for
  time-of-flight cameras}.
\newblock In \bibinfo{editor}{M.~Grzegorzek}, \bibinfo{editor}{C.~Theobalt},
  \bibinfo{editor}{R.~Koch}, \& \bibinfo{editor}{A.~Kolb} (Eds.), {\it
  \bibinfo{booktitle}{Time-of-Flight and Depth Imaging. Sensors, Algorithms,
  and Applications}\/} (pp. \bibinfo{pages}{3--24}).
\newblock \bibinfo{publisher}{Springer} volume \bibinfo{volume}{8200} of {\it
  \bibinfo{series}{Lecture Notes in Computer Science}\/}.
\bibitem[{Lindner \& Kolb(2006)}]{lindnerLAD06}
\bibinfo{author}{Lindner, M.}, \& \bibinfo{author}{Kolb, A.}
  (\bibinfo{year}{2006}).
\newblock \bibinfo{title}{Lateral and depth calibration of pmd-distance
  sensors}.
\newblock In {\it \bibinfo{booktitle}{Proc. Int. Symp. on Visual Computing}\/}
  LNCS (pp. \bibinfo{pages}{524--533}).
\newblock \bibinfo{publisher}{Springer}.
\bibitem[{Lindner \& Kolb(2009)}]{lindnerCOM09}
\bibinfo{author}{Lindner, M.}, \& \bibinfo{author}{Kolb, A.}
  (\bibinfo{year}{2009}).
\newblock \bibinfo{title}{Compensation of motion artifacts for {Time-of-Flight}
  cameras}.
\newblock In {\it \bibinfo{booktitle}{Proc. Dynamic 3D Imaging}\/} (pp.
  \bibinfo{pages}{16--27}).
\newblock \bibinfo{publisher}{Springer} volume \bibinfo{volume}{5742} of {\it
  \bibinfo{series}{LNCS}\/}.
\bibitem[{Lindner et~al.(2010)Lindner, Schiller, Kolb \& Koch}]{lindnerTOF10}
\bibinfo{author}{Lindner, M.}, \bibinfo{author}{Schiller, I.},
  \bibinfo{author}{Kolb, A.}, \& \bibinfo{author}{Koch, R.}
  (\bibinfo{year}{2010}).
\newblock \bibinfo{title}{Time-of-flight sensor calibration for accurate range
  sensing}.
\newblock {\it \bibinfo{journal}{Computer Vision and Image Understanding}\/},
  {\it \bibinfo{volume}{114}\/}, \bibinfo{pages}{1318--1328}.
\bibitem[{Lottner et~al.(2011)Lottner, Langmann, Weihs \&
  Hartmann}]{lottner11scanning}
\bibinfo{author}{Lottner, O.}, \bibinfo{author}{Langmann, B.},
  \bibinfo{author}{Weihs, W.}, \& \bibinfo{author}{Hartmann, K.}
  (\bibinfo{year}{2011}).
\newblock \bibinfo{title}{Scanning 2d/3d monocular camera}.
\newblock In {\it \bibinfo{booktitle}{Proc. 3DTV Conference: The True
  Vision-Capture, Transmission and Display of 3D Video}\/} (pp.
  \bibinfo{pages}{1--4}).
\bibitem[{Macknojia et~al.(2013)Macknojia, Ch{\'a}vez-Arag{\'o}n, Payeur \&
  Lagani{\`e}re}]{macknojia2013KINSLCalib}
\bibinfo{author}{Macknojia, R.}, \bibinfo{author}{Ch{\'a}vez-Arag{\'o}n, A.},
  \bibinfo{author}{Payeur, P.}, \& \bibinfo{author}{Lagani{\`e}re, R.}
  (\bibinfo{year}{2013}).
\newblock \bibinfo{title}{Calibration of a network of kinect sensors for
  robotic inspection over a large workspace}.
\newblock In {\it \bibinfo{booktitle}{IEEE Workshop on Robot Vision (WORV)}\/}
  (pp. \bibinfo{pages}{184--190}).
\newblock \bibinfo{organization}{IEEE}.
\bibitem[{Meister et~al.(2012)Meister, Izadi, Kohli, H{\"a}mmerle, Rother \&
  Kondermann}]{Meister2012can}
\bibinfo{author}{Meister, S.}, \bibinfo{author}{Izadi, S.},
  \bibinfo{author}{Kohli, P.}, \bibinfo{author}{H{\"a}mmerle, M.},
  \bibinfo{author}{Rother, C.}, \& \bibinfo{author}{Kondermann, D.}
  (\bibinfo{year}{2012}).
\newblock \bibinfo{title}{When can we use {KinectFusion} for ground truth
  acquisition?}
\newblock In {\it \bibinfo{booktitle}{Proc. Workshop on Color-Depth Camera
  Fusion in Robotics}\/}.
\bibitem[{Nair et~al.(2013{\natexlab{a}})Nair, Meister, Lambers, Balda,
  Hofmann, Kolb, Kondermann \& J\"ahne}]{nair13ground-truth}
\bibinfo{author}{Nair, R.}, \bibinfo{author}{Meister, S.},
  \bibinfo{author}{Lambers, M.}, \bibinfo{author}{Balda, M.},
  \bibinfo{author}{Hofmann, H.}, \bibinfo{author}{Kolb, A.},
  \bibinfo{author}{Kondermann, D.}, \& \bibinfo{author}{J\"ahne, B.}
  (\bibinfo{year}{2013}{\natexlab{a}}).
\newblock \bibinfo{title}{Ground truth for evaluating time of flight imaging}.
\newblock In \bibinfo{editor}{M.~Grzegorzek}, \bibinfo{editor}{C.~Theobalt},
  \bibinfo{editor}{R.~Koch}, \& \bibinfo{editor}{A.~Kolb} (Eds.), {\it
  \bibinfo{booktitle}{Time-of-Flight and Depth Imaging. Sensors, Algorithms,
  and Applications}\/} (pp. \bibinfo{pages}{52--74}).
\newblock \bibinfo{publisher}{Springer} volume \bibinfo{volume}{8200} of {\it
  \bibinfo{series}{Lecture Notes in Computer Science}\/}.
\bibitem[{Nair et~al.(2013{\natexlab{b}})Nair, Ruhl, Lenzen, Meister,
  Sch\"afer, Garbe, Eisemann, Magnor \& Kondermann}]{nair13tof-stereo-fusion}
\bibinfo{author}{Nair, R.}, \bibinfo{author}{Ruhl, K.},
  \bibinfo{author}{Lenzen, F.}, \bibinfo{author}{Meister, S.},
  \bibinfo{author}{Sch\"afer, H.}, \bibinfo{author}{Garbe, C.},
  \bibinfo{author}{Eisemann, M.}, \bibinfo{author}{Magnor, M.}, \&
  \bibinfo{author}{Kondermann, D.} (\bibinfo{year}{2013}{\natexlab{b}}).
\newblock \bibinfo{title}{A survey on time-of-flight stereo fusion}.
\newblock In \bibinfo{editor}{M.~Grzegorzek}, \bibinfo{editor}{C.~Theobalt},
  \bibinfo{editor}{R.~Koch}, \& \bibinfo{editor}{A.~Kolb} (Eds.), {\it
  \bibinfo{booktitle}{{Time-of-Flight} and Depth Imaging. Sensors, Algorithms,
  and Applications}\/} (pp. \bibinfo{pages}{105--127}).
\newblock \bibinfo{publisher}{Springer} volume \bibinfo{volume}{8200} of {\it
  \bibinfo{series}{Lecture Notes in Computer Science}\/}.
\bibitem[{Newcombe et~al.(2011)Newcombe, Davison, Izadi, Kohli, Hilliges,
  Shotton, Molyneaux, Hodges, Kim \& Fitzgibbon}]{newcombe:2011:KINFU}
\bibinfo{author}{Newcombe, R.~A.}, \bibinfo{author}{Davison, A.~J.},
  \bibinfo{author}{Izadi, S.}, \bibinfo{author}{Kohli, P.},
  \bibinfo{author}{Hilliges, O.}, \bibinfo{author}{Shotton, J.},
  \bibinfo{author}{Molyneaux, D.}, \bibinfo{author}{Hodges, S.},
  \bibinfo{author}{Kim, D.}, \& \bibinfo{author}{Fitzgibbon, A.}
  (\bibinfo{year}{2011}).
\newblock \bibinfo{title}{Kinectfusion: Real-time dense surface mapping and
  tracking}.
\newblock In {\it \bibinfo{booktitle}{Proc. IEEE Int. Symp. Mixed and Augmented
  Reality (ISMAR)}\/} (pp. \bibinfo{pages}{127--136}).
\bibitem[{Nguyen et~al.(2012)Nguyen, Izadi \& Lovell}]{nguyen12kinectsl-noise}
\bibinfo{author}{Nguyen, C.~V.}, \bibinfo{author}{Izadi, S.}, \&
  \bibinfo{author}{Lovell, D.} (\bibinfo{year}{2012}).
\newblock \bibinfo{title}{Modeling kinect sensor noise for improved 3d
  reconstruction and tracking}.
\newblock In {\it \bibinfo{booktitle}{3D Imaging, Modeling, Processing,
  Visualization and Transmission (3DIMPVT), 2012 Second International
  Conference on}\/} (pp. \bibinfo{pages}{524--530}).
\newblock \bibinfo{organization}{IEEE}.
\bibitem[{Nie{\ss}ner et~al.(2013)Nie{\ss}ner, Zollh{\"o}fer, Izadi \&
  Stamminger}]{niessner:2013:KINFUH}
\bibinfo{author}{Nie{\ss}ner, M.}, \bibinfo{author}{Zollh{\"o}fer, M.},
  \bibinfo{author}{Izadi, S.}, \& \bibinfo{author}{Stamminger, M.}
  (\bibinfo{year}{2013}).
\newblock \bibinfo{title}{Real-time {3D} reconstruction at scale using voxel
  hashing}.
\newblock {\it \bibinfo{journal}{ACM Transactions on Graphics (TOG)}\/},  {\it
  \bibinfo{volume}{32}\/}, \bibinfo{pages}{169}.
\bibitem[{Ringbeck et~al.(2007)Ringbeck, M{\"o}ller \&
  Hagebeuker}]{ringbeck2007multidimensional}
\bibinfo{author}{Ringbeck, T.}, \bibinfo{author}{M{\"o}ller, T.}, \&
  \bibinfo{author}{Hagebeuker, B.} (\bibinfo{year}{2007}).
\newblock \bibinfo{title}{Multidimensional measurement by using {3-D PMD}
  sensors}.
\newblock {\it \bibinfo{journal}{Advances in Radio Science}\/},  {\it
  \bibinfo{volume}{5}\/}, \bibinfo{pages}{135--146}.
\bibitem[{Sabov \& Kr\"{u}ger(2010)}]{sabovIAC08}
\bibinfo{author}{Sabov, A.}, \& \bibinfo{author}{Kr\"{u}ger, J.}
  (\bibinfo{year}{2010}).
\newblock \bibinfo{title}{Identification and correction of flying pixels in
  range camera data}.
\newblock In {\it \bibinfo{booktitle}{Proc. Spring Conf. on Computer
  Graphics}\/} (pp. \bibinfo{pages}{135--142}).
\bibitem[{Schmidt \& Jahne(2011)}]{schmidtEAR11}
\bibinfo{author}{Schmidt, M.}, \& \bibinfo{author}{Jahne, B.}
  (\bibinfo{year}{2011}).
\newblock \bibinfo{title}{Efficient and robust reduction of motion artifacts
  for {3D} time-of-flight cameras}.
\newblock In {\it \bibinfo{booktitle}{Proc. Int. Conf. 3D Imaging (IC3D)}\/}
  (pp. \bibinfo{pages}{1--8}).
\newblock \DOIprefix\doi{10.1109/IC3D.2011.6584391}.
\bibitem[{Smisek et~al.(2011)Smisek, Jancosek \& Pajdla}]{smisek3WK11}
\bibinfo{author}{Smisek, J.}, \bibinfo{author}{Jancosek, M.}, \&
  \bibinfo{author}{Pajdla, T.} (\bibinfo{year}{2011}).
\newblock \bibinfo{title}{3d with kinect}.
\newblock In {\it \bibinfo{booktitle}{IEEE Int. Conf. Consumer Depth Cameras
  for Computer Vision (CDC4CV)}\/} (pp. \bibinfo{pages}{1154--1160}).
\bibitem[{Stoyanov et~al.(2011)Stoyanov, Louloudi, Andreasson \&
  Lilienthal}]{stoyanov:2011:DEPTHACCURACY}
\bibinfo{author}{Stoyanov, T.}, \bibinfo{author}{Louloudi, A.},
  \bibinfo{author}{Andreasson, H.}, \& \bibinfo{author}{Lilienthal, A.~J.}
  (\bibinfo{year}{2011}).
\newblock \bibinfo{title}{Comparative evaluation of range sensor accuracy in
  indoor environments.}
\newblock In {\it \bibinfo{booktitle}{European Conference on Mobile Robots
  (ECMR)}\/} (pp. \bibinfo{pages}{19--24}).
\bibitem[{Stoyanov et~al.(2013)Stoyanov, Mojtahedzadeh, Andreasson \&
  Lilienthal}]{Stoyanov:2013:CER}
\bibinfo{author}{Stoyanov, T.}, \bibinfo{author}{Mojtahedzadeh, R.},
  \bibinfo{author}{Andreasson, H.}, \& \bibinfo{author}{Lilienthal, A.~J.}
  (\bibinfo{year}{2013}).
\newblock \bibinfo{title}{Comparative evaluation of range sensor accuracy for
  indoor mobile robotics and automated logistics applications}.
\newblock {\it \bibinfo{journal}{Robot. Auton. Syst.}\/},  {\it
  \bibinfo{volume}{61}\/}, \bibinfo{pages}{1094--1105}.
\bibitem[{Vera et~al.(2011)Vera, Gimeno, Coma \&
  Fern{\'a}ndez}]{vera2011augmented}
\bibinfo{author}{Vera, L.}, \bibinfo{author}{Gimeno, J.},
  \bibinfo{author}{Coma, I.}, \& \bibinfo{author}{Fern{\'a}ndez, M.}
  (\bibinfo{year}{2011}).
\newblock \bibinfo{title}{Augmented mirror: interactive augmented reality
  system based on kinect}.
\newblock In {\it \bibinfo{booktitle}{Human-Computer Interaction--INTERACT}\/}
  (pp. \bibinfo{pages}{483--486}).
\newblock \bibinfo{publisher}{Springer} volume \bibinfo{volume}{6949} of {\it
  \bibinfo{series}{LNCS}\/}.
\bibitem[{Wiedemann et~al.(2008)Wiedemann, Sauer, Driewer \&
  Schilling}]{wiedemann:2008:TOFCharact}
\bibinfo{author}{Wiedemann, M.}, \bibinfo{author}{Sauer, M.},
  \bibinfo{author}{Driewer, F.}, \& \bibinfo{author}{Schilling, K.}
  (\bibinfo{year}{2008}).
\newblock \bibinfo{title}{Analysis and characterization of the pmd camera for
  application in mobile robotics}.
\newblock In {\it \bibinfo{booktitle}{Proc. IFAC World Congress}\/} (pp.
  \bibinfo{pages}{6--11}).
\bibitem[{Xu et~al.(2013)Xu, Perry \& Hills}]{xu2013method}
\bibinfo{author}{Xu, Z.}, \bibinfo{author}{Perry, T.}, \&
  \bibinfo{author}{Hills, G.} (\bibinfo{year}{2013}).
\newblock \bibinfo{title}{Method and system for multi-phase dynamic calibration
  of three-dimensional {(3D)} sensors in a time-of-flight system}.
\newblock \bibinfo{note}{US Patent 8,587,771}.
\bibitem[{Xu et~al.(1998)Xu, Schwarte, Heinol, Buxbaum \& Ringbeck}]{xuSPP99}
\bibinfo{author}{Xu, Z.}, \bibinfo{author}{Schwarte, R.},
  \bibinfo{author}{Heinol, H.}, \bibinfo{author}{Buxbaum, B.}, \&
  \bibinfo{author}{Ringbeck, T.} (\bibinfo{year}{1998}).
\newblock \bibinfo{title}{Smart pixel -- photonic mixer device ({PMD})}.
\newblock In {\it \bibinfo{booktitle}{Proc. Int. Conf. on Mechatron. \& Machine
  Vision}\/} (pp. \bibinfo{pages}{259--264}).
\bibitem[{Zach et~al.(2007)Zach, Pock \& Bischof}]{ZachADB07}
\bibinfo{author}{Zach, C.}, \bibinfo{author}{Pock, T.}, \&
  \bibinfo{author}{Bischof, H.} (\bibinfo{year}{2007}).
\newblock \bibinfo{title}{A duality based approach for realtime {TV-L} 1
  optical flow}.
\newblock In {\it \bibinfo{booktitle}{Proc. German Conference on Pattern
  Recognition (DAGM)}\/} (pp. \bibinfo{pages}{214--223}).
\newblock \bibinfo{publisher}{Springer}.
\bibitem[{Zhang et~al.(2002)Zhang, Curless \& Seitz}]{zhangRSA02}
\bibinfo{author}{Zhang, L.}, \bibinfo{author}{Curless, B.}, \&
  \bibinfo{author}{Seitz, S.~M.} (\bibinfo{year}{2002}).
\newblock \bibinfo{title}{Rapid shape acquisition using color structured light
  and multi-pass dynamic programming}.
\newblock In {\it \bibinfo{booktitle}{{IEEE} Int. Sym. on {3D} Data Processing,
  Visualization, and Transmission}\/} (pp. \bibinfo{pages}{24--36}).
\bibitem[{Zhang \& Huang(2004)}]{zhangHRR04}
\bibinfo{author}{Zhang, S.}, \& \bibinfo{author}{Huang, P.}
  (\bibinfo{year}{2004}).
\newblock \bibinfo{title}{High-resolution, real-time {3D} shape acquisition}.
\newblock In {\it \bibinfo{booktitle}{Proceedings of the 2004 Conference on
  Computer Vision and Pattern Recognition Workshop ({CVPRW})}\/} (pp.
  \bibinfo{pages}{28--36}).
\newblock \bibinfo{address}{Washington, DC, USA}: \bibinfo{publisher}{IEEE
  Computer Society} volume~\bibinfo{volume}{3}.
\bibitem[{Zhang(2000)}]{zhang00Calib}
\bibinfo{author}{Zhang, Z.} (\bibinfo{year}{2000}).
\newblock \bibinfo{title}{A flexible new technique for camera calibration}.
\newblock {\it \bibinfo{journal}{Pattern Analysis and Machine Intelligence,
  IEEE Transactions on}\/},  {\it \bibinfo{volume}{22}\/},
  \bibinfo{pages}{1330--1334}.

\end{thebibliography}
